\documentclass[conference]{IEEEtran}
\IEEEoverridecommandlockouts
\usepackage{amsmath,amsfonts}
\usepackage{array}
\usepackage{textcomp}
\usepackage{stfloats} 
\usepackage{url}
\usepackage{verbatim}
\usepackage[table]{xcolor}
\usepackage{xcolor}
\usepackage{graphicx}
\usepackage[tight,small]{subfigure}
\usepackage{cite}
\usepackage{bm}   
\usepackage{bbm}
\usepackage{amsthm,amsmath,amssymb,amsfonts}
\usepackage{mathrsfs}
\usepackage{newunicodechar}
\usepackage[linesnumbered,ruled,vlined]{algorithm2e}
\usepackage{algpseudocode}
\usepackage{enumitem}
\usepackage{ragged2e} 
\usepackage{adjustbox}
\usepackage{booktabs,makecell, multirow, tabularx}
\usepackage{graphicx}
\usepackage{booktabs}
\usepackage{arydshln}  
\definecolor{verylightgray}{rgb}{0.5, 0.6, 0.85}
\definecolor{veryverylightblue}{rgb}{0.765, 0.824, 0.918}
\usepackage{caption}
\usepackage{fancyhdr}
\newtheorem{theorem}{Theorem}
\newtheorem{assumption}{Assumption}
\usepackage[margin=0.65in, bottom=0.98in]{geometry}

\newtheorem{lemma}{Lemma}
\def\BibTeX{{\rm B\kern-.05em{\sc i\kern-.025em b}\kern-.08em
		T\kern-.1667em\lower.7ex\hbox{E}\kern-.125emX}}
\algnewcommand{\LineComment}[1]{\State \(\triangleright\) #1}
\def\BibTeX{{\rm B\kern-.05em{\sc i\kern-.025em b}\kern-.08em
		T\kern-.1667em\lower.7ex\hbox{E}\kern-.125emX}}
\graphicspath{{Figures/}}
\newif\ifshowrevisioncolors
\showrevisioncolorsfalse 

\ifshowrevisioncolors
\else
  \colorlet{red}{black}
  \colorlet{blue}{black}
\fi

\begin{document}

\title{ELSA: Efficient LLM-Centric Split Aggregation for Privacy-Aware Hierarchical Federated Learning \\ over the Network Edge}
\author{
	\IEEEauthorblockN{
		Xiaohong Yang,
		Tong Xie,
		Minghui Liwang, \textit{Senior Member}, \textit{IEEE}, \\
		Chikai Shang, 
		Yang Lu, \textit{Senior Member}, \textit{IEEE},  
		Zhenzhen Jiao, Liqun Fu, \textit{Senior Member}, \textit{IEEE},\\ 
        Seyyedali Hosseinalipour, \textit{Senior Member}, \textit{IEEE}, Xianbin Wang, \textit{Fellow}, \textit{IEEE}
	}
	\vspace{-0.3 cm}   
    \thanks{This work was supported in part by Shanghai Pujiang Programme under Grant no. 24PJD117; National Natural Science Foundation of China under Grant nos. 62271424, 62503365, U23A20281; Shanghai Municipal Science and Technology Major Project under Grant no. 2021SHZDZX0100; Guizhou Provincial Science and Technology Plan Project ([2026]209); Aeronautical Science Foundation of China under Grant no. 2023Z066038001; Chinese Academy of Engineering, Strategic Research and Consulting Program under Grant no. 2023-XZ-65. Xiaohong Yang, Chikai Shang, Yang Lu and Liqun Fu are with the School of Informatics, Xiamen University, Fujian, China. Tong Xie is with the National University of Singapore. Minghui Liwang is with the Department of Control Science and Engineering, the National Key Laboratory of Autonomous Intelligent Unmanned Systems, and also with Frontiers Science Center for Intelligent Autonomous Systems, Ministry of Education, Tongji University, Shanghai, China. Zhenzhen Jiao is with the Aeromind Technology Co., Ltd., Shenzhen, China. Seyyedali Hosseinalipour is with the Department of Electrical Engineering, University at Buffalo-SUNY, Buffalo, NY USA. Xianbin Wang is with the Department of Electrical and Computer Engineering, Western University, Ontario, Canada. Corresponding author: Minghui Liwang}
}
\maketitle

\makeatletter 
\let\myorg@bibitem\bibitem
\def\bibitem#1#2\par{%
	\@ifundefined{bibitem@#1}{%
		\myorg@bibitem{#1}#2\par
	}{%
		\begingroup
		\color{\csname bibitem@#1\endcsname}%
		\myorg@bibitem{#1}#2\par
		\endgroup
	}%
}
\newcommand*{\bibitem@HSplitLoRA}{red}
\newcommand*{\bibitem@AdaptSFL}{red}
\newcommand*{\bibitem@PipeSFL}{red}
\newcommand*{\bibitem@ESFL}{red}
\newcommand*{\bibitem@CHEESE}{red}
\newcommand*{\bibitem@FedClust}{red}
\newcommand*{\bibitem@liuone}{red}
\newcommand*{\bibitem@EnhancingSL}{red}
\newcommand*{\bibitem@FwdLLM}{red}
\newcommand*{\bibitem@Splitcom}{red}
\newcommand*{\bibitem@linsplitlora}{red}
\newcommand*{\bibitem@KLFedDis}{red}
\newcommand*{\bibitem@HTFL}{red}
\newcommand*{\bibitem@FedCM}{red}
\newcommand*{\bibitem@qiangdeploying}{blue}
\newcommand*{\bibitem@DLGtext}{red}
\newcommand*{\bibitem@chengpersonalized}{red}
\newcommand*{\bibitem@shrivastavasketching}{red}
\newcommand*{\bibitem@FedTKD}{red}
\newcommand*{\bibitem@FLIS}{red}
\newcommand*{\bibitem@TranFL}{red}

\makeatother

\begin{abstract}
Training large language models (LLMs) in resource-constrained edge networks poses fundamental challenges in computation, communication, data heterogeneity, and privacy. To address these challenges, we propose ELSA (Efficient LLM-Centric Split Aggregation), a distributed fine-tuning framework that integrates split learning (SL) with hierarchical federated learning (HFL) to coordinate computation and model aggregation across heterogeneous clients, edge servers, and the cloud. ELSA introduces three key innovations. First, it employs a task-agnostic, behavior-aware client clustering mechanism that constructs semantic fingerprints from public probe inputs using symmetric Kullback-Leibler (KL) divergence, complemented by prediction-consistency-based trust scoring and latency-aware edge assignment to account for data heterogeneity, client reliability, and communication constraints. Second, it adopts a resource-aware dynamic model partitioning strategy that distributes the LLM across clients and edge servers in three segments, while reserving the cloud for adapter aggregation. This design coordinates computation across heterogeneous nodes while balancing on-device computational demands and global convergence behavior. Third, it integrates computational sketches with semantic subspace orthogonal perturbation (SS-OP) to reduce communication overhead and mitigate privacy leakage during model exchanges. Extensive experiments across diverse NLP tasks demonstrate that ELSA outperforms state-of-the-art baselines in model performance, convergence behavior, and robustness. Further, on-device profiling on NVIDIA RTX 4090 GPUs and Jetson AGX Orin edge devices characterizes the memory footprint, energy consumption, and execution latency of key training stages, providing empirical evidence of ELSA's practical feasibility.
\end{abstract}

\begin{IEEEkeywords}
Edge Computing, Distributed Computing, Hierarchical Federated Learning, Split Learning, LLM.
\end{IEEEkeywords}

\vspace{-0.3 cm}
\section{Introduction}
\noindent \IEEEPARstart{L}{arge} language models (LLMs) have demonstrated remarkable performance across a wide range of language understanding and generation tasks, motivating their rapid adoption in both research and real-world systems \cite{TowardEG,EdgeShard,UniOrch}. These models typically undergo a large-scale pre-training phase on centralized corpora to learn general language representations. This is followed by a fine-tuning phase that adapts the model to specific tasks, users, or environments, making fine-tuning the primary focus of LLM deployment at the network edge \cite{RedCT,QSA}. Nevertheless, integrating LLM fine-tuning pipelines at the network edge faces fundamental limitations due to the constrained resources of edge devices, including compute, communication, memory, and energy. Compounding these constraints, data collected across edge devices is often highly skewed and non-independent and identically distributed (non-IID), which induces local model bias and degrades the performance of locally trained models. Together, these challenges highlight the need for a principled, architecture-aware distributed learning framework capable of enabling resource-efficient and robust LLM fine-tuning across edge devices \cite{FewS,FineT}.

Federated learning (FL) offers a natural solution to the aforementioned challenges by enabling collaborative model training across geo-distributed edge devices with non-IID data. \textcolor{blue}{Conventional FL alternates between local model fine-tuning at edge devices and periodic aggregation of model updates at a central server \cite{FedMPS,TranFL}. As the system scales, however, direct communication between all devices and a single server can incur substantial overhead and cause severe backhaul congestion. To alleviate this bottleneck, hierarchical federated learning (HFL) introduces intermediate nodes (e.g., edge servers or gateways) that aggregate local model updates before forwarding them to a central server \cite{Multicenter,ClientSD}. Although HFL improves scalability and reduces backhaul traffic, it typically assumes that devices can compute and transmit full-model updates, an increasingly impractical requirement for LLMs given their large parameter counts and substantial memory demands. These limitations motivate the integration of split learning (SL), which partitions models across devices and servers, limiting device-side training to lightweight segments while offloading resource-intensive components to edge or cloud servers \cite{EfficientP,kim2024battery}. Despite recent efforts to integrate HFL and SL for edge LLM fine-tuning \cite{SplitLLM}, a unified framework jointly addressing semantic heterogeneity, client and network dynamics, hierarchical coordination, and activation-level privacy remains lacking.}

\begin{table*}[t]
	\centering
	\caption{Summary of related works on distributed LLM training and fine-tuning.}
	\label{tab:related_works}
	\renewcommand{\arraystretch}{0.9}
	\footnotesize
	\begin{tabular}{c p{3.5cm} p{12cm}}
		\hline
		\textbf{Reference} & \textbf{Research Field} & \textbf{Summary of Contribution} \\
		\hline
		{\cite{Federal}} & FL + Sentiment Analysis + LLM & Proposes a FL framework combining LLM with multi-scale CNN for sentiment analysis on COVID-19 social media data, enabling collaborative training across edge networks while preserving data privacy and improving communication efficiency through parameter aggregation. \\
		{\cite{FedBERT}} & FL + SL + LLM Pre-training & Introduces a federated SL approach for pre-training LLMs across distributed clients without sharing raw data, demonstrating effectiveness on GLUE tasks while maintaining privacy and reducing computational burden on resource-constrained devices. \\
		{\cite{MT-FBERT}} & FL + LLM + Network Security & Develops an efficient FL framework for malicious traffic detection by pre-training LLM on unlabeled network data and dynamically updating significant neurons, achieving superior detection performance while protecting critical infrastructure data privacy. \\
		{\cite{ROFED}} & FL + LLM + Security & Integrates split FL with adaptive jamming defense for robust LLM training in adversarial wireless environments, enhancing privacy and resilience through multi-modal defense strategies combining differential privacy and beamforming. \\
		{\textcolor{blue}{\cite{EnabledR}}} & \textcolor{blue}{Semantic Comm. + Fed-Split} & \textcolor{blue}{Designs a knowledge-aware federated-split architecture for semantic communications, employing PPO with frozen LLM to optimize resources, reducing training latency and improving reconstruction accuracy.} \\
		{\cite{Privacy-Awa}} & IoT + SFL + Privacy & Develops a privacy-aware split FL scheme for IoT using Fisher information metrics, jointly optimizing split layer and resources to balance convergence speed, energy consumption, and privacy leakage. \\
		{\cite{linsplitlora}} & SL + PEFT & Proposes the first SL framework for LLM fine-tuning combining federated parallel training with model splitting to significantly reduce client computational burden and enhance training efficiency. \\
		{\cite{SplitFrozen}} & Heterogeneous Devices + SL & Introduces a SL framework freezing device-side layers while centralizing LoRA fine-tuning on server, significantly reducing device computations and training time under data imbalance. \\
		{\textcolor{blue}{\cite{SplitLLM}}} & \textcolor{blue}{Wireless + Hierarchical SL} & \textcolor{blue}{Proposes a hierarchical SL scheme with cloud-edge-user three-tier architecture, distributing model parts to reduce peak memory usage through parallel edge training and adapter aggregation.} \\
		{\cite{qiangdeploying}} & Resource-Constrained + SFL & Develops a split FL framework for large AI models on edge devices, incorporating quantization and resource control to lower memory requirements and energy consumption while maintaining scalability. \\
		{\cite{Split-Fine-T}} & Wireless + Split Fine-Tuning & Proposes a split fine-tuning scheme splitting LLM between edge server and devices with activation compression, optimizing resources to reduce fine-tuning delay and communication overhead significantly. \\
		{\cite{PrivacyP}} & Vision + SL & Proposes a lightweight SL framework combining mask mechanism with differential privacy for large-scale vision pre-training in collaborative learning scenarios. \\
        {\textcolor{red}{\cite{HSplitLoRA}}} & \textcolor{red}{LLM + Heterogeneous SL + PEFT} & \textcolor{red}{Adaptively selects client-specific LoRA ranks and model split points based on heterogeneous computing budgets and enables noise-free adapter aggregation.} \\
		\hline
		ELSA & Edge Intelligence + Dist. LLM & Integrates SL with HFL through three-part dynamic model partitioning across client-edge-cloud, incorporating behavior-aware clustering and computational sketches to balance computation cost with convergence stability and privacy. \\
		\hline
	\end{tabular}
    \vspace{-2mm}
\end{table*}

Specifically, this work aims to systematically investigate the potential of jointly integrating  SL and HFL for LLM fine-tuning, forming a novel hybrid SL-HFL framework depicted in Fig. 1 that entails: \textit{(i)} devising SL-driven model partitioning strategies that preserve the effectiveness and convergence of distributed LLM fine-tuning; \textit{(ii)} developing mechanisms to address non-IID data distributions, client unreliability, and communication constraints in distributed LLM fine-tuning under SL-based model partitioning and HFL-style model aggregation; and \textit{(iii)} mitigating privacy risks arising during training and model aggregation in hybrid SL-HFL settings, particularly those associated with data exposure or reconstruction through the observation of shared intermediate representations and model updates across the network. Our key contributions are:

\noindent $\bullet$ We propose ELSA (Efficient LLM-centric Split Aggregation), a framework that integrates SL and HFL to address device data heterogeneity and resource constraints for LLM fine-tuning at the network edge. Within ELSA, devices and edge servers collaboratively train edge-side model segments, while a cloud server performs global model aggregation.

\noindent $\bullet$ To enhance robustness under data heterogeneity, client unreliability, and communication constraints in distributed LLM fine-tuning, we design a task-agnostic, behavior-aware hierarchical clustering mechanism. This strategy is further enhanced with \textit{(i)} a trustworthiness score based on prediction consistency (to filter poisoned or noisy clients), and \textit{(ii)} a latency-aware edge server assignment policy. In addition, we develop an SL-aware dynamic model segmentation strategy that explicitly leverages the heterogeneous capabilities of devices and edge servers, achieving a principled balance between training cost and model performance.

\noindent $\bullet$ To reduce communication overhead and improve system efficiency, we introduce a model compression scheme based on computational sketches. Building on this design, we incorporate semantic subspace orthogonal perturbation (SS-OP) to preserve data privacy throughout the training process, mitigating the risk of adversarial attacks and information leakage during model update exchanges across the network.

\noindent $\bullet$ Extensive evaluations on natural language processing (NLP) tasks, including text classification and information extraction, demonstrate that ELSA consistently outperforms state-of-the-art benchmarks, exhibiting superior model performance, convergence stability, and robustness under heterogeneous data and resource profiles of edge devices.

\section{Literature Review}
\subsection{FL-driven Fine-Tuning of LLMs over the Network Edge}
FL has been recently explored as a promising paradigm for LLM fine-tuning in network edge environments. For example, \cite{Tri_AFLLM} introduces an adaptive asynchronous accelerated FL for LLMs to enhance resource utilization efficiency and model accuracy in edge computing environments. \cite{Federal} designs FL-empowered collaborative training for bidirectional encoder representations from transformers (BERT) models and multi-scale convolutional neural networks (CNNs). Also, \cite{eFedLLM} facilitates multi-user collaborative training of LLMs by integrating FL within model parallelism strategies, while \cite{FM2} proposes a lightweight large-model-based joint multi-task multi-domain learning framework. Moreover, \cite{MT-FBERT} presents an FL-driven LLM fine-tuning malicious traffic detection method using BERT, and \cite{EnablingF} introduce a collaborative training framework for Transformer block classification based on FL.

Despite these contributions, existing FL-based LLM fine-tuning approaches largely overlook the fundamental scalability limitations that arise in large-scale edge networks. Specifically, as the number of participating devices grows, conventional FL architectures, where all devices directly communicate with a single aggregation server, suffer from severe communication congestion, backhaul link bottlenecks, and coordination overhead, issues that are further exacerbated by device data heterogeneity and limited resources. This implies that existing solutions relying on such flat topologies, e.g., \cite{ROFED,EnabledR,Privacy-Awa,linsplitlora,qiangdeploying}, may become ineffective/inefficient in large-scale scenarios due to these bottlenecks. These challenges naturally motivate HFL, which introduces intermediate aggregation points to improve scalability and communication efficiency. However, while HFL has shown promise in narrow-model settings \cite{MultiHFL,OptimizingHFL,FM2}, its application to LLM fine-tuning remains largely unexplored. Further, the massive parameter sizes and memory demands of LLMs introduce new constraints that require HFL architectures to be explicitly adapted to large-model training, a gap that this work aims to address through the integration of SL.

\subsection{Review of the Use of SL over the Network Edge}
\textcolor{blue}{LLM fine-tuning on resource-constrained edge devices typically follows two main strategies: \textit{(i)} parameter-efficient fine-tuning, such as low-rank adaptation (LoRA), to update a limited set of parameters; and \textit{(ii)} model partitioning to distribute training across computing nodes. FlowLLM \cite{FlowLLM} partitions an LLM into two segments for collaborative client-server training. CHEESE \cite{CHEESE} pools edge resources through device-to-device (D2D) communications within a distributed hybrid federated split learning framework, while FedBERT \cite{FedBERT} applies split learning to collaborative BERT training. SplitLoRA \cite{linsplitlora} combines LoRA with SFL to offload the main training workload from clients to the server and provides an open-source benchmark for split LLM fine-tuning.}

\textcolor{blue}{Among two-tier LLM fine-tuning approaches, SplitFrozen \cite{SplitFrozen} freezes device-side layers and combines server-side LoRA fine-tuning with pipeline parallelism to reduce device computation and training time. The privacy-aware split federated learning (SFL) framework in \cite{Privacy-Awa} quantifies layer-wise privacy leakage from smashed data via Fisher information and jointly optimizes the split layer, transmit power, and bandwidth allocation to balance privacy, convergence time, and energy consumption. Split-Fine-T \cite{Split-Fine-T} shares a server-side model across parallel clients and jointly optimizes transformer-block allocation, activation compression, and spectrum allocation to reduce fine-tuning latency and communication overhead. HSplitLoRA \cite{HSplitLoRA} adapts client-specific LoRA ranks and split locations to heterogeneous computing budgets. Related SFL studies address adaptive splitting, scheduling, and resource allocation. AdaptSFL \cite{AdaptSFL} adjusts model splitting and client-side aggregation to balance convergence and communication--computation latency. PipeSFL \cite{PipeSFL} improves server-side scheduling through hybrid asynchronous/synchronous execution, while ESFL \cite{ESFL} jointly allocates user workloads and server computing resources in heterogeneous wireless settings. Beyond two-tier architectures, hierarchical SFL \cite{HSFL} extends model splitting and aggregation to multi-tier cloud--edge systems, jointly optimizing both based on convergence analysis. For hierarchical LLM fine-tuning, SplitLLM \cite{SplitLLM} partitions the pretrained model and LoRA adapters across user, edge, and cloud tiers to alleviate communication congestion and server memory pressure, while uploading user- and edge-side adapters for cloud aggregation.}

\textcolor{red}{Nevertheless, several coupled challenges remain insufficiently addressed. Existing methods characterize heterogeneity primarily via computing capacity, memory, communication conditions, or data distributions, rather than local models' learned semantic behavior \cite{linsplitlora,SplitFrozen,Privacy-Awa,Split-Fine-T,HSplitLoRA,AdaptSFL,ESFL}. Their splitting and scheduling strategies typically assume predefined client--server relationships, without jointly considering client reliability, link quality, and dynamic client--edge association \cite{linsplitlora,Split-Fine-T,HSplitLoRA,AdaptSFL,PipeSFL,ESFL}. Hierarchical SFL \cite{HSFL} addresses multi-tier model splitting and aggregation but focuses on general convergence and system optimization rather than LLM-specific semantic heterogeneity, client reliability, or activation-level privacy. SplitLLM introduces three-tier LLM fine-tuning but aggregates user- and edge-side adapters at the cloud, without behavior-aware client organization, reliability-aware client--edge association, or staged edge--cloud adapter aggregation \cite{SplitLLM}. The privacy-aware SFL framework in \cite{Privacy-Awa} evaluates smashed-data privacy risks and mitigates them through split-point and resource optimization, but neither directly protects exchanged intermediate activations nor integrates such protection with semantic client grouping and reliability-aware association. Thus, existing studies largely address these challenges in isolation rather than within a unified framework.}

\textcolor{blue}{
Accordingly, we note that existing distributed LLM fine-tuning frameworks primarily characterize data and system heterogeneity, often overlooking model behavioral heterogeneity: clients with similar data may learn distinct semantic representations. As summarized in Table \ref{tab:related_works}, ELSA addresses this gap through a semantics-centric framework. Task-agnostic probing quantifies differences in how local models interpret the same inputs, enabling behavior-aware client clustering beyond label statistics. Integrating this approach with resource-adaptive model splitting and privacy-preserving sketching, ELSA unifies semantic alignment, resource efficiency, and privacy.}

\begin{figure}[htbp]
	\centering
	\subfigure{
		\includegraphics[trim=0.8cm 0cm 0cm 0cm, clip, width=0.98\columnwidth]{simulate_results/main_framework0306.pdf}
	}
	\vspace{-5mm}
	\caption{A schematic of our proposed ELSA: clients and edge servers collaboratively fine-tuning LLMs via LoRA, with activations compressed during transmission; the cloud globally aggregates the LoRA parameters.}
	\label{fig_1}
	\vspace{-3mm}
\end{figure}

\section{Proposed Method}
\subsection{Problem Formulation}
We consider distributed learning through an HFL architecture, where the system consists of a cloud server, multiple edge servers $\mathcal{K}= \{1,...,k,...,K\}$ , and a set of clients $\mathcal{N} = \{1,...,n,...,N\}$, as illustrated in the upper right portion of Fig. 1. Each client $n$ owns a local dataset  $\mathcal{D}_n=\{(\mathbf{x}_j,y_j)|1\leq j\leq|\mathcal{D}_n|\}$, where $\mathbf{x}_{j}$ and $y_{j}$ refer to the feature vector and the label of the $j^\text{th}$ local data point, respectively. Subsequently, given a pre-trained LLM with frozen backbone parameters $w^{\mathsf{LLM}}$ and the trainable fine-tuning parameter $\theta$, the loss function for a data sample $(\mathbf{x}_j,y_j)$ is defined via the task specific loss (e.g., cross entropy for classification):
\begin{align}
	f\left(\mathbf{x}_{j}, y_{j} ; w^{\mathsf{LLM}}, \theta\right)=\mathcal{L}_{\text {task }}\left(s\left(\mathbf{x}_{j} ; w^{\mathsf{LLM}}, \theta\right), y_{j}\right)
\end{align}
where $s(\cdot)$ is the LLM's representation output. Collectively, the optimization objective of our interest is to minimize the global loss function across all participating clients, defined as:
\begin{align}
	\underset{ \theta}{\operatorname{argmin}}~~ \widetilde{F}( \theta)=\sum_{n=1}^{N}\frac{|\mathcal{D}_n|}{\mathcal{|D|}}F_n( \theta), \label{2}
\end{align}
where $\mathcal{|D|} = \sum_{n=1}^N|\mathcal{D}_n|$, and
\begin{align}
	F_n(\theta) = \frac{1}{|\mathcal{D}_n|}\sum_{(\mathbf{x}_{j}, y_{j}) \in \mathcal{D}_n}f(\mathbf{x}_{j}, y_{j};w^{\mathsf{LLM}}, \theta).
\end{align}

To tackle the optimization objective in \eqref{2}, we propose \textbf{ELSA}, a HFL framework tailored for distributed LLM fine-tuning under resource and privacy constraints. ELSA systematically addresses three core challenges: \textit{(i)} semantic heterogeneity across clients via behavior-aware clustering, \textit{(ii)} communication overhead and label leakage via tripartite split training with layered compression, and \textit{(iii)} robust global aggregation via trust-weighted fusion at the cloud level. An overview of ELSA is presented in Alg.~\ref{alg:elsa}, and the following subsections elaborate on each technical component in detail.

\subsection{Learning Architecture of ELSA}
In the following, we first introduce a client clustering strategy based on Kullback–Leibler divergence (KLD) to address data heterogeneity and unreliable trustworthiness and communications across clients. Then, we detail the end-to-end workflow of LLM partitioning and collaborative training across clients. Finally, we present a compression framework aimed at simultaneously enhancing communication efficiency and safeguarding the data privacy during the training.

\subsubsection{Behavior-aware Hierarchical Clustering and Robust Aggregation}

\textcolor{blue}{Client clustering in FL, including HFL, commonly relies on model similarity or label distributions (e.g., using KLD) to address heterogeneity. Model-similarity-based methods typically require parameters or gradients from full local models, making them impractical for resource-constrained edge devices fine-tuning LLMs. Label-based methods assume that similar class proportions imply compatible optimization landscapes, particularly in vision tasks with compact models, e.g., CNNs. Although effective in label-rich settings, label-based clustering is ill-suited to distributed LLM fine-tuning: \textit{(i)} downstream tasks (e.g., classification, extractive summarization, sequence labeling) lack a universal label space, limiting label-proportion-based clustering; and \textit{(ii)} heterogeneity extends beyond class frequencies to semantic and behavioral differences in how local LLMs interpret and respond to the same linguistic input\footnote{\textcolor{red}{A detailed comparison with representative methods, including CFL-GP \cite{CFLGP}, Fed-CM \cite{FedCM}, and HT-FL \cite{HTFL}, is provided in \textbf{Appendix~A}.}}.}

\textcolor{blue}{HFL deployments also face two often-overlooked challenges: \textit{unreliable clients} (e.g., due to data poisoning or severe label noise) and \textit{unstable or delayed connectivity}, particularly for geographically distant clients. Existing FL/HFL clustering and association metrics typically consider static statistical divergence or link-level proximity in isolation. Data-driven clustering methods \cite{KLFedDis,Wu2026Hierarchical} neglect client security and propagation latency, whereas communication-centric frameworks \cite{AdaptiveUAV,Fan2026Mobility} overlook client reliability and behavioral data quality.} To address these gaps, we propose a task-agnostic behavioral profiling framework that extends device similarity beyond label distributions by quantifying semantic behavioral divergence, e.g., differences in local models' predictive behavior on identical inputs. Our ELSA comprises three pillars:

\noindent $\bullet$ A probe-based behavioral fingerprint derived from public inputs, enabling KLD-driven clustering without assuming task type or label space of clients;

\noindent $\bullet$ A trustworthiness score grounded in prediction consistency to down-weight clients with polluted or anomalous data;

\noindent $\bullet$ A client-to-server assignment policy that respects physical network constraints between clients and edge servers.

Subsequently, we formalize our approach in four steps:

\noindent \textbf{Step 1: Construction of a Public Probe Set.} We construct a lightweight, public probe dataset $\mathcal{P} = \{x^{(1)},..., x^{(Q)}\}$ by sampling diverse inputs from open benchmarks (e.g., GLUE, TREC, or SQuAD development datasets\cite{SuperGLUE,SQuAD}). \textcolor{blue}{This set is distributed by the cloud server to all clients and serves as a common reference for behavioral evaluation, without exposing any private data\footnote{\textcolor{red}{To address concerns regarding probe distribution mismatch, privacy leakage, and parameter sensitivity, we conduct extensive empirical evaluations in \textbf{Appendix~B}, demonstrating that our fingerprints are highly robust to domain shifts, leakage-free, and comparable within shared tasks.}}.}

\noindent \textbf{Step 2: Task-Agnostic Behavioral Fingerprint Extraction.} For each client $n$, we feed every probe input $x^{(j)} \in \mathcal{P}$ into its locally fine-tuned LLM (e.g., BERT) and extract the contextual representation of the \texttt{[CLS]} token, a standard sentence-level embedding used across transformer-based LLMs, denoted by $\mathbf{T}_n^j \in \mathbb{R}^{\text{D}^{\mathsf{hidden}}}$, where $\text{D}^{\mathsf{hidden}}$ represents the hidden layer dimension. To obtain a unified-dimension, task-agnostic profile that is invariant to downstream task head structures, we model the client’s behavioral fingerprint as a multi-variate Gaussian distribution:
\begin{align}
	R_n = \widetilde{\mathcal{N}}(\boldsymbol{\mu}_n, \Sigma_n),
\end{align}
where $\boldsymbol{\mu}_n = \frac{1}{Q} \sum_{j=1}^Q \mathbf{T}_n^{(j)}$ represents the average semantic representation of the \texttt{[CLS]} vector for client $n$ across all probe samples and $\Sigma_n = \frac{1}{Q} \sum_{j=1}^Q (\mathbf{T}_n^{(j)} - \boldsymbol{\mu}_n)(\mathbf{T}_n^{(j)} - \boldsymbol{\mu}_n)^\top$ captures the dispersion of the \texttt{[CLS]} vectors of client $n$ around its mean.

\noindent \textbf{Step 3: Symmetric KLD for Behavioral Discrepancy.} We quantify the data heterogeneity between clients $n$ and $n'$ via the symmetrized KLD between their Gaussian fingerprints:
\begin{gather}
	\mathcal{R}(n,n') = \text{KL}(R_n \| R_n') + \text{KL}(R_n' \| R_n),
\end{gather}
where the closed-form KLD between multivariate Gaussians is given by
\begin{align}
	\text{KL}(\widetilde{\mathcal{N}}_n \| \widetilde{\mathcal{N}}_{n'}) &= \frac{1}{2} \bigg[ \text{tr}(\boldsymbol{\Sigma}_{n'}^{-1} \boldsymbol{\Sigma}_n) - \text{D}^\mathsf{hidden} + \ln \frac{|\boldsymbol{\Sigma}_{n'}|}{|\boldsymbol{\Sigma}_n|}\notag \\
	&+ (\boldsymbol{\mu}_{n'} - \boldsymbol{\mu}_n)^\top \boldsymbol{\Sigma}_{n'}^{-1} (\boldsymbol{\mu}_{n'} - \boldsymbol{\mu}_n)  \bigg].
\end{align}

This yields an $N \times N$ behavioral distance matrix that reflects semantic divergence beyond label statistics across clients.

\noindent \textbf{Step 4: Trust and Communication-Aware Client Clustering.} To ensure robustness against data pollution and practical deployability under network constraints, we integrate trustworthiness and communication feasibility into a unified client clustering framework. Specifically, we first compute a \textit{trustworthiness score} for each client \(n\) as follows:  
\[
w_n^{\mathsf{trust}} = \exp\left( -\underbrace{\frac{1}{Q} \sum_{j=1}^Q \frac{1}{\|\mathbf{T}^{(j)}_n\|_\text{2}}}_{\text{inverse confidence}} - \underbrace{\frac{1}{N-1} \sum_{n' \neq n} \mathcal{R}(n, n')}_{\text{mean behavioral divergence } \overline{\mathcal{R}}_n} \right),
\]  
where $\frac{1}{\|\mathbf{T}^{(j)}_n\|_\text{2}}$ is the inverse confidence of its \texttt{[CLS]} embeddings across the public probe set $\mathcal{P}$. Intuitively, a client with low prediction entropy (high confidence) and low divergence from peers (consistent behavior) receives a high trust score. We then define the \textit{communication-feasible edge server set} for each client \(n\) as:  
\[
\mathcal{E}_n = \{ k\mid \tau_{n,k} \leq \tau_{\text{max}}, k \in \mathcal{K}\},
\]  
where \(\tau_{n,k}\) is the round-trip latency between client \(n\) and edge server \(k\), and \(\tau_{\text{max}}\) is a system-defined threshold (e.g., 200 ms). We finally construct a \textit{communication-constrained client partition} through a four-stage process.

\noindent \textbf{(Stage 1)} For each edge server \(k\), we define its candidate client set as \(\mathcal{C}_k = \{ n \mid k \in \mathcal{E}_n \}\).

\noindent \textbf{(Stage 2)} We perform spectral clustering \textit{independently within each} \(\mathcal{C}_k\), using a trust-weighted affinity matrix \(\mathbf{A}^{(k)}\) with entries:
\[
\mathbf{A}^{(k)}_{n,n'} = w_n^{\mathsf{trust}} \cdot w_{n'}^{\mathsf{trust}} \cdot \exp\left( -\gamma \cdot \mathcal{R}(n, n') \right), \quad \forall n, n' \in \mathcal{C}_k,
\]
where \(\gamma > 0\) controls the sensitivity to behavioral divergence.

\noindent \textbf{(Stage 3)} For each resulting cluster \(\mathbb{N}_k \subseteq \mathcal{C}_k\), we compute its average trust score \(\bar{w}_k^\mathsf{trust} = \frac{1}{|\mathbb{N}_k|} \sum_{n \in \mathbb{N}_k} w_n^{\mathsf{trust}}\).

\noindent \textbf{(Stage 4)} If~\(\overline{w}_k < w_{\min}\) (a trust threshold), we either: \textit{(i)} merge \(\mathbb{N}_k\) into the nearest high-trust cluster (measured by centroid KLD), or \textit{(ii) }escalate it to cloud-level aggregation if no suitable merge target exists.

The final grouping \(\{ \mathbb{N}_k \}_{k=1}^K\) ensures that each cluster:  
\textit{(i)} consists of behaviorally similar clients,  
\textit{(ii)} excludes or mitigates the influence of polluted data, and  
\textit{(iii)} is assigned to an edge server with reliable connectivity, thereby enabling stable, efficient, and deployable hierarchical model aggregation.
\begin{figure}[htbp]
	\centering
	\scriptsize  
	\subfigure[]{
		
		\includegraphics[trim=0.0cm 0cm 0.0cm 0cm, clip, width=0.46\columnwidth]{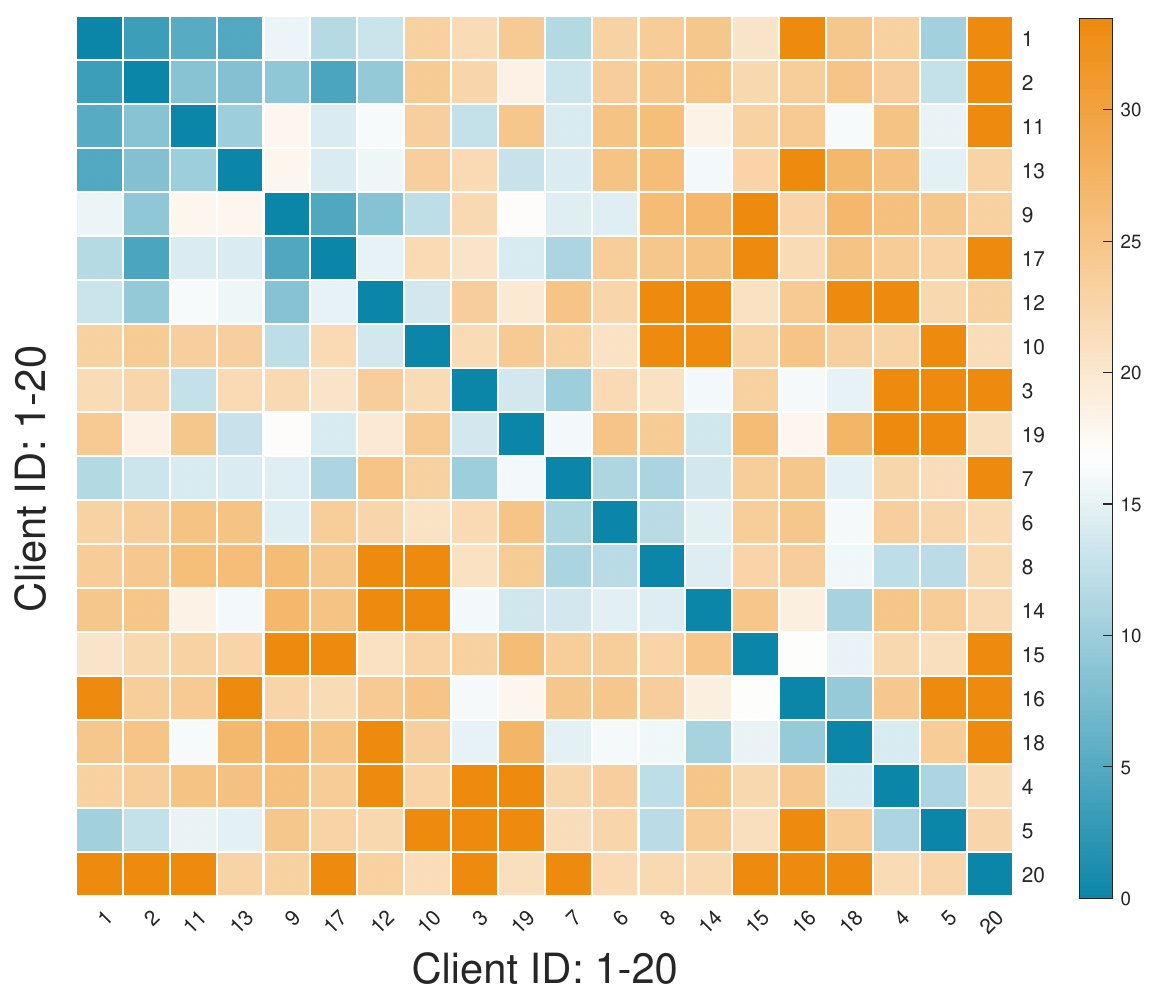}
	}
	\subfigure[]{
		\includegraphics[trim=0.0cm 0.0cm 0.0cm 0.0cm, clip, width=0.48\columnwidth]{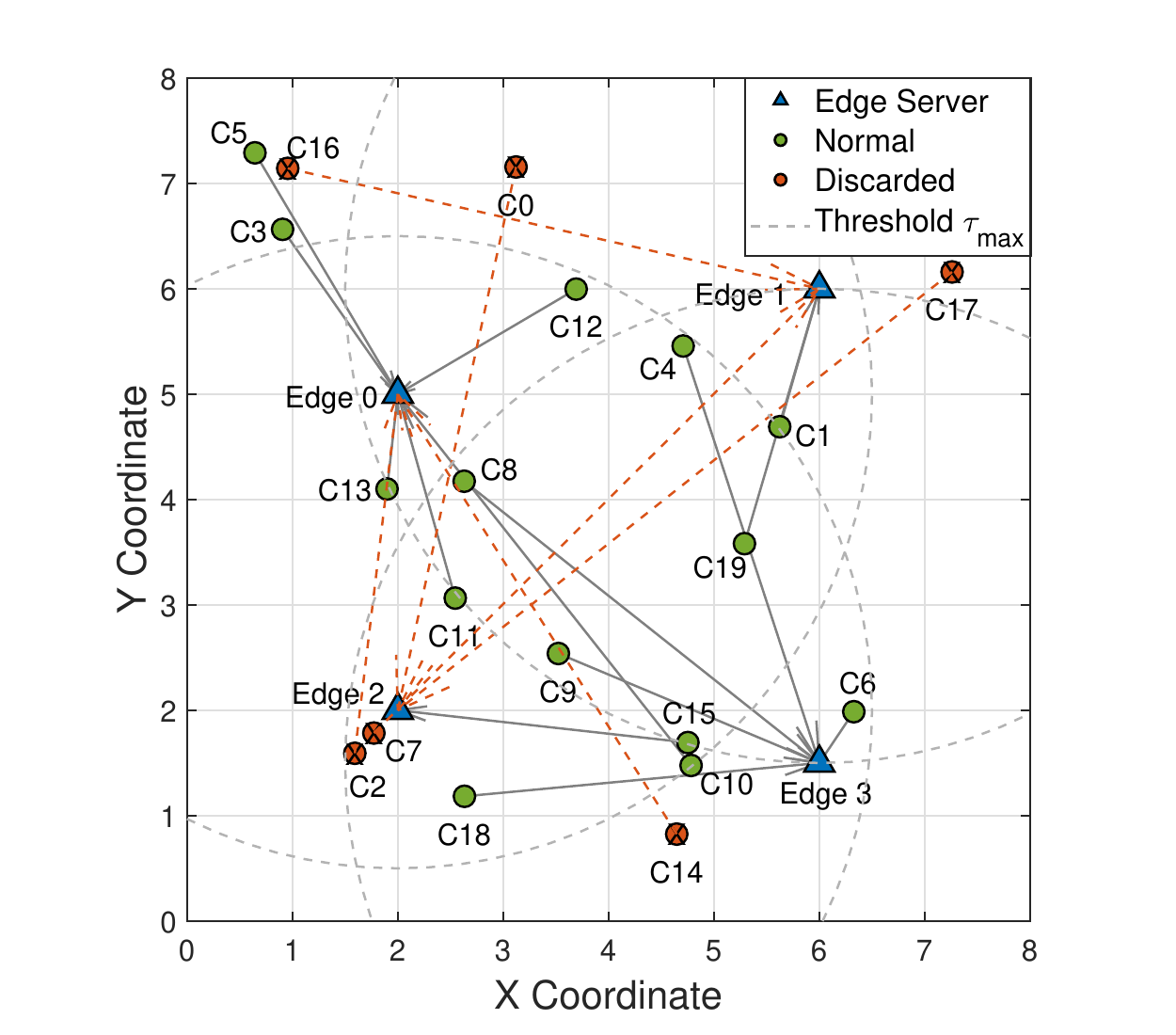}
	}
	\vspace{-3mm}
	\caption{Client behavioral heterogeneity and clustering outcome in a 20-client network. Left: $20 \times 20$ pairwise KLD matrix $\mathcal{R}(n,n')$ across all clients. 
		Right: final client-server association mapping.}
	\label{fig_2}
	\vspace{-3mm}
\end{figure}

To facilitate understanding, in Fig. 2, we consider a network with 20 clients, where each client's local dataset is constructed by partitioning the SQuAD dataset using Dirichlet distribution with concentration parameter $\hat{\alpha}$ = 0.1. Fig. 2(a) provides a heatmap of pairwise KLD ($\mathcal{R}(n, n')$) to visualize data distribution variations, while Fig. 2(b) illustrates the final client clustering  (i.e., client-to-server association) results. Across all edge server groups, clients marked with ‘X’ are either out-of-range or deemed untrustworthy; only high-trust, in-range clients participate in edge-level aggregation, specifically those shown as green circles labeled ‘Normal’.

\subsubsection{Model Splitting and Collaborative Training}
The tripartite model segmentation in ELSA, where both the initial (Part 1) and final (Part 3) segments reside on the client, while the intermediate transformer blocks (Part 2) are offloaded to the server, is deliberately designed to address two critical challenges in edge-based LLM fine-tuning: \textit{privacy preservation} and \textit{label confidentiality}. This design follows a principled insight for LLMs: by retaining the output layer (e.g., task-specific head) on the client side, the ground-truth labels never leave the local device, thereby eliminating the risk of label leakage during training. Moreover, keeping both input embedding and output layers locally prevents the server from directly accessing raw embedding gradients or logits, which are known to be highly susceptible to reconstruction attacks. Consequently, this symmetric client-side placement ensures that sensitive user data, including input tokens, labels, and final predictions, remain strictly within the trusted execution environment of the client, while only intermediate hidden states (obfuscated via SS-OP and sketching, detailed in later) are exchanged with the server.

Building upon this privacy-aware architecture, we integrate SL into our HFL framework, as illustrated in Fig.~1 and Fig.~3. Specifically, in Fig. 3, we consider the LLM architecture (e.g., BERT) as $\mathcal{M}$ transformer blocks stacked in series, and unlike static partitioning, we employ a \textit{resource-aware dynamic splitting strategy} to determine the number of layers for each segment adaptively for each client $n$. The first segment comprises an embedding layer and $p_n$ transformer blocks (Part 1), each of which includes multi-head attention modules, feed-forward networks (FFNs), and related components (e.g., residual connections and layer normalization). The second segment consists of $q_n$ transformer blocks (Part 2), and the third segment contains $o_n$ transformer blocks along with the output layer (Part 3). The first and third segments are deployed on the client-side, while the second segment can be placed on the corresponding server-side. Specifically, each attention module within Parts 1–3 is augmented with trainable low-rank matrices $\theta = [\theta_1, \theta_2, \theta_3]$ (i.e., $\theta = \{ (A^m_i, B^m_i) \mid i \in \{1,2,3\}, m \in \{1,\dots, \mathcal{M}\} \}$) forming the adapters, where $ \theta \in \mathbb{R}^{\text{M}}(\text{M} \ll \text{D})$, and the output layer in Part 3 is also trainable but contributes negligibly to total parameter count. 

To balance computational load and communication costs, the offloading ratio is determined by the client's available computational capacity $\mathcal{H}_n$ (e.g., FLOPS) and uplink bandwidth $\mathcal{B}_n$. 
We define the \textit{offloading preference score} $\mathcal{G}_n$ for client $n$ as:
\begin{align}
	\mathcal{G}_n = \lambda_1 \cdot \left(1 - \frac{\mathcal{H}_n}{\mathcal{H}_{\max}}\right) + \lambda_2 \cdot \frac{\mathcal{B}_n}{\mathcal{B}_{\max}},
	\label{eq:split_score}
\end{align}
where $\mathcal{H}_{\max}$ and $\mathcal{B}_{\max}$ (in bytes/second) are the maximum observed capacity and bandwidth in the system, and $\lambda_1, \lambda_2$ are weighting factors ($\lambda_1+\lambda_2=1$). 

Unlike symmetric partitioning, we fix the depth of the output segment to $o_n = o_{\text{fix}}$ (e.g., $o_{\text{fix}}=2$) for all clients. This design ensures that the task-specific head remains strictly local for label privacy and maintains a consistent model structure across the federation. 
Consequently, the dynamic adjustment focuses on balancing the local encoder depth $p_n$ and the offloaded encoder depth $q_n$. 
To prevent the local feature extractor from becoming overly specialized to non-IID data (a phenomenon known as over-personalization), we constrain the local encoder layers within a safe range $[p_{\min}, p_{\max}]$. 
The number of offloaded layers $q_n$ is then dynamically calculated as:
\begin{align}
	q_n = \mathcal{M} - o_{\text{fix}} - p_n,
\end{align}
where the local encoder depth $p_n$ is determined by mapping the preference score $\mathcal{G}_n$ to the valid range:
\begin{align}
	p_n = p_{\max} - \left\lfloor \mathcal{G}_n \cdot (p_{\max} - p_{\min}) \right\rfloor.
	\label{eq:split_p_dynamic}
\end{align}
Here, $p_{\min}$ (e.g., 1) guarantees basic input embedding privacy, while $p_{\max}$ serves as a theoretical upper bound to safeguard global model generalization. This dynamic mechanism ensures that clients with limited computational resources or high-quality connections offload more work to the edge, while powerful clients or those with poor connectivity perform more local computation. In essence, the rationale for this partitioning is rooted in the client's limited computational resources, the requirement for decision-making autonomy, and the necessity for end-to-end privacy control.

\begin{figure}[htbp]
	\centering
	\subfigure{
		\includegraphics[trim=0cm 0cm 0cm 0cm, clip, width=0.98\columnwidth]{simulate_results/main_framework0306_part2.pdf}
	}
	\vspace{-5mm}
	\caption{Breakdown of ELSA: client trains Part 1, sends compressed activations to edge for Part 2, which then feeds into Part 3 on client. Gradients flow backward symmetrically ($\nabla \widetilde{\mathbf{H}}_{n}^{\mathsf{down}} \xrightarrow{} \text{Edge}$, $\nabla \widetilde{\mathbf{H}}_{n}^{\mathsf{up}} \xrightarrow{} \text{Client}$), completing one round.}
	\vspace{-3mm}
	\label{fig_9}
\end{figure}

Each client $n \in \mathbb{N}_k$ within the coverage edge server $k$, feed-forwards its data through Part 1 of its model, generating a hidden layer state, defined as
\begin{equation}
	\hspace{-1mm}
	\resizebox{0.39\textwidth}{!}{$
		\mathbf{H}_{n}^{\mathsf{up}} {=}\hspace{-.6mm} \left.\left[h_{n,1}^{\mathsf{up}}, \ldots,h_{n,d}^{\mathsf{up}}, \ldots, h_{n,\text{D}}^{\mathsf{up}}\right]\hspace{-.6mm} {\in} \mathbb{R}^{\text{D}}\right|_{f_{\text{embed}}(x; \theta_1) \rightarrow f_{1 \sim p_n}}\hspace{-1mm},
		$}
	\hspace{-1mm}
\end{equation}
where $h_{n,d}^{\mathsf{up}}$ is the $d$-th component of $\mathbf{H}_{n}^{\mathsf{up}}$, $f_{\text{embed}}(x)$ denotes the embedding layer of the LLM, and $f_{1\sim p_n}$ represents the sequential feed-forward through transformer block 1 to block $p_n$ at the client. The client then aims to send this hidden layer state to the edge server to continue the process of model fine-tuning. \textcolor{blue}{However, since the direct transmission of the hidden layer state $\mathbf{H}_{n}^{\mathsf{up}}$ can lead to privacy leakage (e.g., through model inversion attacks) and incur excessive communication overhead, the client first compresses its hidden layer state using a sketch-based method (Section B.3) and then uploads both compressed representation and the attention mask matrix $\widetilde{\mathbf{M}}_n$ to the ES.} Once the information is received, the edge server will decompress it to obtain an approximate information $\widetilde{\mathbf{H}}_{n}^{\mathsf{up}}$ of $\mathbf{H}_{n}^{\mathsf{up}}$. \textcolor{blue}{Then, the edge server uses $\widetilde{\mathbf{H}}_{n}^{\mathsf{up}}$ and $\widetilde{\mathbf{M}}_n$ as inputs to perform the computationally intensive operations, yielding the hidden state $\mathbf{H}_{n}^{\mathsf{down}}$, defined as}
\begin{equation}
	\hspace{-4.5mm}
	\resizebox{0.44\textwidth}{!}{$
		\mathbf{H}_{n}^{\mathsf{down}} {=}\hspace{-.7mm} \left.\left[h_{n,1}^{\mathsf{down}}, \ldots,h_{n,d}^{\mathsf{down}}, \ldots, h_{n,\text{D}}^{\mathsf{down}}\right]\hspace{-.7mm} {\in} \mathbb{R}^{\text{D}}\right|_{f_{p_n + 1\sim p_n + q_n}(\widetilde{\mathbf{H}}_{n}^{\mathsf{up}}, \widetilde{\mathbf{M}}_n; \theta_2)}\hspace{-.5mm},
		$}\hspace{-3.15mm}
\end{equation}
where $h_{n,d}^{\mathsf{down}}$ is the $d$-th component of $\mathbf{H}_{n}^{\mathsf{down}}$ and $f_{p + 1\sim p + q}(.)$ denotes sequential feed-forward through transformer blocks $p_n$ to $p_n+q_n$. Then, $\mathbf{H}_{n}^{\mathsf{down}}$ will be compressed into $\widetilde{\mathbf{H}}_{n}^{\mathsf{down}}$ and then sent to the client.

Finally, the latest $o_n$ model segments on the client-side perform computations to obtain the logits $\mathcal{Z}$ and the loss value, which triggers gradient backpropagation; mathematically,
\begin{align}
	\mathcal{Z} &= f_{p_n + q_n +1 \sim \mathcal{M}}(\widetilde{\mathbf{H}}_{n}^{\mathsf{down}}, \widetilde{\mathbf{M}}_n; \theta_3), \\
	\mathbb{L} &= \mathcal{F}_{\text{loss}}(f_{\text{out}}(\mathcal{Z},y)),
\end{align}
where $ f_{p_n + q_n +1 \sim \mathcal{M}}(.)$ denotes sequential feed-forward through transformer blocks $p_n+q_n$ to $p_n+q_n+o_n$, $f_{\text{out}}(.)$ is the output layer of the model, $y$ is the targeting sentences and $\mathcal{F}_{\text{loss}}(.)$ denotes the loss function.

After computing the loss, each client performs gradient backpropagation locally to update Blocks $o$, then transmits the gradient of hidden states $\widetilde{\mathbf{H}}_{n}^{\mathsf{down}}$ back to the server (i.e., $\nabla \widetilde{\mathbf{H}}_{n}^{\mathsf{down}}$). The server executes gradient backpropagation on Blocks $q$, updates its parameters, and subsequently transmits the gradients of hidden states $\widetilde{\mathbf{H}}_{n}^{\mathsf{up}}$ back to the client. Lastly, Blocks $p$ perform gradient backpropagation, which completes a training round.

While client–edge collaboration improves training efficiency, the restricted representational power of isolated local updates can compromise global model expressiveness. To mitigate this, we periodically aggregate model parameters at the cloud: after every fixed number of collaborative training rounds (i.e., $\mathbf{t}$), each edge server $k$ uploads its locally refined adapter parameters to the cloud server for global model aggregation. Consistent with our behavior-aware clustering and robustness-oriented design, the contribution of each edge server to the global aggregation is weighted according to two factors: \textit{(i)} the internal behavioral coherence of its client group, and \textit{(ii)} the aggregated trustworthiness of the participating clients. Specifically, we define:
\begin{align}
	\alpha_k = \frac{1}{1 + \overline{\mathcal{R}}}_k \times \bar{w}_k^{\text{trust}},
\end{align}
where $\overline{\mathcal{R}}_k = \frac{1}{|\mathbb{N}_k|(|\mathbb{N}_k|-1)} \sum_{n \neq n' \in \mathbb{N}_k} \mathcal{R}(n, n')$ is the average pairwise symmetric KLD among clients in group $\mathbb{N}_k$. Next, we obtain the normalized weights
$\widetilde{\alpha}_k = \frac{\alpha_k}{\sum_{k=1}^K \alpha_k}$, using which we obtain the global model $\theta_g$, as follows:
\begin{align}
	\theta_g &= \sum_{k=1}^K \widetilde{\alpha}_k\theta_{g,k},
\end{align}
where $\theta_{g,k}$ represents adapter parameters obtained through collaborative training between edge server $k$ and clients within its coverage area during the $g^\text{th}$ global iteration. Further, we consider that the training ends at global aggregation once two consecutive global models satisfy the following condition:
\begin{align}
	\Vert \theta_{g} - \theta_{g-1}\Vert_\text{2} \leq \xi,
	\label{13}
\end{align}
where $\xi$ is a small positive value used to control the convergence criterion.
\begin{algorithm}[t]
	{
		\small
		\caption{Overview of ELSA: Efficient LLM-centric Split Aggregation}
		\label{alg:elsa}
		\SetKwInOut{Input}{Input}\SetKwInOut{Output}{Output}
		\Input{Client set $\mathcal{N}$, Edge server set $\mathcal{K}$, Public probe set $\mathcal{P}$, Pre-trained LLM $w^{\mathsf{LLM}}$, Local rounds $\mathbf{t}$.}
		\Output{Global optimized parameters $\theta_g$.}
		{\bf{Initialization:}} $w_n^{\mathsf{trust}}=0$, $\mathcal{E}_n = \emptyset$.  
		
		\textbf{Phase 1: Behavior-aware Client Clustering}
		
		\textit{1.1 Behavioral Fingerprinting:} Each client $n$ computes Gaussian fingerprint $R_n = \widetilde{\mathcal{N}}(\boldsymbol{\mu}_n, \Sigma_n)$ using probe set $\mathcal{P}$\;
		
		\textit{1.2 Discrepancy Measurement:} Compute symmetric KL divergence matrix $\mathcal{R}(n, n')$ between all client pairs\;
		
		\textit{1.3 Trust \& Assignment:} Calculate trust score $w_n^{\mathsf{trust}}$ and communication feasibility $\mathcal{E}_n$. Perform spectral clustering within feasible edge groups to obtain clusters $\{\mathbb{N}_k\}_{k=1}^K$\;
		
		\textbf{Phase 2: Collaborative Training \& Aggregation}
		
		\If{The global model parameter $\theta_{g}$ changes does not satisfy (\ref{13})}{
			\ForEach{edge server $k \in \mathcal{K}$}{
				\For{$t = 1$ \KwTo $\mathbf{t}$}{
					\ForEach{client $n \in \mathbb{N}_k$}{
						\textit{Forward Pass (Split):} Client computes Part 1 output $\mathbf{H}_{n}^{\mathsf{up}}$\;
						\textit{Compression \& Privacy:} Apply SS-OP and sketching to generate $\widetilde{\mathbf{H}}_{n}^{\mathsf{up}}$. Upload to Edge $k$\;
						\textit{Edge Processing:} Edge computes Part 2 output $\mathbf{H}_{n}^{\mathsf{down}}$ using $\widetilde{\mathbf{H}}_{n}^{\mathsf{up}}$. Compress and send back $\widetilde{\mathbf{H}}_{n}^{\mathsf{down}}$\;
						\textit{Client Completion:} Client computes Part 3 output, calculates loss $\mathbb{L}$, and initiates backpropagation\;
						\textit{Backward Pass:} Gradients flow symmetrically: Client (Part 3) $\to$ Edge (Part 2) $\to$ Client (Part 1)\;
					}
				}
				Edge $k$ aggregates local adapter parameters to obtain $\theta_{g,k}$\;
			}
			
			\textbf{Phase 3: Cloud-level Global Aggregation}
			
			Compute edge weights $\alpha_k = \frac{1}{1 + \overline{\mathcal{R}}_k} \times \bar{w}_k^{\text{trust}}$\;
			Normalize weights $\widetilde{\alpha}_k$ and update global model: $\theta_g = \sum_{k=1}^K \widetilde{\alpha}_k \theta_{g,k}$\;
		}
		\Return{$\theta_g$}\;
	}
\end{algorithm}


\newtheorem{remark}{Remark}
\begin{remark}[\textcolor{red}{Applicability to Encoder-Only and Decoder-Only LLMs}]
\textcolor{red}{Although described using encoder-only notation, ELSA's behavioral clustering and Transformer-block partitioning also apply to decoder-only LLMs. Decoder blocks can be partitioned along the client--edge--client pipeline, with the input embedding, final causal blocks, and LM head retained on the client. The hidden state of the EOS token or last valid prompt token can replace the \texttt{[CLS]} representation as the behavioral fingerprint. Appendix~C details this extension, and our experiments evaluate both architectures.}
\end{remark}

\subsubsection{Layered Compression for Secure and Efficient Communications}
To simultaneously reduce communication overhead and mitigate the privacy leakage during the information exchange between clients and edge servers, we design a layered compression-and-obfuscation mechanism based on computational sketches, which is tailored to preserve semantic fidelity in NLP model updates. \textcolor{blue}{This separates local feature processing into two stages: semantics-guided privacy obfuscation followed by sketch-based transmission compression. For obfuscation, we propose \textit{semantic subspace orthogonal perturbation} (SS-OP), in which each client $n$ targets the dominant semantic subspace of its local hidden activations rather than applying a global orthogonal transform. Specifically, let $\mathbf{J}_n \in
\mathbb{R}^{Q \times D^{\mathsf{hidden}}}$ denote the reference
hidden-state matrix constructed from $Q$ hidden states (e.g.,
\texttt{[CLS]} vectors) obtained using a reference batch of public probe inputs or previously sampled local training inputs. This reference batch is disjoint from the current protected mini-batch.} Client $n$ computes the top-$r$ principal components via truncated singular value decomposition (or power iteration):
\begin{align}
	\mathbf{U}_n = \operatorname{Top}_r\big(\text{SVD}(\mathbf{J}_n)\big) \in \mathbb{R}^{\text{D}^{\mathsf{hidden}} \times r}, \label{17}
\end{align}
where $\mathbf{U}_n^\top \mathbf{U}_n = \mathbf{I}_r$ spans the most informative semantic directions. Next, a client-specific random orthogonal matrix $\mathbf{V}_n \in \mathbb{R}^{r \times r}$ is generated by QR decomposition of a pseudo-random Gaussian matrix seeded from the client ID and a pre-shared salt:
\begin{align}
	\mathbf{V}_n = \text{QR}\big( \Phi(n) \big),
	\label{18}
\end{align}
where $[\Phi(n)]_{i,j} \sim \widetilde{\mathcal{N}}(0,1)$ with $\text{seed}_n = \text{Hash}(s \parallel n)$, and $s$ is a pre-shared salt value. \textcolor{blue}{The full perturbation matrix is then constructed as\footnote{\textcolor{red}{A comprehensive cryptographic formalization regarding master salt management, cloud-edge trust boundary isolation, and the computational hardness of protecting $U_n$ is provided in \textbf{Appendix~D}.}}}:
\begin{align}
	\mathcal{Q}_n = \mathbf{U}_n \mathbf{V}_n \mathbf{U}_n^\top + (\mathbf{I}_{\text{D}^{\mathsf{hidden}}} - \mathbf{U}_n \mathbf{U}_n^\top).
\end{align}

This matrix is orthogonal ($\mathcal{Q}_n^\top \mathcal{Q}_n = \mathbf{I}_{\text{D}^{\mathsf{hidden}}}$) and performs a random rotation \textit{only within the semantic subspace}, leaving the orthogonal complement unchanged. This design ensures that:

\noindent \textbf{(1)} \textbf{Privacy is strengthened}: Adversaries cannot recover the original semantics without knowing both $\mathbf{U}_n$ (data-dependent) and $\mathbf{V}_n$ (secret-seeded).

\noindent \textbf{(2)} \textbf{Training remains stable}: During backpropagation, the client applies $\mathcal{Q}_n^\top$ to restore exact gradients, as $\mathcal{Q}_n$ is orthogonal.

\noindent \textbf{(3)} \textbf{Activation structure is preserved}: Zero or low-magnitude dimensions outside the semantic subspace remain unaltered, maintaining compatibility with downstream layers (e.g., attention heads or classification heads).

\textcolor{blue}{Having obfuscated the local feature privacy via SS-OP, the client enters the second stage: lightweight activation compression. To substantially minimize transmission overhead without requiring full-model parameter updates on edge clients (in contrast to standard federated sketching paradigms detailed in \textbf{Appendix~E}), ELSA directly applies computational sketching to the perturbed activations. Specifically, the uploaded hidden state is then computed as $\widetilde{\mathbf{H}}_n = \mathcal{Q}_n \mathbf{H}_n$, which is compressed into a $Y \times Z$ sketch matrix $U$ using $Y$ independent hash functions, where $Z \ll \text{D}$. Specifically, each sketch element is calculated as:}
\begin{align}
	U_{n,[j,k]} = \sum_{d:h_{n,j}(d)=u}\text{sign}_j(d)\cdot h_{n,d}^{\mathsf{up}}, \label{20}
\end{align}
where $u\in \{1, 2,..., Z\}$ represents the column index of sketch matrix $U$ and ${\text{sign}_{j: 1 \leq j \leq Y}: \text{D} \rightarrow \{+1, -1\}}$ denotes $Y$ pairwise independent sign hash functions. \textcolor{blue}{Besides, we define the compression ratio as $\rho = D / (Y Z)$, which directly reduces to $\rho = d / d'$ when sketching is applied along the feature dimension, where $d$ and $d'$ represent the raw and compressed feature dimensions, respectively.}

At the edge, we recover the original hidden layer information based on the received $U_{n,[j,k]}$ as follows:
\begin{align}
	\widetilde{\mathbf{H}}_{n}^{\mathsf{up}} = \text{Median}_{1 \leq j \leq Y}\{\text{sign}_j(d)\cdot 	U_{n,[j,h_{n,j}(d)]}\},
\end{align}
where $\text{Median}(.)$ denotes the median of all approximated values. Edge servers aggregate $\{\widetilde{\mathbf{H}}^{\mathsf{up}}_n\}$ directly without decryption, as the orthogonal transformation preserves inner products and norms, ensuring meaningful model averaging. Besides, during the downloading (or distribution) of the hidden layer state, we compress and encrypt it to obtain $\widetilde{\mathbf{H}}_{n}^{\mathsf{down}}$.


\begin{lemma}[\textcolor{red}{Reconstruction Error Lower Bound under SS-OP and Sketching}]
\label{lem:reconstruction_lower_bound}
    \textcolor{red}{Let $H_n=[h_{n,1},\ldots,h_{n,T}]\in\mathbb{R}^{d\times T}$ denote the intermediate activations of client $n$, where the activation vectors are independently drawn from an isotropic Gaussian model $h_{n,t}\sim\mathcal{N}(0,\sigma_H^2\mathbf{I}_d),t=1,\ldots,T.$ Accordingly, $ \Sigma_H = \frac{1}{T}\mathbb{E}[H_nH_n^\top] = \sigma_H^2\mathbf{I}_d.$ For the protected activation batch, let the SS-OP matrix be $
\mathcal{Q}_n = \mathbf{U}_n\mathbf{V}_n\mathbf{U}_n^\top + \left( \mathbf{I}_d-\mathbf{U}_n\mathbf{U}_n^\top \right),$ where $\mathbf{U}_n^\top\mathbf{U}_n=\mathbf{I}_r$ and $\mathbf{V}_n^\top\mathbf{V}_n=\mathbf{I}_r$. Conditioned on the current model parameters, the matrix $\mathcal{Q}_n$
is generated before processing the protected activation batch using a disjoint public-probe or reference batch, independently of the protected activations, and remains fixed throughout the corresponding forward and backward passes. Let $\mathbf{R}\in\mathbb{R}^{d'\times d}$ be the sketching matrix,
where $d'=d/\rho$, $\rho>1$, and $\operatorname{rank}(\mathbf{R})\leq d'$. Given the observed sketch
$\mathbf{S}_n=\mathbf{R}\mathcal{Q}_nH_n,$ any reconstruction adversary that does not possess the secret
SS-OP matrix satisfies \[ \mathbb{E} \left[ \left\| H_n-\widehat{H}_n \right\|_F^2 \right] \geq T\left(1-\frac{1}{\rho}\right) \operatorname{Tr}(\Sigma_H) >0. \] }
\end{lemma}
\textcolor{red}{A comprehensive and rigorous proof of Lemma~\ref{lem:reconstruction_lower_bound} is provided in \textbf{Appendix F}.}

\subsubsection{Communication Modeling}
\label{sec:comm_model}
ELSA employs symmetric transmission: intermediate activations are sent during the forward pass, and their corresponding gradients are returned during the backward pass. Since both have identical dimensions, the total communication overhead in global iteration $g$ is
\begin{equation}
	C_g = 
	\underbrace{
		\frac{2\mathbf{t} \zeta \mu D^\mathsf{hidden}}{\rho}\sum_{k=1}^{K} \sum_{n \in \mathbb{N}_k} B_n
	}_{\text{Client $\leftrightarrow$ Edge }} + \underbrace{ K \cdot |\theta^{\text{LoRA}}|}_{\text{Edge $\rightarrow$ Cloud}},
	\label{eq:comm_cost_corrected}
\end{equation}
where $\mathbf{t}$ is the number of client–edge collaborative fine-tuning rounds performed within each global aggregation, $\zeta$ denotes the byte size per floating-point parameter (e.g., $\zeta = 4$ for FP32), $\mu$ represents the input sequence length in tokens, and $|\theta^{\text{LoRA}}|$ is the total communication volume (in bytes) of the aggregated LoRA adapter parameters uploaded by a single edge server to the cloud. Here, $B_n$ is the average local mini-batch size used by client $n$ during training.

The per-client communication latency in the $g^\text{th}$ global round is determined by its local batch size and effective network throughput. Specifically, during $\mathbf{t}$ rounds of client--edge collaborative training, client $n$ transmits and receives compressed intermediate activations for $B_n$ samples per round. Accounting for symmetric forward/backward passes, the total transmitted volume is $2 \mathbf{t} B_n \mu \zeta D^{\mathsf{hidden}} / \rho$. Dividing by the bandwidth $\mathcal{B}_{\text{n}}$ of client $n$, the communication time is:
\begin{equation}
	\mathcal{T}_{g, n} = 
	\frac{
		2 \mathbf{t} B_n \mu \zeta D^{\mathsf{hidden}} / \rho
	}{
		\mathcal{B}_{\text{n}}
	},
	\label{eq:client_comm_time_corrected}
\end{equation}

Subsequently, the total communication time until convergence is given by
\begin{equation}
	T_{\text{total}} \approx G \times \max_{n \in \mathbb{N}_k,\, k \in \mathcal{K}}(\mathcal{T}_{g,n}),
	\label{eq:total_comm_time}
\end{equation} 
where $G$ is the number of global iterations required to satisfy the convergence criterion in \eqref{13}.

\subsection{Convergence Analysis}
\begin{theorem}[Convergence of ELSA]
	Under Assumptions 1–3 (in the \textbf{Appendix G}), let the learning rate be $\eta = \frac{1}{\mathcal{L}\sqrt{G}}$, where $G$ is the total number of global iterations. Then, the sequence of global adapter parameters $\{\theta_g\}_{g=0}^{G-1}$ generated by ELSA satisfies:
	\begin{equation}
		\hspace{-4.5mm}
		\resizebox{0.47\textwidth}{!}{$
			\frac{1}{G} \sum_{g=0}^{G-1} \mathbb{E}\left[ \left\| \nabla \widetilde{F}(\theta_g) \right\|^2 \right] 
			\leq \underbrace{\frac{4\mathcal{L}(\widetilde{F}(\theta_0) - \widetilde{F}^*)}{\sqrt{G}}}_{\text{optimization error}} 
			\nonumber \\+ \underbrace{\frac{\sigma_{\text{local}}^2}{\sqrt{G}}}_{\text{vanishing noise}} 
			+ \underbrace{\sigma_2^2}_{\text{non-IID bias}},
			$}\hspace{-3.15mm}
	\end{equation}
	where $\widetilde{F}^* = \min_\theta \widetilde{F}(\theta)$ denotes the optimal value of the global objective (proof is provided in the \textbf{Appendix G}).
\end{theorem}

\begin{remark}
	\textcolor{blue}{Theorem~1 shows ELSA converges to an $\mathcal{O}(\sigma_2^2)$-neighborhood of a stationary point at a rate of $\mathcal{O}(1/\sqrt{G})$. Although $\mathcal{O}(1/G)$ non-IID bias reduction is possible under idealized settings (e.g., single-step updates and unbiased aggregation), our multi-round client-edge collaboration, model splitting, and compression introduce a persistent bias between the expected update and the true gradient.} This results in a non-vanishing residual error $\sigma_2^2$. In contrast, sketching and SGD noise ($\sigma_{\text{local}}^2$) vanish during training. Convergence is ensured by bounding intra-cluster semantic heterogeneity via KLD-based clustering, which aligns local SS-OP bases $\{\mathbf{U}_n\}$ and controls aggregation bias.
\end{remark}

\section{Experiments and Evaluations}

In the following, we aim to evaluate the proposed ELSA framework with respect to two key research questions:

\noindent $\bullet$ How does ELSA perform compared to existing methods?

\noindent $\bullet$ How do various hyperparameters affect ELSA's performance, and how can their optimal values be determined?

\subsection{Experimental Setup}
\noindent \textbf{Datasets:} We consider eight diverse datasets spanning various NLP tasks: \textit{(i)} \textit{Text Classification (TC)}: \textit{Trec}, \textit{AG\_News}, \textit{Emotion}, and \textit{Banking77}, covering question, news, sentiment, and fine-grained intent classification (up to 77 categories)\cite{trec,AGNews,Bank}; \textit{(ii)} \textit{Natural Language Inference (NLI)}: \textit{RTE} and \textit{CB}, which involve premise-hypothesis entailment; \textit{MultiRC} and \textit{SQuAD}, representing multi-sentence reading comprehension and extractive question answering tasks respectively \cite{SuperGLUE,SQuAD}. 

\noindent \textbf{Baselines:} \textcolor{blue}{We evaluate ELSA against 11 representative baselines, covering three categories: \textit{(i)} general FL methods, including FedAvg \cite{FedAvg} and the performance of FedAvg under randomly selected client conditions (i.e., FedAvg (Ran.)), where clients are randomly selected in each round; \textit{(ii)} aggregation-optimized and edge FL/HFL methods, including FedProx \cite{FedProx}, FedAMS \cite{FedAms}, RaSA \cite{RaSA}, and FedCAda \cite{FedCAda}; and \textit{(iii)} SL for LLMs \& SL FL/HFL methods, including RoFed \cite{ROFED}, SFLAM \cite{qiangdeploying}, SplitLLM \cite{SplitLLM}, Privacy-Aware SFL \cite{Privacy-Awa}, and HSFL \cite{HSFL}.}

\noindent \textbf{Parameter settings:} \textcolor{blue}{Following the semi-simulation paradigm in prior work \cite{FwdLLM, SplitCom}, we construct a system comprising $N= $20 logical clients and $K=$ 4 logical ESs. First, the locations of clients and ESs are configured within an 8~km $\times$ 8~km region, where their spatial distances are used to model client--ES link conditions, transmission latency, available bandwidth, and association/scheduling decisions. The end-to-end federated training loop, including local client updates, edge aggregation, and global aggregation, is executed serially on a workstation equipped with 4$\times$ NVIDIA RTX 4090 GPUs to emulate system-level behavior under multiple clients and edge servers, as illustrated in Fig.~\ref{fig_2}. Second, to ground our computation and communication analyses on representative hardware, we empirically measure the peak memory footprint, energy consumption, and execution latency for each client--edge training phase directly on NVIDIA RTX 4090 GPUs and NVIDIA Jetson AGX Orin\footnote{\textcolor{red}{\url{https://www.nvidia.com/en-us/autonomous-machines/embedded-systems/jetson-orin/}}} edge devices (detailed jointly in later discussions)\footnote{\textcolor{red}{Details of the relevant hardware experiments can be found in \textbf{Appendix H}.}}. To simulate unreliable client data, we randomly selected 4 clients and injected small batches of mislabeled samples into their local datasets. We evaluate ELSA on both encoder-only and decoder-only architectures. For the encoder-only setting, we adopt BERT-base-uncased, which consists of 12 Transformer blocks, a hidden dimension of 768, and 12 attention heads, totaling approximately 110 million parameters \cite{MT-FBERT,FedBERT}. For the decoder-only setting, we employ Qwen3-4B\footnote{\textcolor{red}{\url{https://huggingface.co/Qwen/Qwen3-4B}}}, which consists of 36 Transformer blocks, a hidden dimension of 2,560, 32 attention heads, and 8 key--value heads, totaling approximately 4 billion parameters. To rigorously evaluate the performance of ELSA, we implement a hybrid heterogeneity scheme that simultaneously accounts for quantity skew and label distribution skew: \textit{Quantity Skew}: each client $n$ allocates a local subset $\mathcal{D}_n$, with the size $|\mathcal{D}_n|$ proportional to a performance factor $\chi_n = \frac{n+1}{\Omega_k}$, where $\Omega_k$ denotes the overall samples related to edge $k$; \textit{Label Distribution Skew}: we impose class imbalance within each local dataset using the Dirichlet distribution $\text{Dir}(\hat{\alpha})$. We set $\hat{\alpha}=$\{0.1, 0.2\} to simulate varying statistical heterogeneity \cite{FedDCC}.}
\begin{figure*}[!t]
	\centering
	\subfigure[BERT Trec]{
		\includegraphics[trim=0.1cm 0.05cm 0.6cm 0cm, clip, width=0.465\columnwidth]{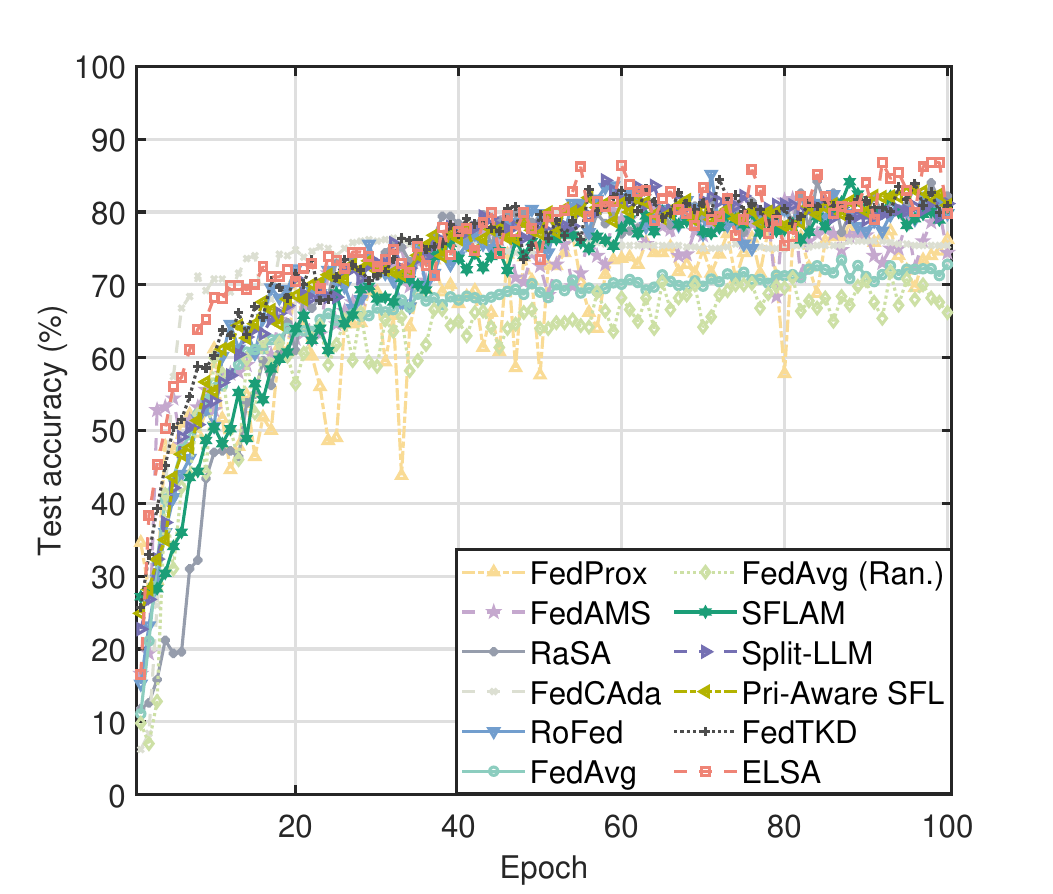}
	}
    \hfill
	\subfigure[BERT AG\_News]{
		\includegraphics[trim=0.1cm 0.05cm 0.6cm 0cm, clip, width=0.465\columnwidth]{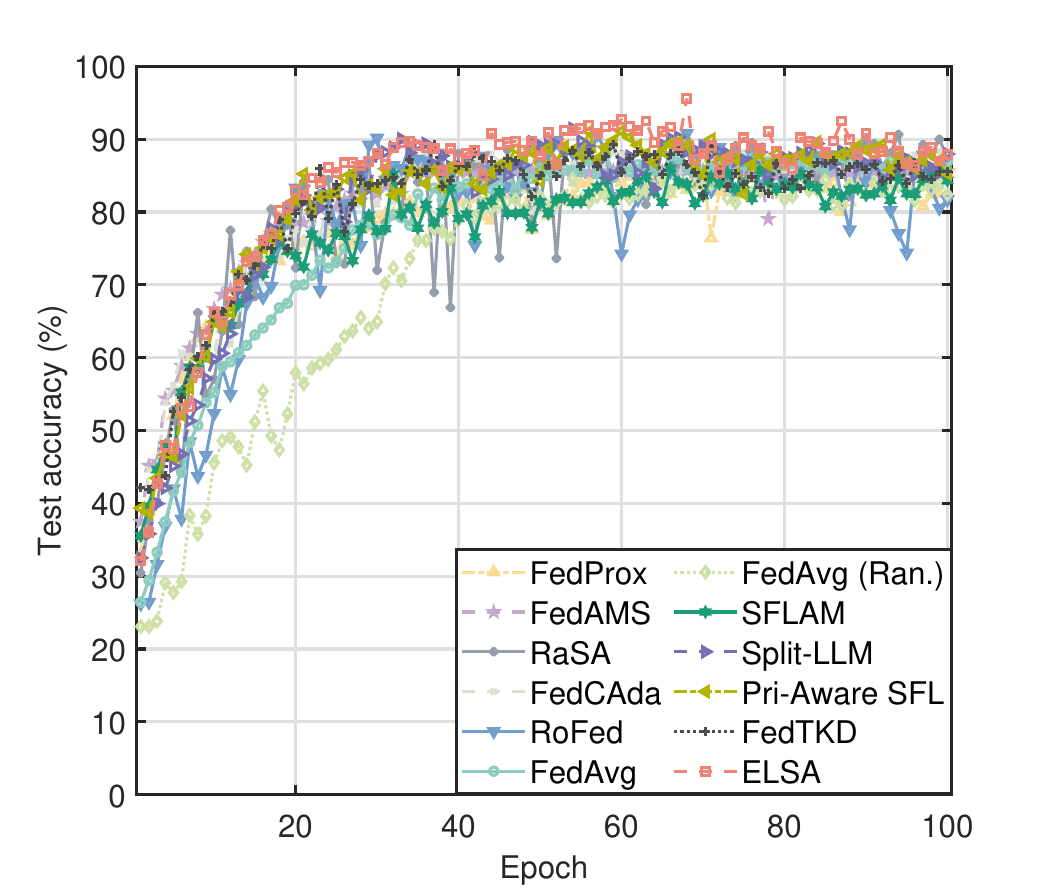}
	}
	\subfigure[BERT Emotion]{
		\includegraphics[trim=0.1cm 0.05cm 0.6cm 0cm, clip, width=0.465\columnwidth]{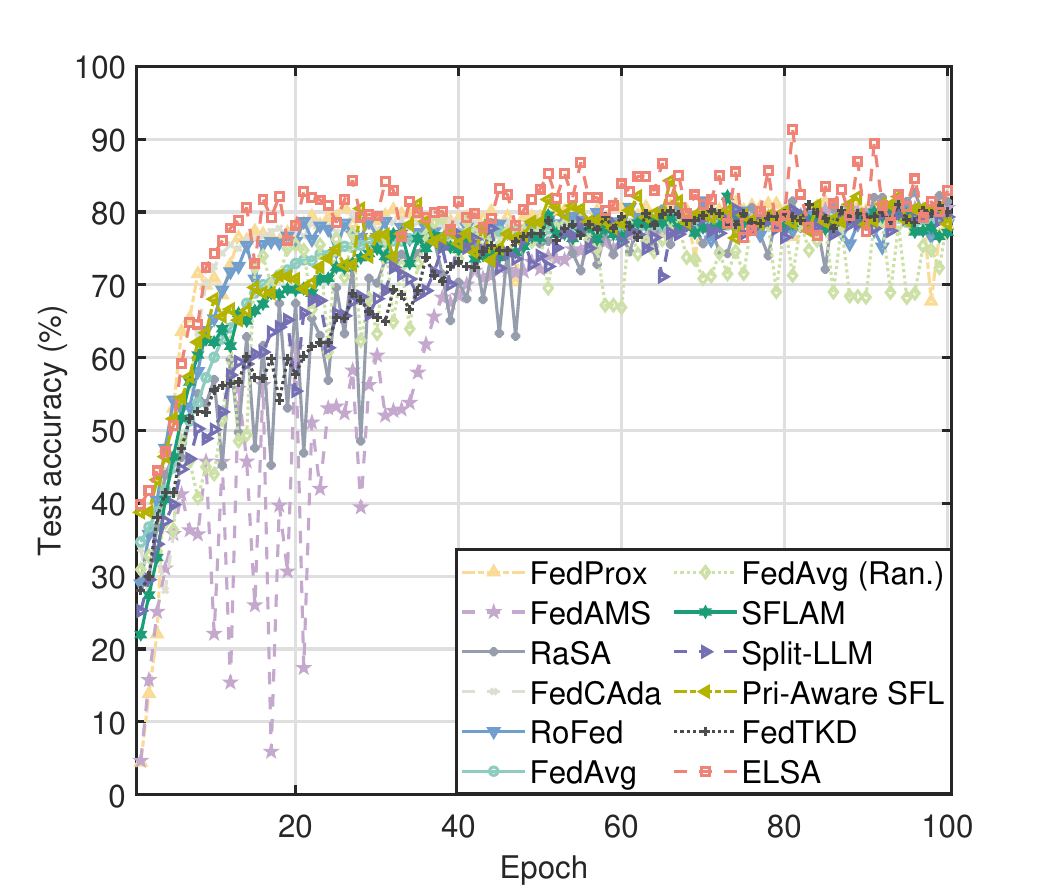}
	}
	\subfigure[BERT Banking]{
		\includegraphics[trim=0.1cm 0.05cm 0.6cm 0cm, clip, width=0.465\columnwidth]{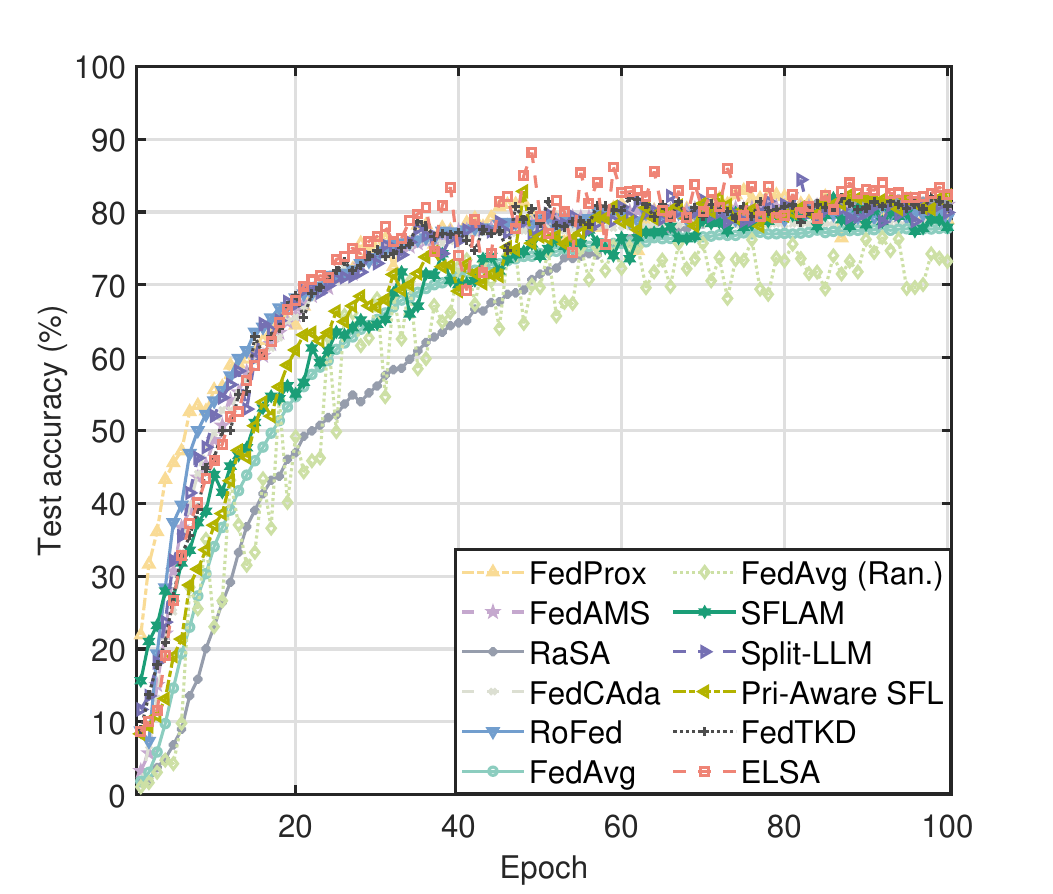}
	}
	\subfigure[BERT Trec]{
		\includegraphics[trim=0.1cm 0.05cm 0.6cm 0cm, clip, width=0.465\columnwidth]{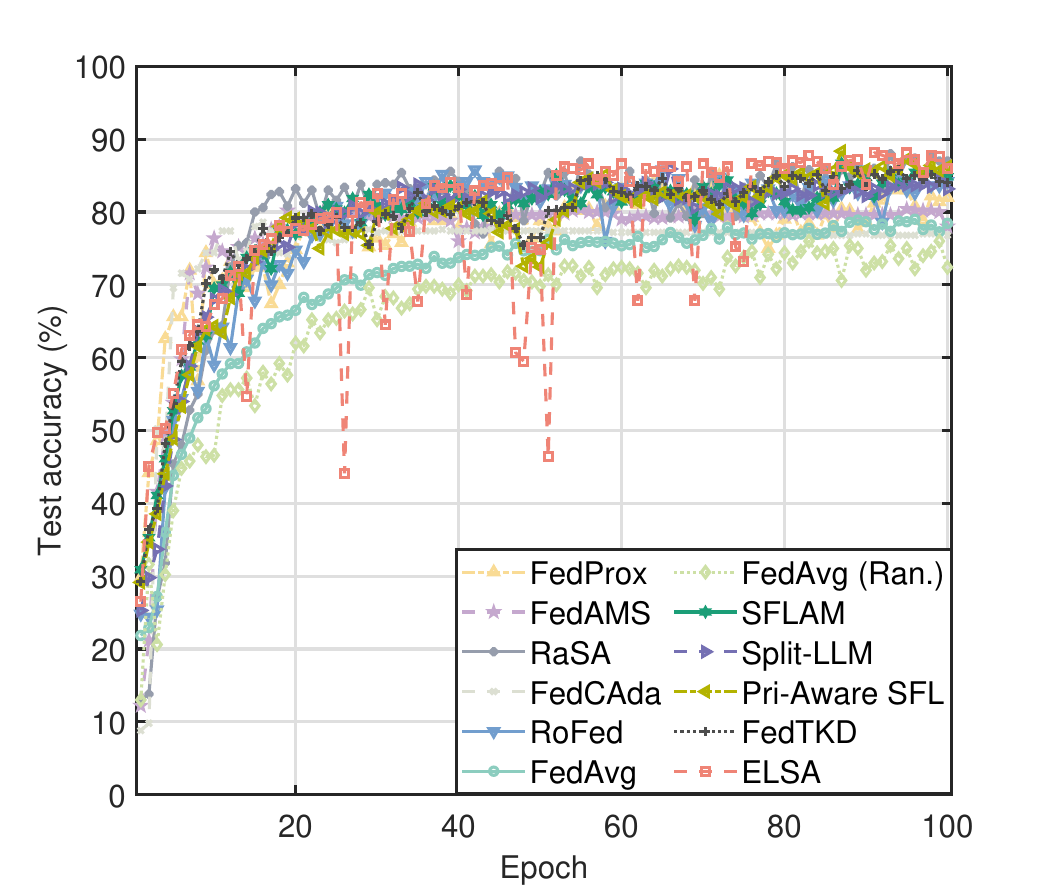}
	}
	\subfigure[BERT AG\_News]{
		\includegraphics[trim=0.1cm 0.05cm 0.6cm 0cm, clip, width=0.465\columnwidth]{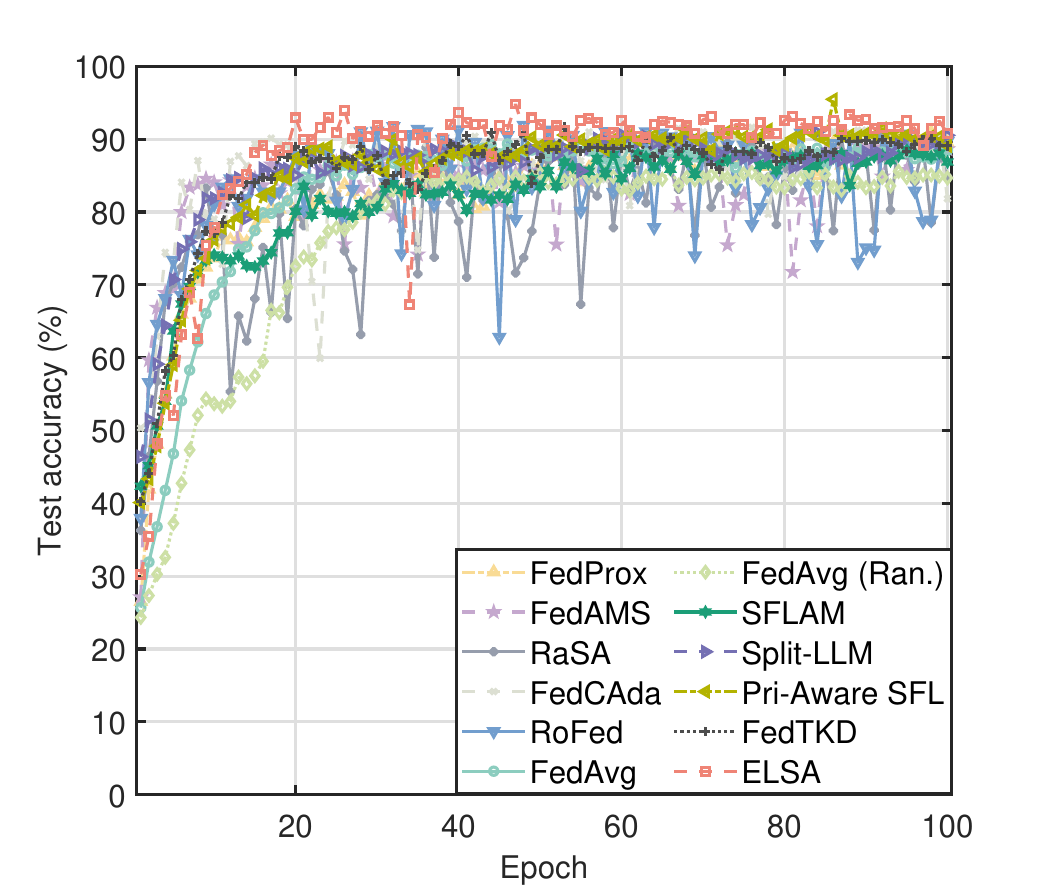}
	}
	\subfigure[BERT Emotion]{
		\includegraphics[trim=0.1cm 0.05cm 0.6cm 0cm, clip, width=0.465\columnwidth]{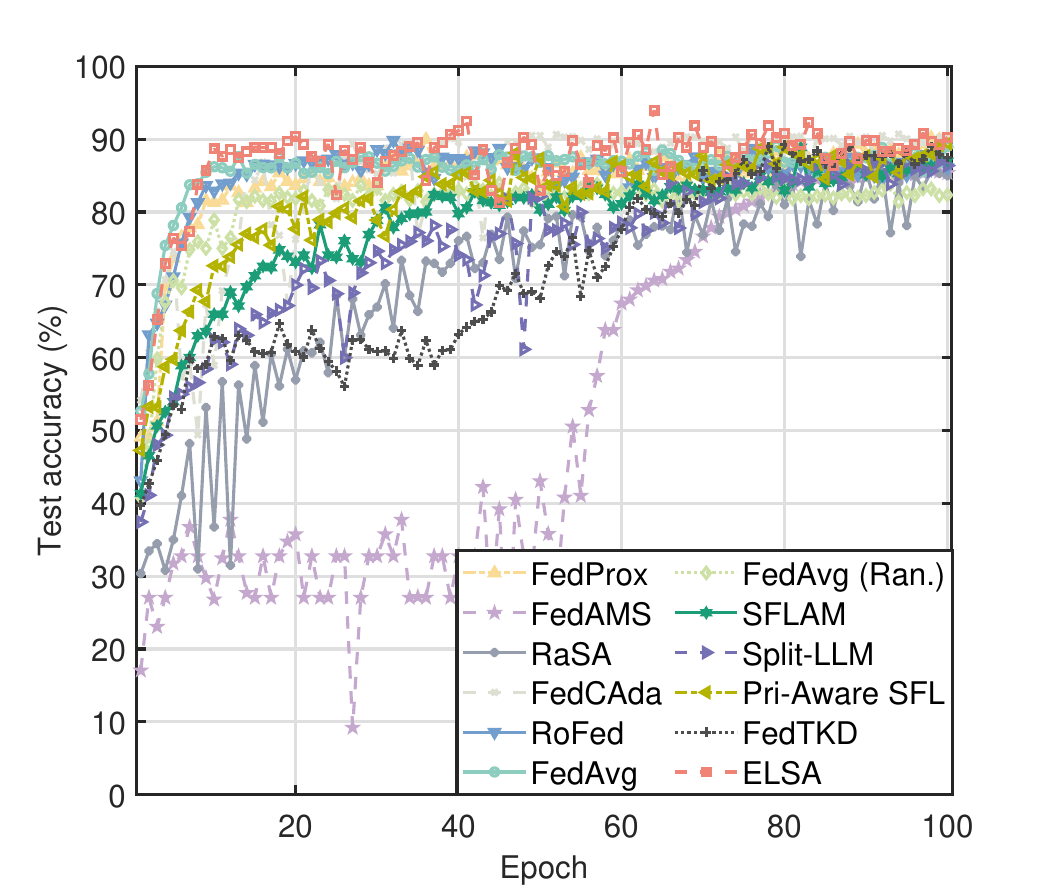}
	}
	\subfigure[BERT Banking]{
		\includegraphics[trim=0.1cm 0.05cm 0.6cm 0cm, clip, width=0.465\columnwidth]{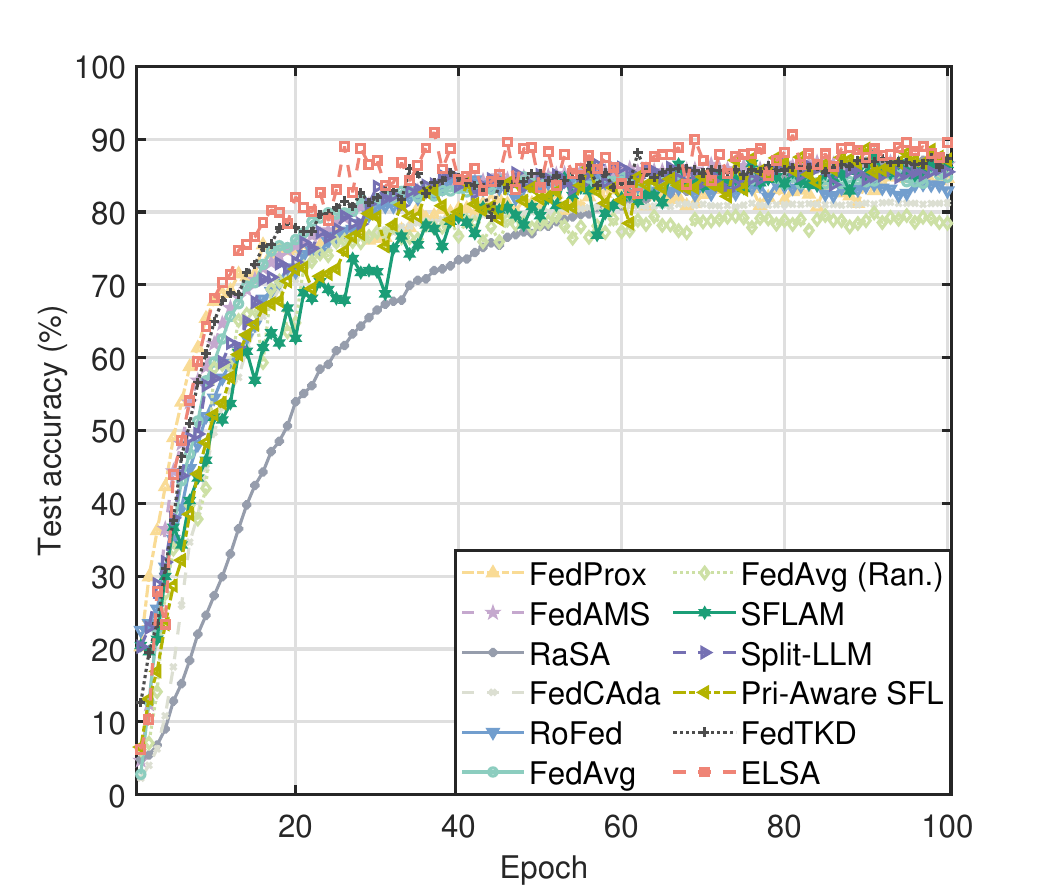}
	}

	\subfigure[Qwen3-4B Trec]{
		\includegraphics[trim=0.1cm 0.05cm 0.6cm 0cm, clip, width=0.465\columnwidth]{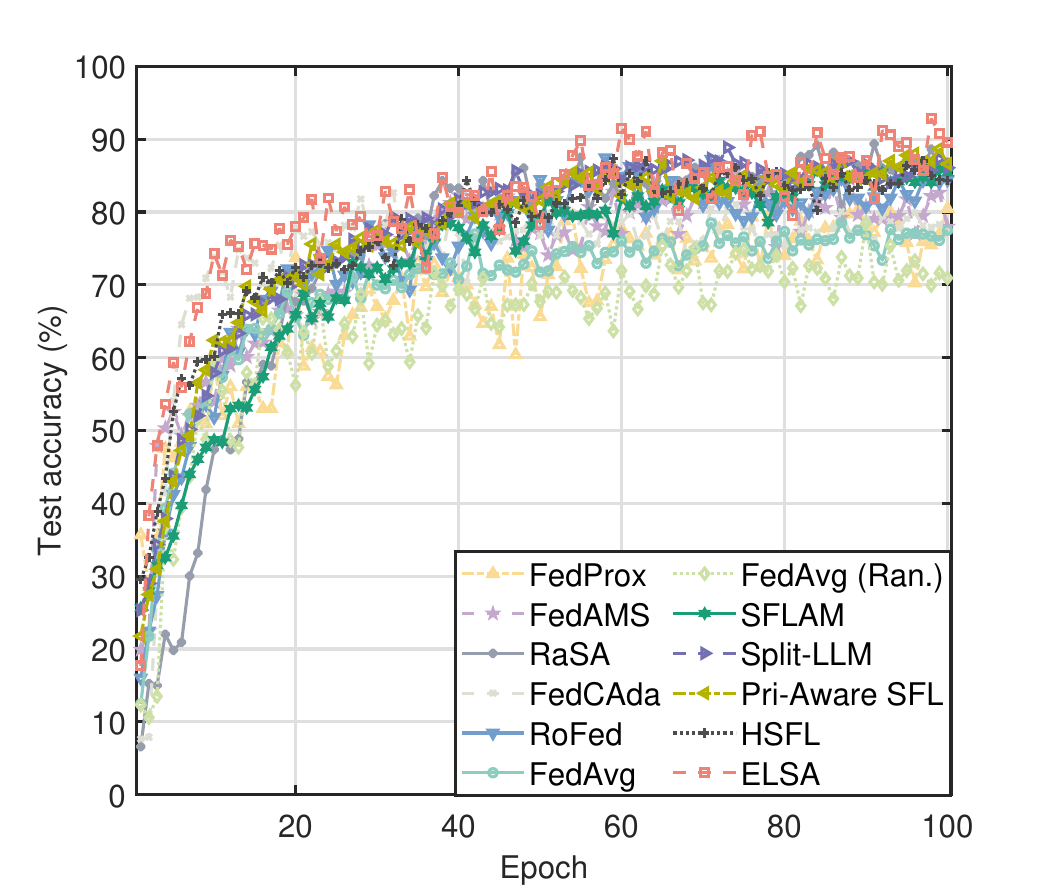}
	}
	\subfigure[Qwen3-4B AG\_News]{
		\includegraphics[trim=0.1cm 0.05cm 0.6cm 0cm, clip, width=0.465\columnwidth]{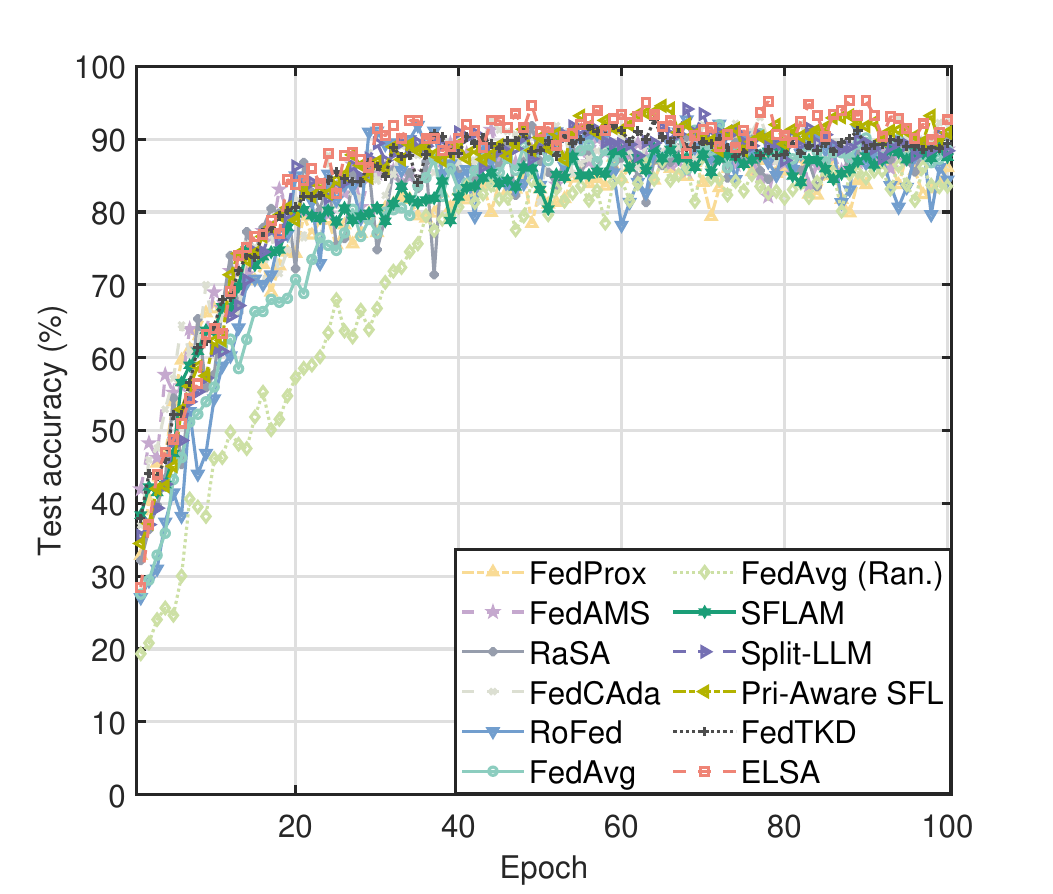}
	}
	\subfigure[Qwen3-4B Emotion]{
		\includegraphics[trim=0.1cm 0.05cm 0.6cm 0cm, clip, width=0.465\columnwidth]{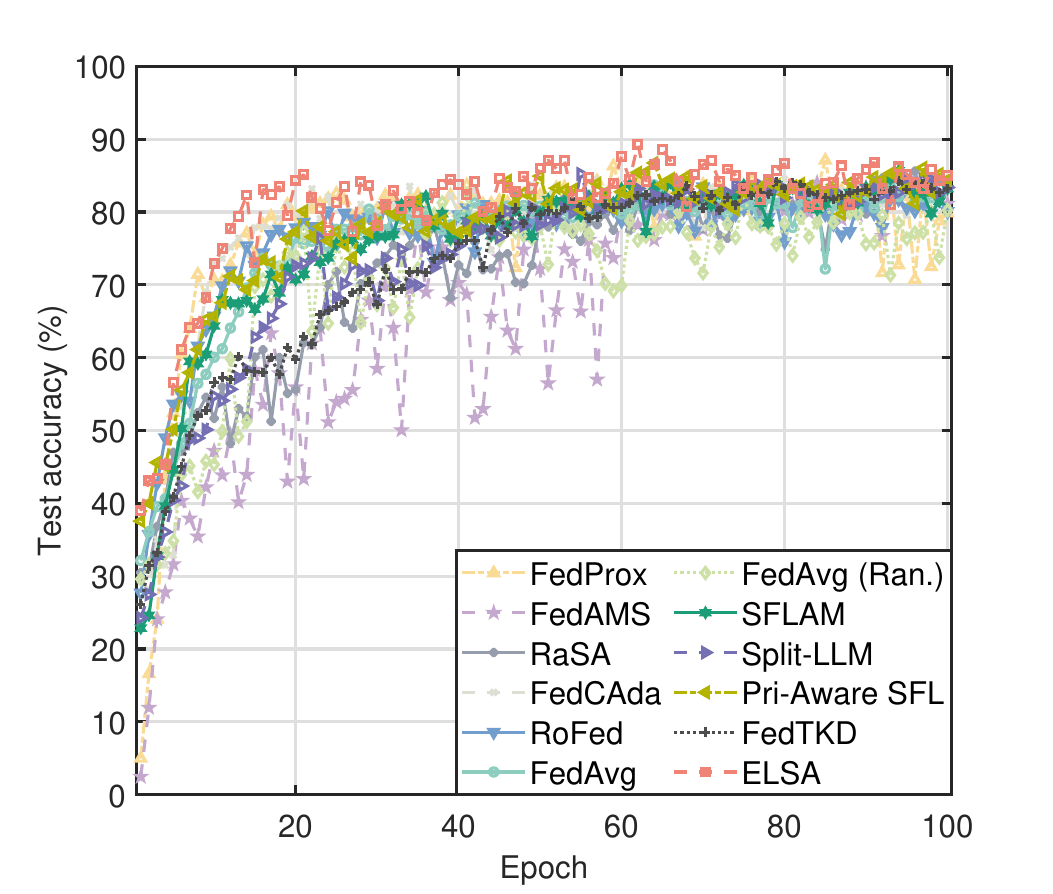}
	}
	\subfigure[Qwen3-4B Banking]{
		\includegraphics[trim=0.1cm 0.05cm 0.6cm 0cm, clip, width=0.465\columnwidth]{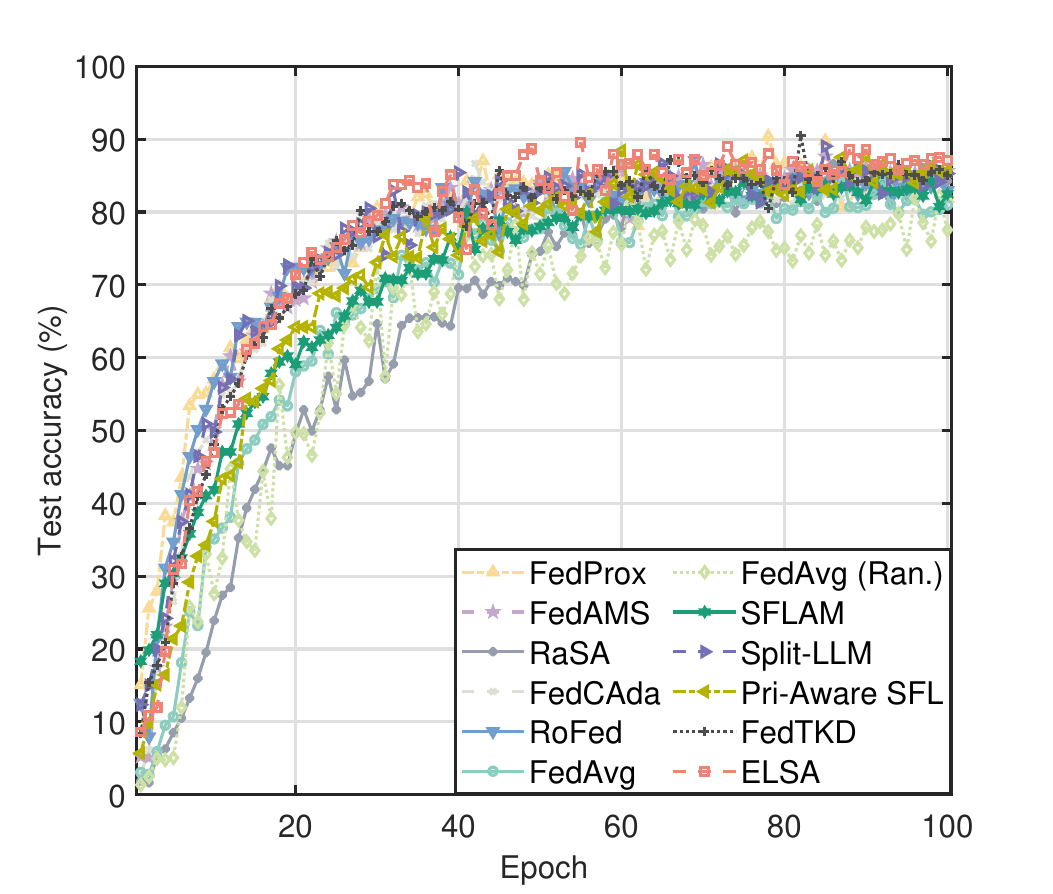}
	}
	\subfigure[Qwen3-4B Trec]{
		\includegraphics[trim=0.1cm 0.05cm 0.6cm 0cm, clip, width=0.465\columnwidth]{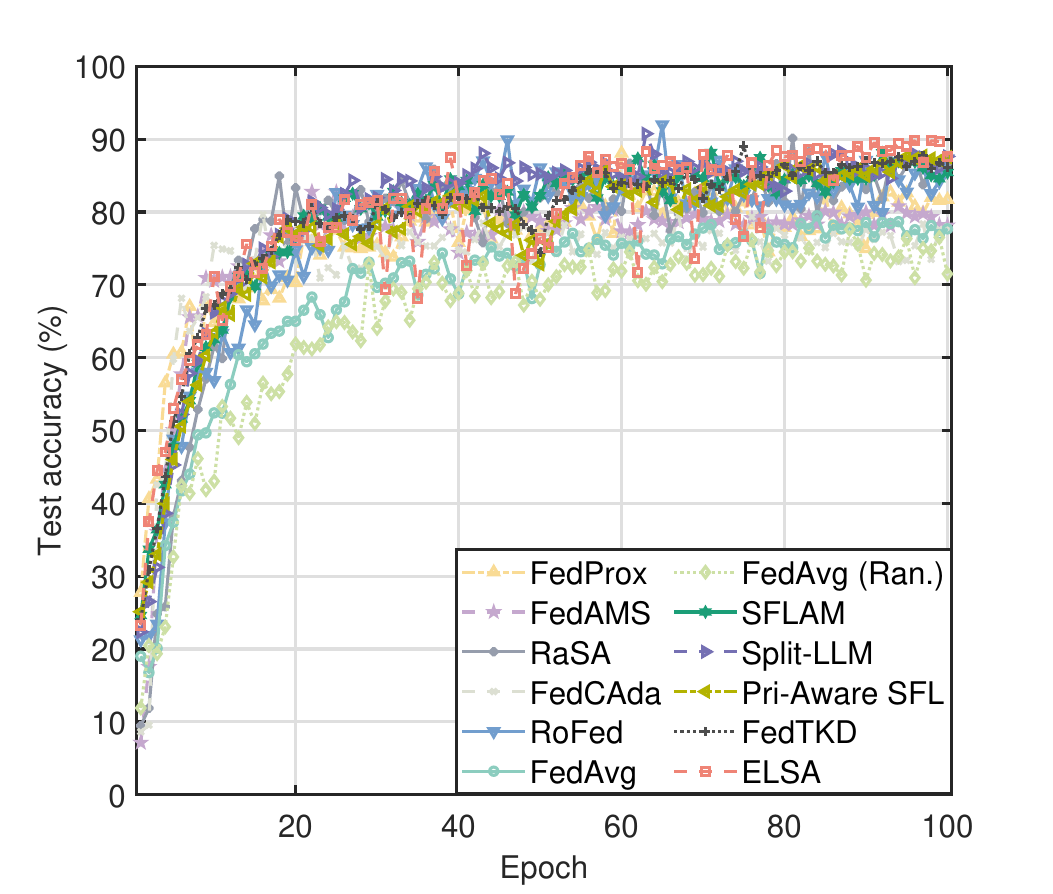}
	}
	\subfigure[Qwen3-4B AG\_News]{
		\includegraphics[trim=0.1cm 0.05cm 0.6cm 0cm, clip, width=0.465\columnwidth]{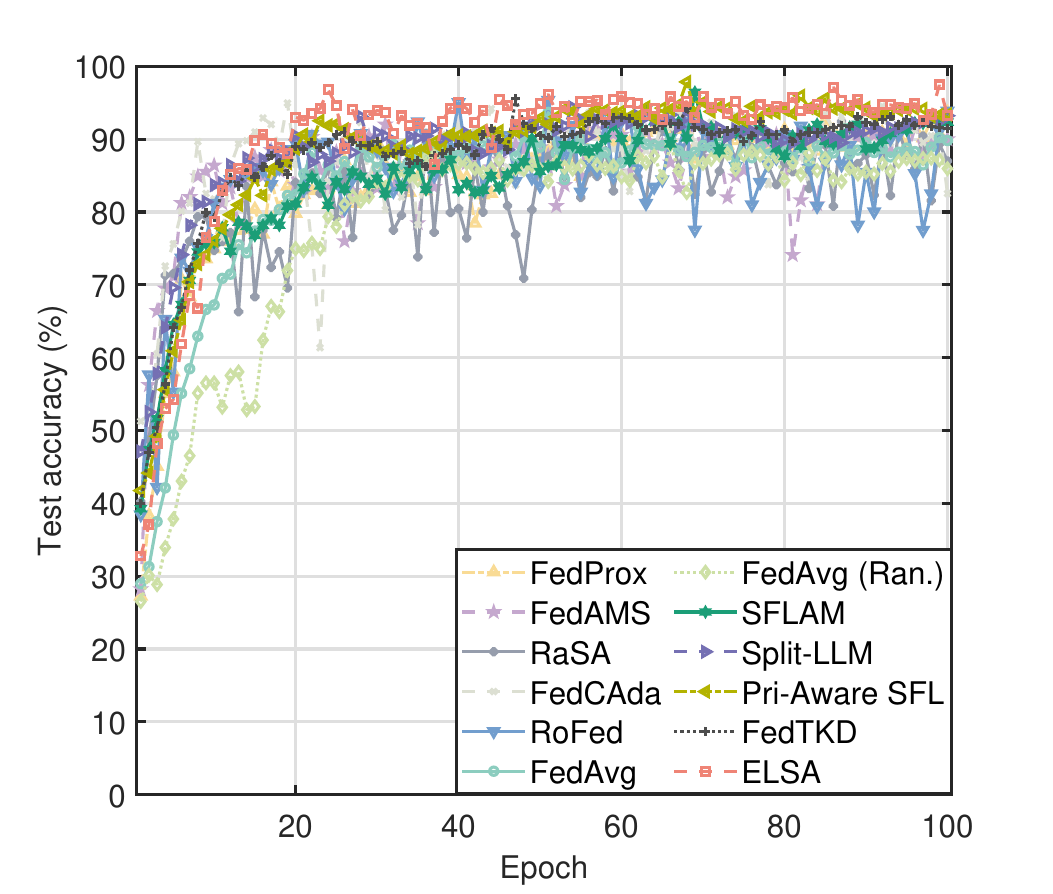}
	}
	\subfigure[Qwen3-4B Emotion]{
		\includegraphics[trim=0.1cm 0.05cm 0.6cm 0cm, clip, width=0.465\columnwidth]{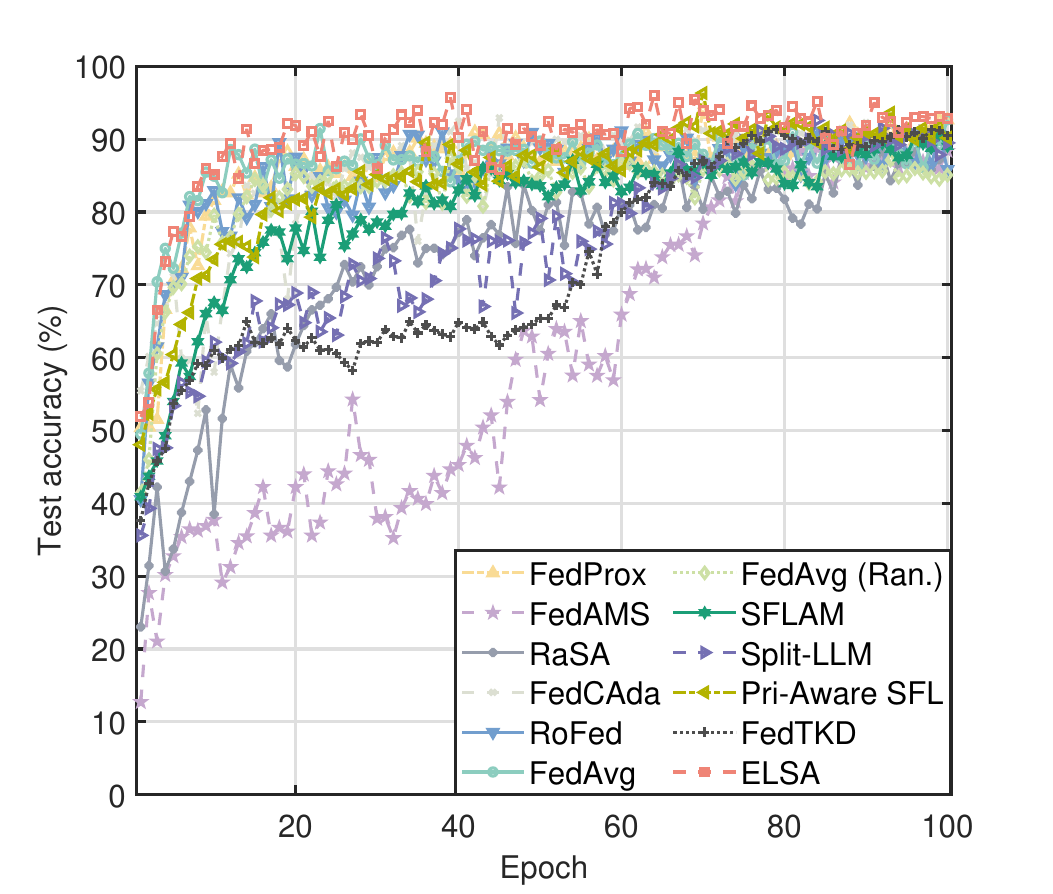}
	}
	\subfigure[Qwen3-4B Banking]{
		\includegraphics[trim=0.1cm 0.05cm 0.6cm 0cm, clip, width=0.465\columnwidth]{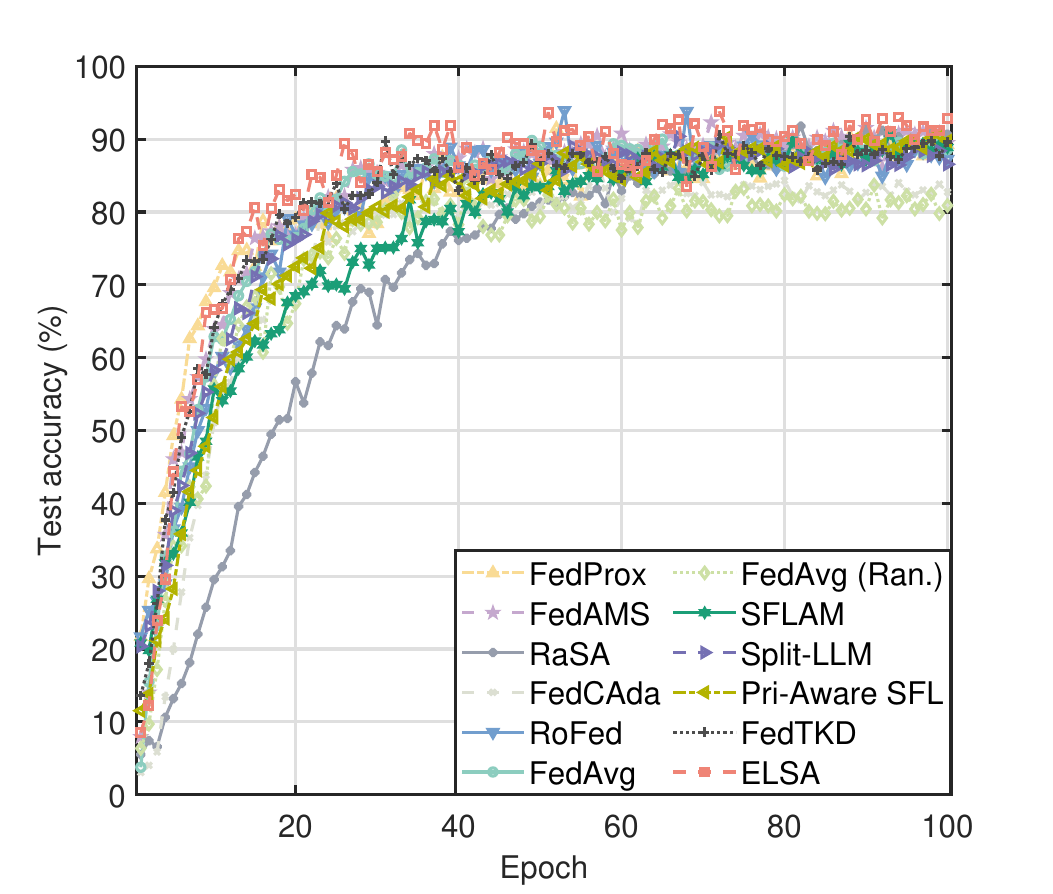}
	}
	
	\vspace{-2mm}
	\caption{\textcolor{blue}{Test performance comparison of BERT and Qwen3-4B across TC datasets under two levels of data heterogeneity: (a)–(d) and (i)–(l)  correspond to $\hat{\alpha}=$ 0.1, while (e)–(h) and (m)–(p) use $\hat{\alpha}=$ 0.2.}}
	\label{fig:4}
    \vspace{-3mm}
\end{figure*}

\begin{table*}[!t]
    \centering
    \caption{\textcolor{blue}{Performance comparison on four NLP tasks under different data heterogeneity settings ($\hat{\alpha} =$ 0.1 and $\hat{\alpha} =$ 0.2) and $\rho=$ 4.2. Accuracy (\%) is reported for RTE and CB, while F1/EM is reported for MultiRC and SQuAD. Bold indicates the best performance in each column (mean ± std over 5 independent runs).}}
    \label{tab:main_results}
    \scriptsize
    \renewcommand{\arraystretch}{1.05}
    \begin{adjustbox}{max width=\textwidth}
        \setlength{\tabcolsep}{3.5pt}
        \begin{tabular}{c|c|cccc|cccc}
            \toprule
            \multirow{4}{*}{\textbf{Model}} & \multirow{4}{*}{\textbf{Method}} & \multicolumn{4}{c|}{\textbf{$\hat{\alpha}$ = 0.1}} & \multicolumn{4}{c}{\textbf{$\hat{\alpha}$ = 0.2}} \\
            \cmidrule(lr){3-6} \cmidrule(lr){7-10}
            & & RTE & CB & MultiRC & SQuAD & RTE & CB & MultiRC & SQuAD \\
            & & Acc. & Acc. & F1/EM & EM/F1 & Acc. & Acc. & F1/EM & EM/F1 \\
            \midrule
            \multirow{13}{*}{\textbf{\shortstack{BERT-\\base-uncased}}} 
            & FedAvg          & 79.21{\tiny$\pm$0.31} & 80.23{\tiny$\pm$0.58} & 79.61{\tiny$\pm$0.25}/50.17{\tiny$\pm$0.33} & 76.92{\tiny$\pm$0.29}/87.06{\tiny$\pm$0.22} & 79.91{\tiny$\pm$0.27} & 81.03{\tiny$\pm$0.42} & 79.96{\tiny$\pm$0.24}/52.03{\tiny$\pm$0.31} & 77.22{\tiny$\pm$0.26}/87.72{\tiny$\pm$0.20} \\
            & FedAvg (Ran.)   & 78.48{\tiny$\pm$0.45} & 79.26{\tiny$\pm$0.63} & 78.23{\tiny$\pm$0.29}/48.29{\tiny$\pm$0.38} & 75.33{\tiny$\pm$0.32}/85.66{\tiny$\pm$0.27} & 79.03{\tiny$\pm$0.31} & 79.77{\tiny$\pm$0.51} & 79.11{\tiny$\pm$0.28}/49.06{\tiny$\pm$0.35} & 76.52{\tiny$\pm$0.30}/85.98{\tiny$\pm$0.25} \\
            & FedProx         & 78.92{\tiny$\pm$0.29} & 82.14{\tiny$\pm$0.46} & 80.68{\tiny$\pm$0.23}/49.19{\tiny$\pm$0.31} & 78.33{\tiny$\pm$0.27}/\textbf{87.31}{\tiny$\pm$0.21} & 79.54{\tiny$\pm$0.28} & 83.62{\tiny$\pm$0.38} & 81.33{\tiny$\pm$0.22}/50.02{\tiny$\pm$0.30} & 79.01{\tiny$\pm$0.26}/87.70{\tiny$\pm$0.19} \\
            & FedAMS          & 80.13{\tiny$\pm$0.37} & 79.82{\tiny$\pm$0.50} & 81.44{\tiny$\pm$0.24}/51.33{\tiny$\pm$0.29} & 77.66{\tiny$\pm$0.31}/86.21{\tiny$\pm$0.24} & 81.22{\tiny$\pm$0.26} & 82.38{\tiny$\pm$0.48} & 81.76{\tiny$\pm$0.23}/52.05{\tiny$\pm$0.27} & 78.02{\tiny$\pm$0.29}/87.05{\tiny$\pm$0.22} \\
            & RaSA             & 78.92{\tiny$\pm$0.30} & 81.24{\tiny$\pm$0.42} & 79.55{\tiny$\pm$0.26}/51.27{\tiny$\pm$0.30} & 78.38{\tiny$\pm$0.28}/86.79{\tiny$\pm$0.23} & 80.53{\tiny$\pm$0.29} & 83.24{\tiny$\pm$0.36} & 80.62{\tiny$\pm$0.25}/52.08{\tiny$\pm$0.28} & 78.99{\tiny$\pm$0.27}/86.95{\tiny$\pm$0.21} \\
            & FedCAda         & 79.23{\tiny$\pm$0.28} & 82.66{\tiny$\pm$0.39} & 79.81{\tiny$\pm$0.24}/51.39{\tiny$\pm$0.28} & 78.52{\tiny$\pm$0.27}/86.45{\tiny$\pm$0.22} & 81.31{\tiny$\pm$0.27} & 84.23{\tiny$\pm$0.34} & 80.12{\tiny$\pm$0.23}/52.41{\tiny$\pm$0.27} & 80.03{\tiny$\pm$0.26}/87.29{\tiny$\pm$0.20} \\
            & RoFed            & 79.18{\tiny$\pm$0.29} & 82.71{\tiny$\pm$0.41} & 80.22{\tiny$\pm$0.25}/51.09{\tiny$\pm$0.29} & 78.50{\tiny$\pm$0.28}/86.55{\tiny$\pm$0.23} & 82.33{\tiny$\pm$0.31} & 83.19{\tiny$\pm$0.40} & 82.19{\tiny$\pm$0.24}/52.32{\tiny$\pm$0.28} & 79.33{\tiny$\pm$0.27}/87.16{\tiny$\pm$0.21} \\
            & SFLAM            & 79.45{\tiny$\pm$0.38} & 82.90{\tiny$\pm$0.45} & 80.51{\tiny$\pm$0.32}/51.18{\tiny$\pm$0.36} & 78.30{\tiny$\pm$0.34}/86.62{\tiny$\pm$0.28} & 82.05{\tiny$\pm$0.36} & 83.85{\tiny$\pm$0.39} & 81.05{\tiny$\pm$0.31}/52.15{\tiny$\pm$0.34} & 79.35{\tiny$\pm$0.33}/87.35{\tiny$\pm$0.26} \\
            & Split-LLM        & 79.52{\tiny$\pm$0.36} & 81.75{\tiny$\pm$0.48} & 80.10{\tiny$\pm$0.31}/50.85{\tiny$\pm$0.35} & 77.88{\tiny$\pm$0.33}/86.48{\tiny$\pm$0.27} & 81.15{\tiny$\pm$0.34} & 83.40{\tiny$\pm$0.41} & 80.78{\tiny$\pm$0.30}/51.90{\tiny$\pm$0.33} & 79.12{\tiny$\pm$0.32}/87.18{\tiny$\pm$0.25} \\
            & Pri-Aware SFL    & 79.60{\tiny$\pm$0.34} & 82.15{\tiny$\pm$0.43} & 80.73{\tiny$\pm$0.29}/51.15{\tiny$\pm$0.33} & 78.02{\tiny$\pm$0.31}/86.55{\tiny$\pm$0.26} & 81.42{\tiny$\pm$0.32} & 84.05{\tiny$\pm$0.37} & 80.92{\tiny$\pm$0.28}/52.05{\tiny$\pm$0.31} & 79.33{\tiny$\pm$0.30}/87.25{\tiny$\pm$0.24} \\
            & HSFL           & 79.45{\tiny$\pm$0.30} & 81.68{\tiny$\pm$0.44} & 80.42{\tiny$\pm$0.27}/51.05{\tiny$\pm$0.30} & 77.62{\tiny$\pm$0.29}/86.42{\tiny$\pm$0.23} & 81.07{\tiny$\pm$0.28} & 83.91{\tiny$\pm$0.38} & 81.03{\tiny$\pm$0.25}/51.88{\tiny$\pm$0.28} & 79.44{\tiny$\pm$0.27}/87.05{\tiny$\pm$0.21} \\
            \cmidrule(lr){2-10}
            & \textbf{ELSA (Ours)} & \textbf{80.93}{\tiny$\pm$0.22} & \textbf{83.93}{\tiny$\pm$0.31} & \textbf{81.78}{\tiny$\pm$0.19}/\textbf{52.06}{\tiny$\pm$0.24} & \textbf{79.24}{\tiny$\pm$0.21}/87.17{\tiny$\pm$0.18} & \textbf{83.13}{\tiny$\pm$0.20} & \textbf{84.99}{\tiny$\pm$0.28} & \textbf{82.72}{\tiny$\pm$0.18}/\textbf{53.16}{\tiny$\pm$0.22} & \textbf{80.44}{\tiny$\pm$0.20}/\textbf{88.04}{\tiny$\pm$0.17} \\
            & \textit{Gain} & \textit{+0.80} & \textit{+1.03} & \textit{+0.34/+0.67} & \textit{+0.72/-0.14} & \textit{+0.80} & \textit{+0.76} & \textit{+0.53/+0.75} & \textit{+0.41/+0.32} \\
            \midrule
            \multirow{13}{*}{\textbf{Qwen3-4B}} 
            & FedAvg          & 79.45{\tiny$\pm$0.38} & 82.03{\tiny$\pm$0.55} & 80.82{\tiny$\pm$0.24}/51.35{\tiny$\pm$0.30} & 78.15{\tiny$\pm$0.27}/87.82{\tiny$\pm$0.21} & 81.30{\tiny$\pm$0.32} & 83.05{\tiny$\pm$0.49} & 81.25{\tiny$\pm$0.33}/53.10{\tiny$\pm$0.29} & 78.65{\tiny$\pm$0.25}/88.40{\tiny$\pm$0.19} \\
            & FedAvg (Ran.)   & 79.60{\tiny$\pm$0.42} & 81.05{\tiny$\pm$0.61} & 79.40{\tiny$\pm$0.28}/49.30{\tiny$\pm$0.36} & 76.80{\tiny$\pm$0.30}/86.40{\tiny$\pm$0.25} & 80.25{\tiny$\pm$0.38} & 81.50{\tiny$\pm$0.53} & 80.20{\tiny$\pm$0.25}/50.10{\tiny$\pm$0.33} & 77.80{\tiny$\pm$0.21}/86.80{\tiny$\pm$0.23} \\
            & FedProx         & 80.12{\tiny$\pm$0.35} & 83.85{\tiny$\pm$0.42} & 81.50{\tiny$\pm$0.22}/50.40{\tiny$\pm$0.28} & 79.20{\tiny$\pm$0.26}/88.10{\tiny$\pm$0.20} & 80.95{\tiny$\pm$0.31} & 85.10{\tiny$\pm$0.37} & 82.15{\tiny$\pm$0.21}/51.20{\tiny$\pm$0.27} & 79.80{\tiny$\pm$0.25}/88.50{\tiny$\pm$0.18} \\
            & FedAMS          & 81.35{\tiny$\pm$0.31} & 81.60{\tiny$\pm$0.52} & 82.05{\tiny$\pm$0.23}/52.25{\tiny$\pm$0.27} & 78.90{\tiny$\pm$0.25}/87.05{\tiny$\pm$0.22} & 82.40{\tiny$\pm$0.29} & 83.80{\tiny$\pm$0.44} & 82.60{\tiny$\pm$0.26}/52.95{\tiny$\pm$0.26} & 79.25{\tiny$\pm$0.27}/87.90{\tiny$\pm$0.20} \\
            & RaSA             & 81.63{\tiny$\pm$0.28} & 82.90{\tiny$\pm$0.45} & 80.75{\tiny$\pm$0.24}/52.10{\tiny$\pm$0.28} & 79.45{\tiny$\pm$0.27}/88.03{\tiny$\pm$0.21} & 81.85{\tiny$\pm$0.30} & 84.60{\tiny$\pm$0.39} & 81.80{\tiny$\pm$0.23}/52.85{\tiny$\pm$0.27} & 80.10{\tiny$\pm$0.26}/87.80{\tiny$\pm$0.19} \\
            & FedCAda         & 80.50{\tiny$\pm$0.33} & 84.10{\tiny$\pm$0.38} & 81.05{\tiny$\pm$0.22}/52.20{\tiny$\pm$0.26} & 79.60{\tiny$\pm$0.25}/87.25{\tiny$\pm$0.20} & 82.50{\tiny$\pm$0.28} & 84.52{\tiny$\pm$0.35} & 81.35{\tiny$\pm$0.21}/53.20{\tiny$\pm$0.25} & 80.85{\tiny$\pm$0.24}/88.10{\tiny$\pm$0.18} \\
            & RoFed            & 81.08{\tiny$\pm$0.30} & 84.21{\tiny$\pm$0.17} & 82.13{\tiny$\pm$0.23}/52.00{\tiny$\pm$0.27} & 79.55{\tiny$\pm$0.26}/87.30{\tiny$\pm$0.21} & 83.20{\tiny$\pm$0.27} & 84.90{\tiny$\pm$0.41} & 83.10{\tiny$\pm$0.22}/53.15{\tiny$\pm$0.26} & 80.20{\tiny$\pm$0.25}/87.95{\tiny$\pm$0.19} \\
            & SFLAM            & 81.55{\tiny$\pm$0.41} & 83.70{\tiny$\pm$0.44} & 81.65{\tiny$\pm$0.30}/52.15{\tiny$\pm$0.34} & 79.85{\tiny$\pm$0.32}/87.65{\tiny$\pm$0.26} & 83.45{\tiny$\pm$0.35} & 85.35{\tiny$\pm$0.38} & 82.80{\tiny$\pm$0.29}/53.05{\tiny$\pm$0.32} & 80.95{\tiny$\pm$0.31}/88.35{\tiny$\pm$0.24} \\
            & Split-LLM        & 81.20{\tiny$\pm$0.39} & 83.25{\tiny$\pm$0.46} & 81.30{\tiny$\pm$0.29}/51.85{\tiny$\pm$0.33} & 79.50{\tiny$\pm$0.31}/87.45{\tiny$\pm$0.25} & 83.05{\tiny$\pm$0.34} & 85.05{\tiny$\pm$0.40} & 82.45{\tiny$\pm$0.28}/52.80{\tiny$\pm$0.31} & 80.55{\tiny$\pm$0.30}/88.31{\tiny$\pm$0.23} \\
            & Pri-Aware SFL    & 81.76{\tiny$\pm$0.36} & 83.50{\tiny$\pm$0.43} & 82.21{\tiny$\pm$0.27}/52.00{\tiny$\pm$0.31} & 81.03{\tiny$\pm$0.29}/87.55{\tiny$\pm$0.24} & 83.25{\tiny$\pm$0.32} & 85.20{\tiny$\pm$0.37} & 83.03{\tiny$\pm$0.26}/52.95{\tiny$\pm$0.29} & 80.75{\tiny$\pm$0.28}/88.30{\tiny$\pm$0.22} \\
            & HSFL           & 80.98{\tiny$\pm$0.32} & 83.25{\tiny$\pm$0.40} & 81.25{\tiny$\pm$0.23}/52.05{\tiny$\pm$0.28} & 80.05{\tiny$\pm$0.27}/87.37{\tiny$\pm$0.21} & 82.65{\tiny$\pm$0.30} & 84.85{\tiny$\pm$0.39} & 83.00{\tiny$\pm$0.23}/53.05{\tiny$\pm$0.26} & 79.85{\tiny$\pm$0.25}/87.24{\tiny$\pm$0.19} \\
            \cmidrule(lr){2-10}
            & \textbf{ELSA (Ours)} & \textbf{82.09}{\tiny$\pm$0.20} & \textbf{84.25}{\tiny$\pm$0.26} & \textbf{82.54}{\tiny$\pm$0.17}/\textbf{52.88}{\tiny$\pm$0.21} & \textbf{81.42}{\tiny$\pm$0.19}/\textbf{88.15}{\tiny$\pm$0.16} & \textbf{84.55}{\tiny$\pm$0.18} & \textbf{85.65}{\tiny$\pm$0.23} & \textbf{83.57}{\tiny$\pm$0.16}/\textbf{53.95}{\tiny$\pm$0.20} & \textbf{82.61}{\tiny$\pm$0.18}/\textbf{88.92}{\tiny$\pm$0.15} \\
            & \textit{Gain} & \textit{+0.33} & \textit{+0.04} & \textit{+0.33/+0.63} & \textit{+0.39/+0.05} & \textit{+1.10} & \textit{+0.30} & \textit{+0.47/+0.75} & \textit{+1.66/+0.42} \\
            \bottomrule
        \end{tabular}
    \end{adjustbox}
    \vspace{0.5em}
    
    \tiny
    \textcolor{blue}{\textit{Note:} Values are reported as mean{\tiny$\pm$}std over 5 independent runs. \textit{Gain} measures the performance difference between ELSA and the best baseline in each metric. Baselines are categorized as: General Baselines: (FedAvg, FedAvg (Ran.)); Aggregation Optimization \& Edge FL/HFL (FedProx, FedAMS, RaSA, FedCAda); SL for LLMs \& SL FL/HFL (RoFed, SFLAM, Split-LLM, Pri-Aware SFL, HSFL).}
    \vspace{-3mm}
\end{table*}

\subsection{Evaluation of Learning Efficiency and Model Performance}

To evaluate the model training performance for TC datasets, we prioritize the convergence rate to evaluate the communication efficiency of each method. Conversely, for NLI tasks that demand complex semantic reasoning, we focus on the steady-state performance (e.g., Accuracy, F1 and EM) once training has fully converged. This differentiation ensures that we capture both the learning efficiency on standard tasks and the ultimate representative capacity on more challenging reasoning benchmarks \cite{MT-FBERT,ROFED}.

\textcolor{blue}{Fig.~\ref{fig:4} compares the convergence behavior of BERT-base-uncased and Qwen3-4B on four TC datasets under two levels of data heterogeneity. Across both encoder-only and decoder-only architectures, ELSA generally converges to a higher or comparable accuracy than the 11 baselines. Under the more heterogeneous setting of $\hat{\alpha}=0.1$, ELSA maintains stable convergence despite the stronger divergence among local client updates.
When the data distributions become less heterogeneous ($\hat{\alpha}=0.2$), most methods converge more smoothly, while ELSA continues to achieve a consistently high accuracy ceiling. Although ELSA occasionally exhibits slower initial progress or transient fluctuations, particularly during the early training epochs (e.g., during the 40-60 epochs in Fig. 4(d) and the 30-50 epochs in Fig. 4(m)), its later-stage convergence is generally more stable. The similar trends observed for BERT and Qwen3-4B indicate that the effectiveness of ELSA is not tied to a specific Transformer architecture.}

\textcolor{blue}{Table~\ref{tab:main_results} further reports the steady-state performance on RTE, CB, MultiRC, and SQuAD. For BERT-base-uncased, ELSA achieves the best results on nearly all metrics under both heterogeneity settings. For example, at $\hat{\alpha}=0.1$, it reaches 80.93\% accuracy on RTE and 81.78/52.06 in F1/EM on MultiRC, improving over the strongest baselines by 0.80 and 0.34/0.67 percentage points, respectively. The only exception is the SQuAD F1 at $\hat{\alpha}=0.1$, where ELSA is 0.14 percentage points below FedProx. For Qwen3-4B, ELSA also consistently achieves the best steady-state performance. Its advantage becomes particularly evident at $\hat{\alpha}=0.2$, where it improves over the strongest baselines by 1.10 percentage points on RTE, 0.47/0.75 on MultiRC, and 1.66/0.42 on SQuAD. Furthermore, the performance gains on the test set driven by the stronger foundational model demonstrate that ELSA does not compromise the inherent capabilities or expressiveness of the decoder-only architecture (e.g., while $\hat{\alpha} = $0.1, comparing to BERT, ELSA on Qwen yields an average performance gain of around 0.95\% across the eight datasets). These results demonstrate that ELSA remains effective when scaling from the 110-million-parameter encoder-only BERT model to the approximately 4-billion-parameter decoder-only Qwen model.}

\textcolor{blue}{The comparison across the three baseline categories further reveals the contributions of ELSA. First, compared with the general FL baselines, FedAvg and FedAvg (Ran.), ELSA benefits from behavior- and trust-aware client organization instead of relying on fixed or random client participation. For instance, on BERT RTE with $\hat{\alpha}=0.1$, ELSA achieves 80.93\% accuracy, compared with 79.21\% for FedAvg and 78.48\% for FedAvg (Ran.). Second, aggregation-optimized and edge FL/HFL methods, including FedProx, FedAMS, RaSA, and FedCAda, improve local-update coordination but do not jointly account for semantic behavioral heterogeneity and split execution. Third, the LLM split-learning baselines, including RoFed, SFLAM, SplitLLM, Privacy-Aware SFL, and HSFL, reduce client-side resource requirements or improve hierarchical execution, but lack the joint integration of behavioral clustering, reliability-aware association, dynamic three-part splitting, and coherence-weighted aggregation. The consistent improvements of ELSA across these three categories are aligned with the benefits of jointly coordinating semantic, system, and network heterogeneity throughout the training process. Furthermore, to validate scalability in edge environments, we extend our evaluation to $N=$ 50 and $N=$ 100, presenting detailed analyses on convergence, communication costs, transmission efficiency, and clustering overheads (\textbf{Appendix I}).}

\begin{table*}[t]
    \centering
    \caption{\textcolor{blue}{Comprehensive comparison of dataset-specific communication time ($\times \text{10}^\text{3}$ s), average computation latency ($\times \text{10}^\text{3}$ s), and task failure rate (\%) across different baselines (using BERT-base-uncased).}}
    \label{tab:total_comm_time}
    \small
    \setlength{\tabcolsep}{4.5pt} 
    \begin{adjustbox}{max width=\textwidth} 
        \begin{tabular}{lcccccccccc}
            \toprule
            & \multicolumn{8}{c}{\textbf{Communication Time to Convergence ($\times \text{10}^\text{3}$ s)}} & \multicolumn{2}{c}{\textbf{System Efficiency \& Reliability}} \\
            \cmidrule(lr){2-9} \cmidrule(lr){10-11}
            \textbf{Method} & \textbf{AG\_News} & \textbf{Banking} & \textbf{Emotion} & \textbf{Trec} & \textbf{RTE} & \textbf{CB} & \textbf{MultiRC} & \textbf{SQuAD} & \begin{tabular}[c]{@{}c@{}}\textbf{Avg. Latency}\\\textbf{(s $\times \text{10}^\text{3}$)}\end{tabular} & \begin{tabular}[c]{@{}c@{}}\textbf{Avg. Failure}\\\textbf{Rate (\%)}\end{tabular} \\
            \midrule
            \multicolumn{11}{l}{\textit{\textbf{Vanilla \& General Baselines}}} \\
            Vanilla Model   & 10.51 & 6.01 & 7.51 & 3.38 & 14.41 & 18.02 & 39.64 & 36.93 & 28.45 (+14.20, -20.07) & - \\
            
            FedAvg          & 3.79  & 2.14 & 2.32 & 1.07 & 4.58  & 6.29  & 14.87 & 13.94 & 10.54 (+5.93, -4.03)   & - \\
            
            FedAvg (Random) & 3.93  & 2.22 & 2.50 & 1.09 & 5.00  & 7.15  & 16.87 & 14.58 & 11.89 (+6.38, -4.88)  & - \\
            
            \midrule
            \multicolumn{11}{l}{\textit{\textbf{Aggregation Optimization \& Edge FL/HFL}}} \\
            FedProx         & 3.65  & 1.97 & 2.29 & 1.04 & 4.65  & 5.72  & 13.73 & 13.08 & 10.12 (+5.31, -3.78)   & - \\
            
            FedAMS          & 3.57  & 1.89 & 2.47 & 1.00 & 4.58  & 6.43  & 14.58 & 14.16 & 10.35 (+5.41, -4.10)   & - \\
            
            RaSA            & 3.43  & 1.93 & 2.43 & 0.98 & 4.79  & 6.58  & 15.73 & 13.73 & 10.61 (+5.73, -4.33)   & - \\
            
            FedCAda         & 3.43  & 2.32 & 2.36 & 0.95 & 4.72  & 6.15  & 14.30 & 13.30 & 10.28 (+5.12, -3.92)   & - \\
            \midrule
            \multicolumn{11}{l}{\textit{\textbf{SL for LLMs \& SL FL/HFL}}} \\
            
            RoFed           & 3.36  & 1.89 & 2.32 & 0.97 & 4.43  & 6.01  & 13.44 & 13.08 & 9.82 (+3.99, -3.65)    & 3.0 (+0.4, -0.2) \\
            SFLAM           & 3.52  & 1.98 & 2.38 & 1.01 & 4.61  & 6.12  & 14.10 & 13.50 & 9.95 (+5.10, -3.69)    & 5.2 (+0.4, -0.8) \\
            Split-LLM       & 3.48  & 1.95 & 2.35 & 0.99 & 4.52  & 6.15  & 13.85 & 13.25 & 9.68 (+4.80, -3.52)    & 4.8 (+0.2, -0.6) \\
            
            Pri-Aware SFL   & 3.50  & 1.92 & 2.41 & 0.96 & 4.55  & 6.05  & 13.80 & 13.40 & 9.74 (+4.88, -3.66)    & 3.5 (+0.8, -0.2) \\
            
            HSFL          & 3.61  & 2.05 & 2.45 & 1.03 & 4.68  & 6.25  & 14.25 & 13.62 & 10.22 (+5.03, -4.02)   & 2.4 (+0.3,-0.4)\\
            \midrule
            \textbf{ELSA (Ours)} & \textbf{3.22/\textit{-0.14}} & \textbf{1.86/\textit{-0.03}} & \textbf{2.25/\textit{-0.04}} & \textbf{0.89/\textit{-0.06}} & \textbf{4.29/\textit{-0.14}} & \textbf{5.43/\textit{-0.29}} & \textbf{13.15/\textit{-0.29}} & \textbf{12.23/\textit{-0.85}} & \textbf{8.76 (+4.33, -3.14)} & \textbf{1.2 (+0.4, -0.2)} \\
            \bottomrule
        \end{tabular}
    \end{adjustbox}
    
    \vspace{0.5em}
    \footnotesize
    \textcolor{blue}{\textit{Note:} Communication time values are formatted as `time / reduction vs. best baseline'. `Avg. Latency' denotes non-communication computation duration (i.e., $T_{\text{latency}} = T_{\text{total}} - T_{\text{comm}}$). Parentheses (+max, -min) denote the maximum positive and negative deviations from the mean across all 8 evaluation datasets. `–' indicates N/A, as we compare exclusively with SL algorithms to highlight our advantages over them.}
    
    \vspace{-1.5mm}
\end{table*}

\begin{table*}[t]
	\centering
	\caption{Sensitivity of ELSA to compression ratio $\rho$: performance and communication benefit (using BERT-base-uncased).}
	\label{tab:performance_analysis}
	\small
	\setlength{\tabcolsep}{4pt} 
	\begin{tabular}{@{}lccccccccc@{}} 
		\toprule
		& \textbf{AG\_News} & \textbf{Banking} & \textbf{Emotion} & \textbf{Trec} & \textbf{RTE} & \textbf{CB} & \textbf{MultiRC} & \textbf{SQuAD} & \textbf{Avg. Benefit} \\
		& Accuracy & Accuracy & Accuracy & Accuracy & Accuracy & Accuracy & F1/EM & EM/F1 & --- \\ \midrule
		
		$\rho =$ 2.1 & 92.33 & 86.34 & 85.22 & 83.33 & 84.24 & 84.18 & 82.45/53.12 & 85.66/87.88 & --- \\
		Comm. Benefit & 1.67 & 1.81 & 1.66 & 1.82 & 1.65 & 1.26 & 1.46 & 1.29 & \textbf{1.58} \\ \midrule
		
		$\rho =$ 3.3 & 91.08 & 84.21 & 85.02 & 83.08 & 83.18 & 84.03 & 81.87/53.01 & 84.13/87.48 & --- \\
		Comm. Benefit & 2.98 & 2.68 & 2.77 & 2.68 & 2.73 & 2.01 & 2.45 & 2.26 & \textbf{2.57} \\ \midrule

        \textcolor{blue}{$\rho$ = 4.2} & \textcolor{blue}{89.58} & \textcolor{blue}{83.55} & \textcolor{blue}{83.90} & \textcolor{blue}{82.20} & \textcolor{blue}{80.93} & \textcolor{blue}{83.93} & \textcolor{blue}{81.78/52.06} & \textcolor{blue}{79.24/87.17} & \textcolor{blue}{---} \\ \textcolor{blue}{Comm. Benefit} & \textcolor{blue}{3.26} & \textcolor{blue}{3.30} & \textcolor{blue}{3.33} & \textcolor{blue}{3.78} & \textcolor{blue}{3.64} & \textcolor{blue}{2.40} & \textcolor{blue}{3.15} & \textcolor{blue}{2.84} & \textcolor{blue}{\textbf{3.21}} \\ \midrule
		
		$\rho =$ 6.4 & 86.33 & 81.03 & 81.11 & 81.15 & 80.25 & 81.36 & 81.37/51.71 & 77.13/86.52 & --- \\
		Comm. Benefit & 3.65 & 3.84 & 3.39 & 4.85 & 4.46 & 3.28 & 3.86 & 3.12 & \textbf{3.81} \\ \midrule
		
		$\rho =$ 8.4 & 83.22 & 75.46 & 78.22 & 74.66 & 75.97 & 73.66 & 73.34/47.22 & 76.22/82.11 & --- \\
		Comm. Benefit & 3.39 & 4.27 & 4.33 & 5.56 & 3.45 & 2.24 & 2.46 & 3.80 & \textbf{3.69} \\ \midrule
		
		$\rho =$ 11.8 & 78.03 & 70.31 & 72.84 & 71.81 & 72.25 & 68.34 & 68.19/43.18 & 72.55/80.06 & --- \\
		Comm. Benefit & 3.98 & 4.57 & 4.72 & 5.82 & 4.98 & 5.90 & 3.77 & 3.87 & \textbf{4.70} \\ \bottomrule
	\end{tabular}
	\vspace{-3mm}
\end{table*}

\subsection{Evaluation of Communication Costs and Transmission Efficiency}

In the following, we evaluate ELSA by analyzing its communication-utility trade-offs, architectural advantages in model segmentation, and the effectiveness of ``SS-OP + Sketch" mechanism in balancing privacy and representation quality, focusing on a setting with $\hat{\alpha} = $ 0.1, $\mathcal{B}_{\text{n}} \in$  [50,100] Mbps.

\noindent \textbf{Analysis of System Utility and Communication Efficiency.}

\begin{figure}[!t]
	\centering
	
	\subfigure[BERT TC tasks]{
		\includegraphics[trim=0.1cm 0.05cm 0.6cm 0cm, clip, width=0.46\columnwidth]{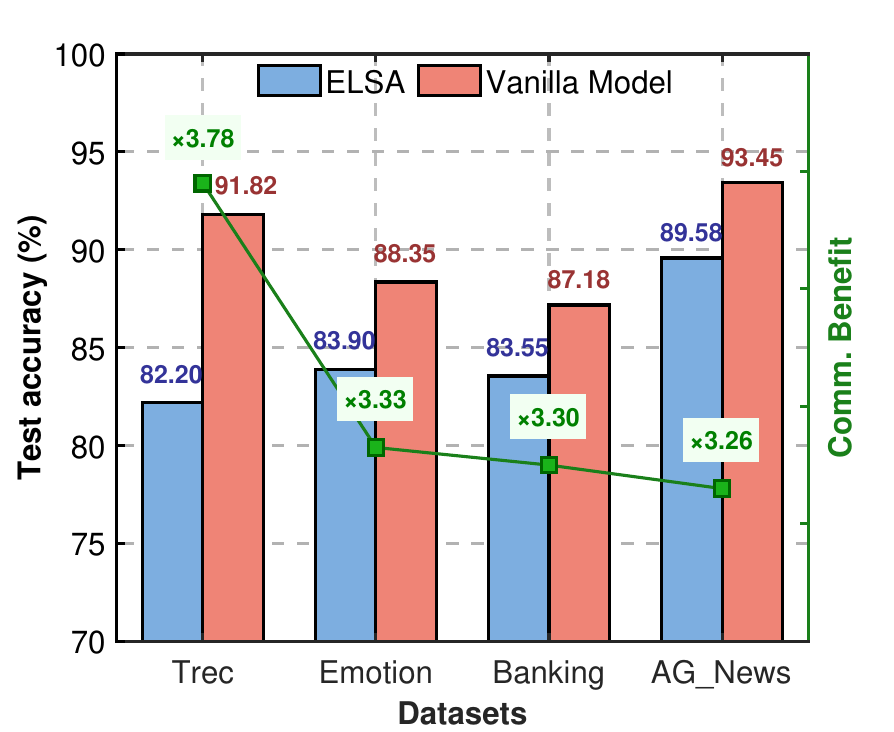}
	}
	\subfigure[BERT NLI tasks]{
		\includegraphics[trim=0.1cm 0.05cm 0.6cm 0cm, clip, width=0.46\columnwidth]{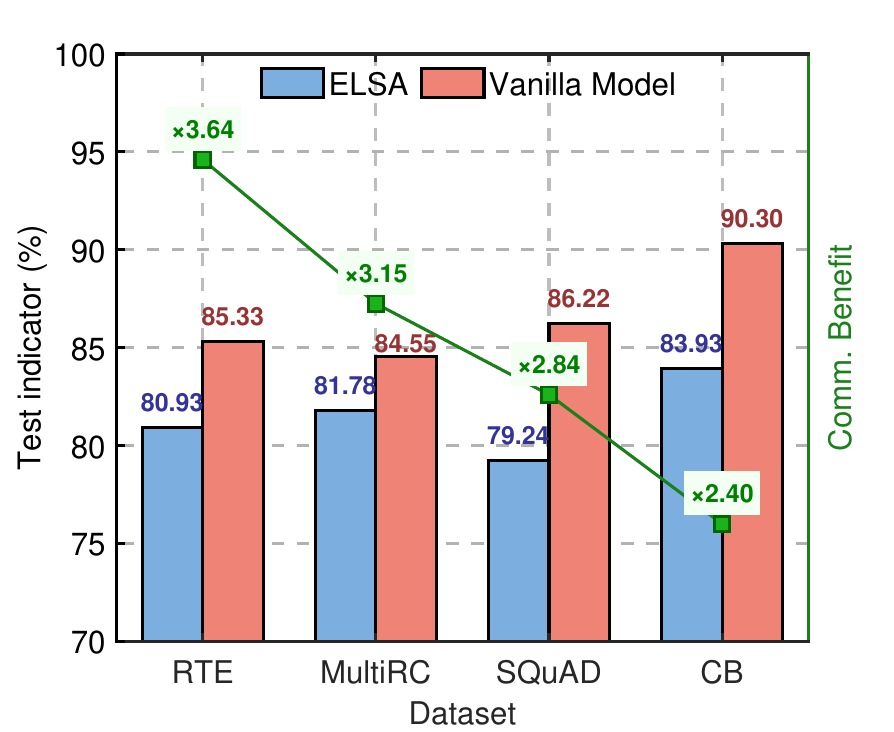}
	}
	\subfigure[Qwen TC tasks]{
		\includegraphics[trim=0.1cm 0.05cm 0.6cm 0cm, clip, width=0.46\columnwidth]{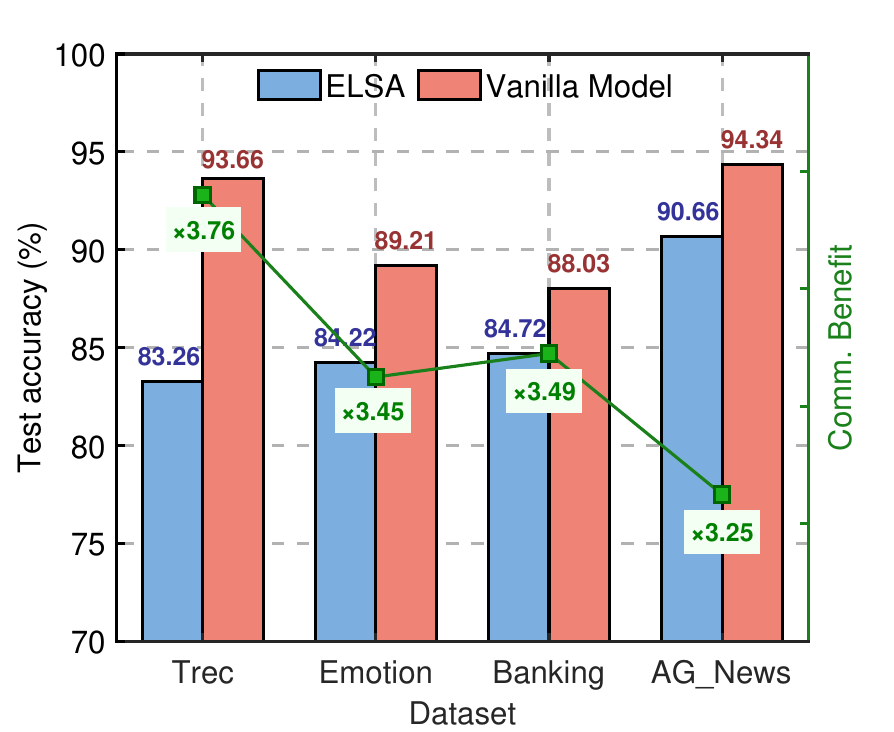}
	}
	\subfigure[Qwen NLI tasks]{
		\includegraphics[trim=0.1cm 0.05cm 0.6cm 0cm, clip, width=0.46\columnwidth]{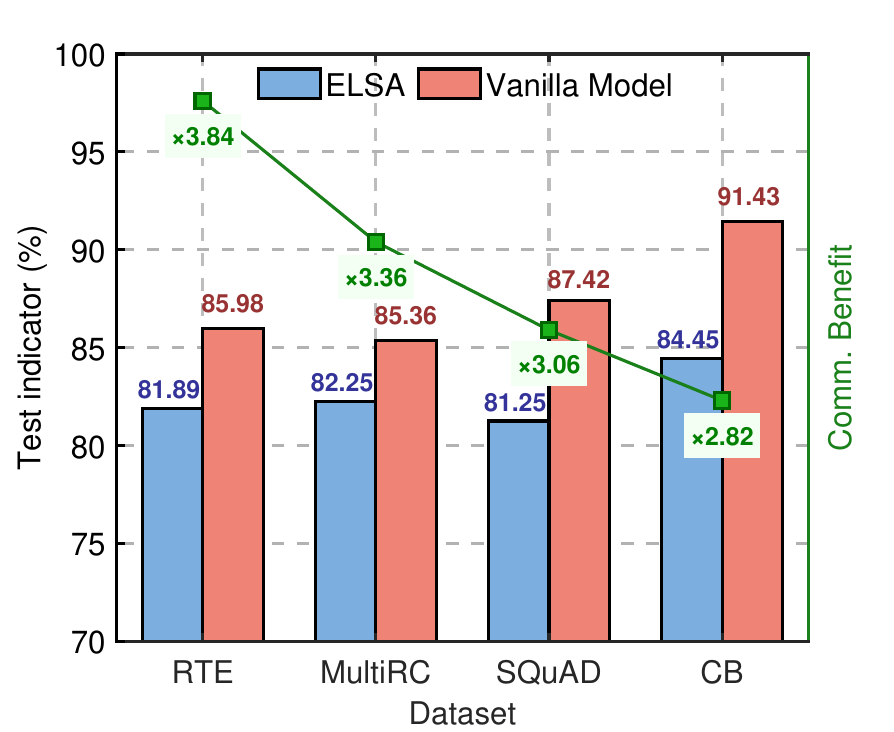}
	}
		
	\vspace{-3mm}
	\caption{\textcolor{blue}{Evaluation performance and communication reduction of ELSA versus uncompressed schemes on various datasets (Indicators are Accuracy (RTE, CB) , plus F1 and EM (MultiRC, SQuAD), compression ratio $\rho$ = 4.2 and $\hat{\alpha} =$ 0.1).}}
	\vspace{-5mm}
	\label{fig:5}
\end{figure}

\noindent In Fig. \ref{fig:5}, we evaluate the practical utility of ELSA by comparing it with an uncompressed Vanilla Model (representing the upper bound of performance without transmission constraints). To achieve comparable performance, the communication volume required by the baseline uncompressed model is 3.26$\times$ to 3.78$\times$ that of ELSA for TC tasks, and up to 3.64$\times$ for NLI tasks. Despite the high compression, ELSA maintains competitive performance, achieving 83.93\% accuracy on CB and 81.78\% on MultiRC with only marginal degradation compared to the uncompressed upper bound. These results validate that our sketch-based layered compression effectively captures essential model updates with high fidelity, making it suitable for resource-constrained edge environments. \textcolor{red}{Furthermore, as the model scale increases from 110 million parameters to the billion-parameter level, ELSA continues to deliver competitive task performance while achieving greater communication benefits. In particular, Qwen3-4B improves the average communication benefit across all datasets by 5.17\% over BERT. This result demonstrates that ELSA remains effective when scaled from relatively compact encoder models to substantially larger decoder-only LLMs.}

Furthermore, we extend our evaluation to the total communication time required to reach a predefined convergence target across eight diverse datasets, as summarized in Table \ref{tab:total_comm_time}, following \eqref{eq:total_comm_time}. Results demonstrate that ELSA consistently achieves the lowest communication latency in all scenarios. Specifically, compared to the uncompressed Vanilla Model, ELSA reduces the total required communication time by approximately 69.3\% to 73.7\% (e.g., from 10.51$\times \text{10}^\text{3}$ s to 3.22$\times \text{10}^\text{3}$ s on AG\_News). \textcolor{blue}{Across the three baseline categories, the general FL methods incur relatively high communication and computation costs because they do not explicitly exploit model splitting or behavior-aware client organization. Aggregation-optimized and edge FL/HFL methods improve update coordination, but their communication time remains higher because aggregation optimization alone does not jointly address intermediate-representation transmission and client--edge association. The LLM-oriented SL methods achieve stronger system efficiency by
distributing model execution. Nevertheless, compared with RoFed, the strongest SL baseline in terms of communication time on these two datasets, ELSA reduces the total communication time to convergence by 0.29$\times$10$^\text{3}$~s on MultiRC (from 13.44$\times$10$^\text{3}$~s to 13.15$\times$10$^\text{3}$~s) and by 0.85$\times$10$^\text{3}$~s on SQuAD (from 13.08$\times$10$^\text{3}$~s to 12.23$\times$10$^\text{3}$~s). Moreover, ELSA achieves an
average computation latency of only 8.76$\times$10$^\text{3}$~s across the eight datasets, representing a 9.5\% reduction over Split-LLM
(9.68$\times$10$^\text{3}$~s), the most computation-efficient competing SL method, which benefits from the synergy of ELSA's behavior-aware clustering, dynamic three-part splitting, and compressed cross-tier transmission. Besides, ELSA also achieves the lowest average computation latency (8.76 $\times \text{10}^\text{3}$~s), compared with 9.68 $\times \text{10}^\text{3}$~s for Split-LLM, the most efficient SL baseline. Its average task failure rate is 1.2\%, versus 2.4\%--5.2\% for the SL baselines, indicating greater training reliability under heterogeneous client resources and network conditions alongside reduced communication and computation costs.}

Next, in Table~\ref{tab:performance_analysis}, we present the system performance and communication benefit (it's measured as the ratio of total communication volume between Vanilla Model and ELSA under identical convergence conditions) of ELSA under different compression strengths $\rho $ to characterize its sensitivity to compression. We observe that there is a non monotonic relationship between overall communication advantage and compression strength. A lower compression ratio (e.g. $\rho=$ 2.1) can ensure high representation fidelity and stable convergence, achieving an accuracy of 92.33\% on AG\_News, but the communication burden per training round is significant. While excessively high compression ratios (e.g. $\rho=$ 11.8) can significantly reduce the amount of data transmitted per round, but can lead to a significant decrease in accuracy (e.g. 78.03\% on AG\_News). \textit{How to choose the value of $\rho$?}
Table~\ref{tab:performance_analysis} illustrates the fundamental tradeoff between communication efficiency and model performance induced by the compression ratio $\rho$. As $\rho$ increases, ELSA achieves larger communication benefits (up to a 4.70 $\times$ average reduction at $\rho=$ 11.8) but this comes with a noticeable degradation in accuracy and F1/EM across tasks. For small to moderate compression levels ($\rho=$ 2.1 and $\rho=$ 3.3), ELSA preserves near-peak performance while already providing meaningful communication savings (1.58$\times$--2.57$\times$ on average). Beyond this regime, performance drops accelerate, particularly for semantically complex tasks such as MultiRC and SQuAD, indicating that excessive compression increasingly distorts informative hidden representations. Overall, these results suggest that $\rho \in$ [2.1,\,4.2] offers a practical operating region that balances communication efficiency and learning performance, while larger $\rho$ values should be reserved for severely bandwidth-constrained scenarios.
\begin{table*}[!t]
	\centering
	\caption{\textcolor{blue}{Comparison of resource utilization and task failure rate between static and dynamic splitting strategies across BERT-base-uncased and Qwen3-4B. Under a heterogeneous network where 40\% of clients are resource-constrained, a training task is marked as ``Failed'' if the iteration latency exceeds the system timeout threshold.}}
	\label{tab:splitting_comparison}
	\small
	\renewcommand{\arraystretch}{1.05} 
	\begin{adjustbox}{max width=\textwidth}
		\begin{tabular}{clcccc} 
			\toprule
			\textbf{Model} & \textbf{Strategy} & \textbf{Avg. Comp. Util. (\%)} & \textbf{Avg. Comm. Util. (\%)} & \textbf{Overall Eff. (\%)} & \textbf{Task Failure Rate (\%)} \\ 
			\midrule
			\multirow{5}{*}{\textbf{\makecell{BERT-base\\-uncased}}} 
			& Static (Aggressive, $p=1$)  & 32.4 & \textbf{88.5} & 54.2 & 2.1 \\
			& Static (Balanced A, $p=3$)  & 45.8 & 76.3          & 58.5 & 8.5 \\
			& Static (Balanced B, $p=6$)  & 61.2 & 54.1          & 56.8 & 24.3 \\
			& Static (Conservative, $p=9$)& \textbf{78.5} & 31.2 & 48.6 & 41.7 \\ 
			\cmidrule(lr){2-6}
			& \textbf{Ours ELSA (Dynamic)}     & \textbf{72.3} & \textbf{79.8} & \textbf{84.6} & \textbf{1.2} \\ 
			\midrule
			\multirow{5}{*}{\textbf{Qwen3-4B}} 
			& Static (Aggressive, $p=3$)  & 35.1 & \textbf{91.2} & 52.4 & 2.8 \\
			& Static (Balanced A, $p=10$) & 48.6 & 78.5          & 60.1 & 9.2 \\
			& Static (Balanced B, $p=20$) & \textbf{78.5} & 51.2          & 54.3 & 28.6 \\
			& Static (Conservative, $p=30$)& \textbf{84.2} & 28.4 & 45.2 & 48.5 \\ 
			\cmidrule(lr){2-6}
			& \textbf{Ours ELSA (Dynamic)}     & \textbf{74.8} & \textbf{81.6} & \textbf{86.2} & \textbf{1.8} \\ 
			\bottomrule
		\end{tabular}
	\end{adjustbox}
	\vspace{-3mm}
\end{table*}

\begin{table*}[!t]
	\centering
	\caption{Comparison of privacy and utility metrics under different compression rates $\rho$.}
	\label{tab:comparison_updated}
	\small
	\setlength{\tabcolsep}{5pt} 
	\begin{tabular}{lccccccccc}
		\toprule
		\multirow{2}{*}{\textbf{Method}} & \multicolumn{3}{c}{$\rho$ = 2.1} & \multicolumn{3}{c}{$\rho$ = 4.2} & \multicolumn{3}{c}{$\rho$ = 8.4} \\
		\cmidrule(lr){2-4} \cmidrule(lr){5-7} \cmidrule(lr){8-10}
		& Cos Sim & MSE & Token Acc. & Cos Sim & MSE & Token Acc. & Cos Sim & MSE & Token Acc. \\
		\midrule
		Direct Transmission & 1.0000 & 0.0000 & 53.13\% & -- & -- & -- & -- & -- & -- \\
		Gaussian Noise      & 0.9137 & 0.0617 & 50.11\% & -- & -- & -- & -- & -- & -- \\
		\midrule
		Sketch Only & 0.3599 & 2.1509 & 4.07\%  & 0.2655 & 4.1584 & 1.96\%  & 0.1975 & 7.0615 & 1.13\% \\
		\textbf{ELSA ($r=$ 8)} & \textbf{-0.0334} & \textbf{2.6979} & \textbf{1.21\%}  & \textbf{-0.0244} & \textbf{4.7059} & \textbf{0.38\%} & \textbf{-0.0220} & \textbf{7.7735} & \textbf{0.09\%}\\
		\textbf{ELSA ($r=$ 16)} & \textbf{-0.0167} & \textbf{2.7669} & \textbf{0.45\%}  & \textbf{-0.0122} & \textbf{4.7046} & \textbf{0.08\%} & \textbf{-0.0078} & \textbf{7.7492} & \textbf{0.00\%}\\
		\bottomrule
	\end{tabular}
	
	\vspace{0.5em}
	\scriptsize
	\textit{Note:} ``--'' indicates that the method is independent of the compression ratio $\rho$ (i.e., no compression is applied).
	\vspace{-5mm}
\end{table*}

\noindent \textbf{Impact of Model Segmentation.}
\noindent \textcolor{blue}{We evaluate the performance of ELSA by varying the number of transformer blocks trained locally on the client side, specifically, under the 12-layer BERT-base architecture, we evaluate $p \in$ \{1,3,6,9\}, with corresponding $(q, o)$ pairs of (9, 2), (7, 2), (4, 2) and (1,2)}\footnote{In our experiments, $o$ was empirically set to 2, but it has no bearing on the overall findings since our analysis focuses on the relative changes in $p$.}, respectively, ensuring the total number of transformer blocks remains 12. As illustrated in the radar chart in Fig. 6(a), a significant performance degradation is observed as the number of local blocks increases to 6. We attribute this phenomenon to a critical factors inherent in SL paradigms. Although our hierarchical aggregation mechanism, to some extent, can avoid local optima of the model at the client end, upon having large number of training layers, it still has a certain impact, that is, the feature extractor becomes ``too personalized" to the local (usually non-IID) data distribution and cannot capture a universal semantic representation. \textcolor{blue}{Furthermore, these findings suggest that maintaining a client-side training depth of fewer than 6 Transformer blocks is sufficient to yield robust results. For Qwen3-4B, we consider $p\in$ \{3,10,20,30\} and observe behavior consistent with BERT, indicating that this phenomenon is not specific to a particular Transformer architecture or model scale. Building on this insight, for BERT, we configure our \textit{resource-aware dynamic splitting} strategy with an upper bound $p_{\max}=$ 6 to prevent performance degradation from over-personalization (for Qwen3-4B, the upper bound is $p_{\max}=$ 20), while allowing $p_n$ to adaptively decrease for clients with better resources.}

Table~\ref{tab:splitting_comparison} illustrates the performance trade-offs between fixed splitting configurations and our proposed dynamic strategy in a heterogeneous environment. 
Static strategies face a dilemma: \textcolor{blue}{aggressive offloading (e.g., $p=$ 1 for BERT-base and $p=$ 3 for Qwen3-4B) incurs high communication overhead and underutilizes client computing resources. Conversely, conservative local training (e.g., $p=$ 9 and $p=$ 30, respectively) creates computation bottlenecks for resource-constrained clients, resulting in low overall efficiencies of 48.6\% and 45.2\% and high task failure rates of 41.7\% and 48.5\%. In contrast, our \textit{resource-aware dynamic splitting} adaptively assigns $p_n$ based on each client's specific computational and network profile. 
This approach achieves the highest overall efficiency (84.6\%, 86.2\%) while maintaining the lowest failure rate (1.2\%, 1.8\%), demonstrating its ability to effectively balance system efficiency and robustness against device heterogeneity\footnote{\textit{Comp. Util.} denotes the average percentage of client computational capacity (FLOPS) actively engaged in training; \textit{Comm. Util.} represents the average fraction of available network bandwidth effectively used for data transmission; \textit{Overall Eff.} is a composite score evaluating the joint efficiency of computation and communication resources; \textit{Task Failure Rate} indicates the proportion of clients unable to complete an iteration within the system timeout limit, reflecting robustness against stragglers.}. In Fig. 6, while the static configuration with minimal local layers (e.g., $p=$ 1) achieves the marginal highest accuracy in ideal conditions, it incurs prohibitive communication costs and high failure rates as discussed in Table~\ref{tab:splitting_comparison}.  In contrast, for BERT, our dynamic approach surpasses both the $p=$ 3 and $p=$ 6 static baselines on average, while significantly outperforming the conservative $p=$ 9 setting. Specifically, our method closely tracks the performance of the $p=$ 3 configuration on most datasets (e.g., SQuAD, MultiRC, CB) while occasionally approaching the $p=1$ peak on others (e.g., AG\_News and Emotion). This demonstrates that by capping the local depth at the empirically determined threshold (e.g.,  $p_{\max}=$ 6) and dynamically offloading the rest, ELSA successfully navigates the trade-off: it retains the generalization benefits of shallow local training (e.g., avoiding the $p=$ 9 drop) while accommodating heterogeneous resource constraints without the instability of aggressive static offloading (e.g., $p=$ 1)\footnote{\textcolor{red}{Deeper local partition depths ($p \ge $ 6) assigned to low-bandwidth clients under Non-IID settings may induce minor local overfitting, presenting a trade-off between communication efficiency and device-level fairness. We elaborate on this phenomenon, present a preliminary mitigation scheme, and provide full empirical validations in \textbf{Appendix~K.}}}.}

\begin{table*}[!t]
	\centering
	\caption{\textcolor{blue}{Ablation study of the ELSA framework on eight NLP datasets under $\hat{\alpha}=$ 0.1 and BERT-base-uncased.}}
	\label{tab:elsa_ablation}
	\small
	\renewcommand{\arraystretch}{1.05}
	\setlength{\tabcolsep}{3.2pt}
	
	\begin{adjustbox}{max width=\textwidth}
		\begin{tabular}{@{} l *{8}{c} c c c @{}}
			\toprule
			
			\multicolumn{1}{c}{\multirow{2}{*}{\textbf{Variant}}}
			& \multicolumn{4}{c}{\textbf{TC Tasks (Acc.\,\%)}}
			& \multicolumn{4}{c}{\textbf{NLI Tasks (Acc.\,\%, F1 and EM)}}
			& \multicolumn{1}{c}{\multirow{2}{*}{\textbf{\shortstack{Client\\Comp.}}}}
			& \multicolumn{1}{c}{\multirow{2}{*}{\textbf{\shortstack{Comm.\\/ Lat.}}}}
			& \multicolumn{1}{c}{\multirow{2}{*}{\textbf{\shortstack{Token Rec.\\Acc.$^\dagger$ (\%)}}}} \\
			\cmidrule(lr){2-5} \cmidrule(lr){6-9}
			
			\multicolumn{1}{c}{}
			& \textbf{Trec} & \textbf{Emotion} & \textbf{Banking} & \textbf{AG\_News}
			& \textbf{RTE} & \textbf{MultiRC} & \textbf{SQuAD} & \textbf{CB}
			& \multicolumn{1}{c}{}
			& \multicolumn{1}{c}{}
			& \multicolumn{1}{c}{} \\
			\midrule
			
			ELSA (Full)
			& \textbf{82.20} & \textbf{83.90} & \textbf{83.55} & \textbf{89.58}
			& \textbf{80.93} & \textbf{81.78} & \textbf{79.24} & \textbf{83.93}
			& 1.82\,TFLOPs & 14.2\,MB/1.12\,s & $<$ 1.50 \\
			\addlinespace[0.3em]
			
			\multicolumn{12}{@{}l}{\textit{1.\;Behavior-Aware \& Selection}} \\
			\addlinespace[0.2em]
			
			w/o Behavior Fingerprint (Ran.)
			& 81.45 & 81.02 & 78.10 & 86.50
			& 78.48 & 78.23 & 76.55 & 81.25
			& 1.82\,TFLOPs & 14.2\,MB/1.25\,s & $<$ 1.50 \\
			
			w/o Trust Scoring (Noise Sensitive)
			& 80.10 & 82.65 & 79.30 & 87.12
			& 75.20 & 76.10 & 73.42 & 77.80
			& 1.82\,TFLOPs & 14.2\,MB/1.20\,s & $<$ 1.50 \\
			\addlinespace[0.3em]
			
			\multicolumn{12}{@{}l}{\textit{2.\;Dynamic Model Splitting Strategy}} \\
			\addlinespace[0.2em]
			
			w/o Dynamic Splitting (Static $p = $6)
			& 79.55 & 83.01 & 81.78 & 89.02
			& 80.01 & 80.10 & 78.89 & 81.03
			& 3.64\,TFLOPs & 16.8\,MB/1.68\,s & $<$ 1.50 \\
			\addlinespace[0.3em]
			
			\multicolumn{12}{@{}l}{\textit{3.\;Sketching \& SS-OP }} \\
			\addlinespace[0.2em]
			
			w/o SS-OP (Sketch Only)
			& 82.12 & 83.88 & 83.54 & 89.55
			& 80.90 & 81.73 & 79.20 & 83.84
			& 1.82\,TFLOPs & 14.2\,MB/1.08\,s & 10.76 \\
			
			w/o Sketching (SS-OP Only)
			& \underline{91.82} & \underline{88.35} & \underline{87.18} & \underline{93.45}
			& \underline{85.33} & \underline{84.55} & \underline{86.22} & \underline{90.30}
			& 1.82\,TFLOPs & 52.8\,MB/2.95\,s & 4.33 \\
			
			\addlinespace[0.3em]
			
			\multicolumn{12}{@{}l}{\textit{4.\;Coherence \& Trust-Weighted}} \\
			\addlinespace[0.2em]
			
			w/o Coherence Weighting
			& 80.30 & 81.80 & 79.96 & 85.85
			& 79.65 & 80.40 & 77.95 & 82.85
			& 1.82\,TFLOPs & 14.2\,MB/1.12\,s & $<$ 1.50 \\
			
			\bottomrule
			\multicolumn{12}{@{}l}{\scriptsize \textcolor{blue}{\textbf{Bold} indicates the best performance, while \underline{underline} denotes the unconstrained upper bound (w/o Sketching)}} \\
			\multicolumn{12}{@{}l}{\scriptsize \textcolor{blue}{$^\dagger$DLG attack token recovery accuracy; higher value indicates greater privacy leakage.}} \\
			\multicolumn{12}{@{}l}{ \scriptsize
			\textcolor{blue}{Performance varies across different static slicing layers (it may be better or worse); therefore we focus on the residual performance (see Table V in the revised manuscript).}} \\
		\end{tabular}
	\end{adjustbox}
    \vspace{-5mm}
\end{table*}

\subsection{\textcolor{red}{Evaluation of Component Contributions and Ablation Analysis}}
\textcolor{red}{To quantify the individual contribution and overhead of each core component in ELSA, we conduct comprehensive ablation experiments using BERT across eight NLP benchmarks under non-IID conditions ($\hat{\alpha}=0.1$). As summarized in Table~\ref{tab:elsa_ablation}, each mechanism plays a crucial role in balancing system efficiency, global accuracy, and privacy preservation. \textit{(i) Behavior-Aware Clustering and Trust Selection}: As shown in Table~\ref{tab:elsa_ablation}, task-agnostic behavioral fingerprinting and trust scoring incur zero additional client-side compute (maintaining 1.82\,TFLOPs) and communication payload (14.2\,MB). However, omitting behavior fingerprinting (\textit{w/o Behavior Fingerprint}) slows down model convergence, increasing total latency by 11.6\% (from 1.12\,s to 1.25\,s) and causing a 0.75\%--5.45\% performance degradation across tasks. Furthermore, removing trust scoring (\textit{w/o Trust Scoring}) exposes the system to noisy clients, severely harming accuracy on complex tasks (e.g., dropping SQuAD EM score from 79.24\% to 73.42\%). This validates that behavioral clustering significantly accelerates convergence and enhances robustness at near-zero overhead. \textit{(ii) Dynamic Model Splitting Strategy}: Replacing our dynamic layer splitting with a fixed allocation (e.g., \textit{Static $p=$ 6}) doubles client-side computation from 1.82\,TFLOPs to 3.64\,TFLOPs and increases per-round latency by 50.0\% (from 1.12\,s to 1.68\,s). While aggressive static slicing ($p=$ 1) reduces client compute, it severely congests the network backhaul; conversely, conservative static splits ($p=$ 9) overload resource-constrained edge devices, causing high task failure rates. ELSA's dynamic splitting adaptively balances compute and communication across heterogeneous devices, achieving superior task completion rates (see Table \ref{tab:splitting_comparison}). \textit{(iii) Activation Sketching and SS-OP Perturbation}: Omitting Sketching (\textit{w/o Sketching (SS-OP Only)}) expands the per-round upload payload by 271.8\% (from 14.2\,MB to 52.8\,MB) and nearly triples the round latency (from 1.12\,s to 2.95\,s), while Sketching encoding introduces negligible computational overhead on clients (maintaining 1.82\,TFLOPs). On the other hand, removing SS-OP cryptographic protection (\textit{w/o SS-OP (Sketch Only)}) has negligible effects on latency (1.08\,s) and accuracy, but causes the DLG attack token recovery rate to surge from $<$ 1.50\% to 10.76\%. This highlights SS-OP as a highly cost-effective safeguard against activation reconstruction attacks. \textit{(iv) Coherence and Trust-Weighted Aggregation}: Since coherence weighting operates entirely on the server side, it introduces zero client-side compute or bandwidth overhead. Under severe non-IID data distributions, removing this mechanism (\textit{w/o Coherence Weighting}) leads to consistent accuracy drops across all datasets (e.g., Trec accuracy falls from 82.20\% to 80.30\%), proving its efficacy in mitigating semantic drift during aggregation. Collectively, the ablation results demonstrate that ELSA's performance gains are not driven by any single component, but rather stem from the tight synergy between sketching-based efficiency, behavior-aware grouping, and privacy-preserving mechanisms.}

\subsection{Evaluation of Data Desensitization and Privacy Protection}
\textcolor{blue}{Table~\ref{tab:comparison_updated} evaluates the trade-off between privacy and utility in our SS-OP + Sketch under varying compression ratios. To simulate potential attacks on intermediate activations during client–edge transmissions, we conduct two levels of evaluation: basic static recoverability analysis (Table~\ref{tab:comparison_updated}) and advanced active attack evaluation against stronger threat models (Fig.~\ref{fig:strong_attacks}). We compare against three baselines:~\textit{(i)} Direct Transmission (no protection), \textit{(ii)} Gaussian Noise (adding calibrated noise for differential privacy $\widetilde{\mathcal{N}}$(0, 0.25), and \textit{(iii) }Sketch Only (compression without orthogonal perturbation). Detailed attack formulations and hyperparameter configurations are provided in \textbf{Appendix J}.}

\noindent \textcolor{blue}{\textbf{Static Recoverability Analysis:} Following \cite{SentenceEm,InformationL}, we first assess basic recoverability via \textit{(i)} Reconstruction attacks by a semi-honest server, measured via cosine similarity (Cos Sim) and mean squared error (MSE) between original and reconstructed hidden states; \textit{(ii)} Token identification attacks, where an adversary attempts to infer input tokens from compressed representations, evaluated by token recovery accuracy (Token Acc.).} As shown in Table~\ref{tab:comparison_updated}, Direct Transmission is highly vulnerable (Cos Sim = 1.0, Token Acc. = 53.13\%), while Gaussian Noise offers only marginal protection (Cos Sim = 0.9137). Sketch Only reduces token recovery (e.g., 1.13\% at $\rho=$ 8.4) but retains exploitable semantic structure, as indicated by its positive cosine similarity. In contrast, ELSA integrates SS-OP with sketching, driving cosine similarity to near or below zero and suppressing token accuracy to near-zero levels. Notably, increasing the semantic subspace dimension from $r=$ 8 to $r=$ 16 further reduces token identification accuracy (e.g., from 0.38\% to 0.08\% at $\rho=$ 4.2) while maintaining similarly low cosine similarity. This suggests that, under appropriate compression ratios, a larger $r$ enhances privacy by perturbing a richer set of semantic directions without degrading utility. The slight MSE increase over Sketch Only is benign for training, confirming that ELSA\footnote{Note that while these modules introduce additional operations, their computational cost is designed to be lightweight relative to the LLM forward/backward pass (see \textbf{Appendix L} for detailed complexity analysis).} achieves strong privacy without sacrificing utility.

\begin{figure}[!t]
	\centering
	\subfigure[]{
		\includegraphics[trim=0cm 0.05cm 0cm 0cm, clip, width=0.46\columnwidth]{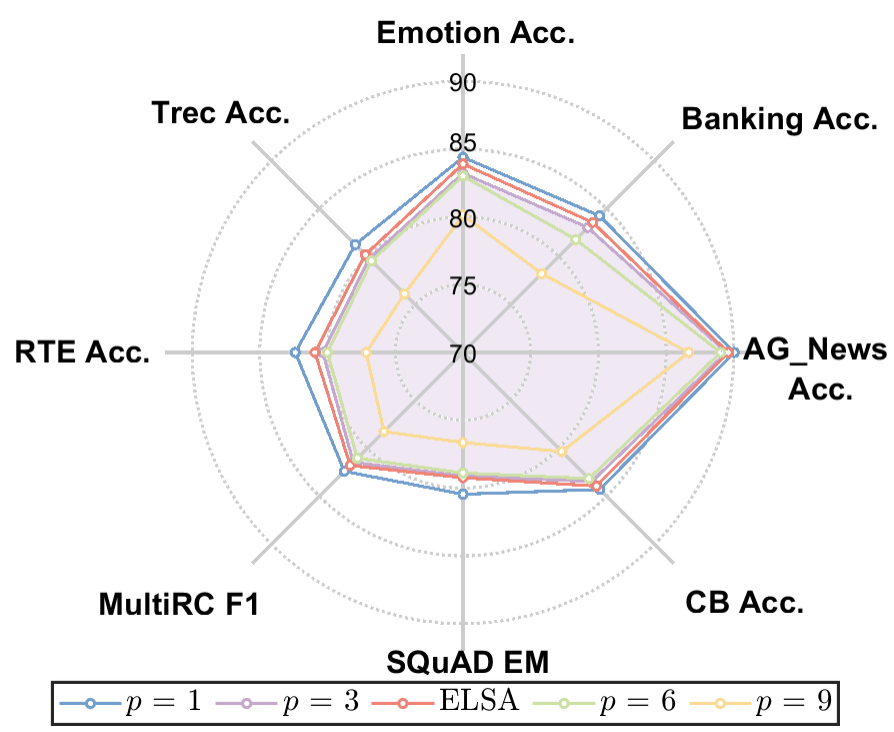}
		}
	\subfigure[]{
		\includegraphics[trim=0.1cm 0.05cm 0cm 0cm, clip, width=0.46\columnwidth]{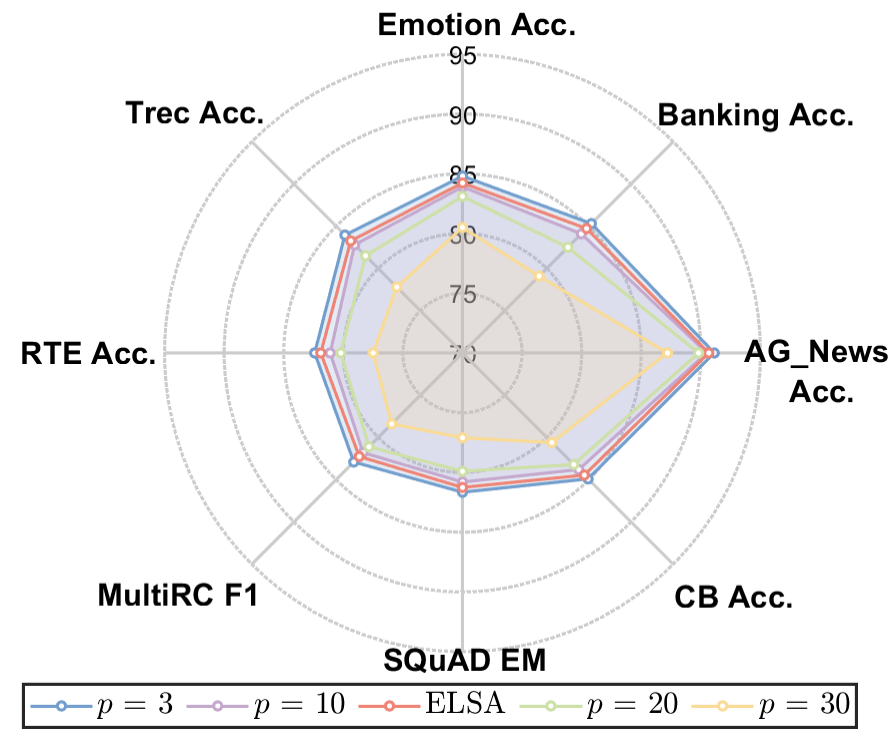}
	}
	
	\caption{\textcolor{blue}{Effects of collaborative training schemes on model performance (Left subplot: BERT, Right subplot: Qwen3-4B).}}
	\vspace{-3mm}
	\label{fig:leftright}
\end{figure}

\begin{figure}[htbp]
	\centering
	\subfigure{
		\includegraphics[trim=0.63cm 0cm 0.7cm 0cm, clip, width=0.93\columnwidth]{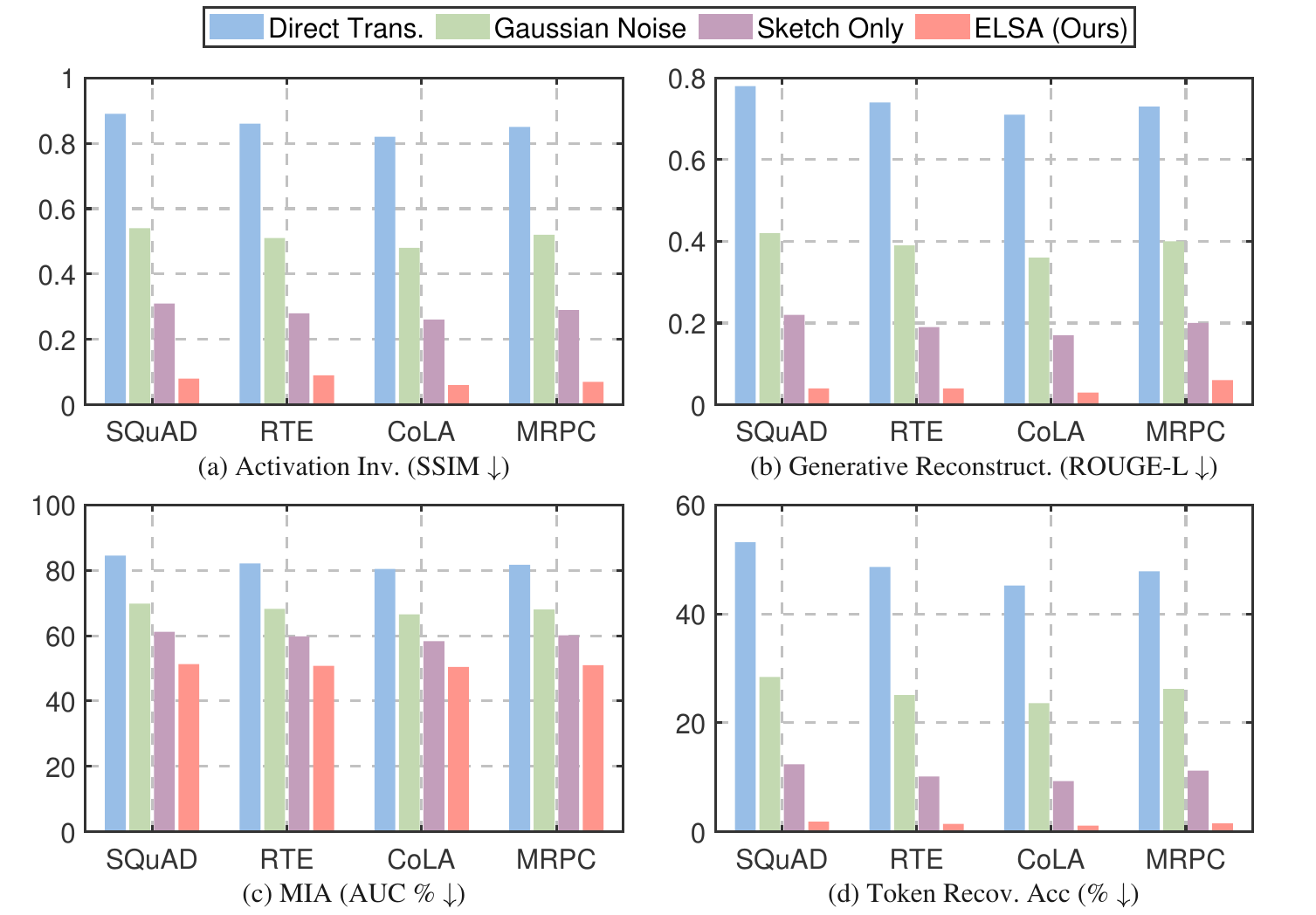}
	}
	\caption{\textcolor{blue}{Privacy protection performance under stronger attack baselines across multiple datasets.}}
	\label{fig:strong_attacks}
	\vspace{-3mm}
\end{figure}

\noindent \textcolor{red}{\textbf{Defense Against Stronger Attack Baselines:} To rigorously validate privacy robustness, we extend our evaluation across four NLP tasks (SQuAD, RTE, CoLA\footnote{\textcolor{red}{https://nyu-mll.github.io/CoLA/}}, and MRPC\footnote{\textcolor{red}{\url{https://huggingface.co/datasets/SetFit/mrpc}}}) under three state-of-the-art attack paradigms: \textit{(i) Activation Inversion Attack (DLG/iDLG)}: An optimization-based adversary attempting to reconstruct input embeddings via gradient matching (Fig. \ref{fig:strong_attacks}(a)). \textit{(ii) Generative Embedding Inversion (Vec2Text)}: A generative adversary using a pre-trained language model to decode token sequences from intermediate activations (Fig. \ref{fig:strong_attacks}(b)). \textit{(iii) Membership Inference Attack (MIA)}: A shadow-model-based adversary determining whether a given sample belonged to the client's training set (Fig. \ref{fig:strong_attacks}(c)). As illustrated in Fig.~\ref{fig:strong_attacks}, ELSA consistently achieves superior defense capabilities across all tasks. Under Activation Inversion and Generative Reconstruction attacks, ELSA significantly reduces the SSIM and ROUGE-L scores compared to Direct Transmission and Gaussian Noise baselines. Furthermore, ELSA suppresses the Token Recovery Accuracy to $<$ 2.0\% under DLG attacks (Fig. \ref{fig:strong_attacks}(d)) and reduces the MIA AUC to nearly 50.0\% (equivalent to random guessing). These empirical findings indicate that SS-OP effectively obfuscates semantic representations and provides strong empirical resistance under the evaluated generative and optimization-based attack settings.}

\section{Conclusion}
We proposed ELSA, a privacy-aware and efficient HFL framework designed to optimize distributed LLM fine-tuning in heterogeneous and resource-constrained edge environments. By synergistically integrating SL with a hierarchical architecture, ELSA addresses the coupled challenges of computational fragmentation, data heterogeneity, and communication overhead. Extensive evaluations across eight diverse NLP benchmarks, demonstrated that ELSA consistently outperforms state-of-the-art baselines in both learning efficiency and final performance. \textcolor{blue}{Future research directions include the extension of ELSA to ultra-large-scale models scaling from 4B+ edge-friendly variants up to 10B+ parameter architectures (including MoE backbones). Furthermore, incorporating knowledge distillation and adaptive straggler mitigation strategies could further alleviate heterogeneous device delays and enhance client-side parameter efficiency. In addition, joint intelligent mechanisms for model segmentation and communication compression can be explored to enable dynamic optimization of split depth and compression ratios in response to more severe real-time edge resource availability and network dynamics.} 

\bibliographystyle{ieeetr}
\bibliography{reference.bib}

\clearpage
\section*{Appendix}

\subsection{\textcolor{red}{Comparison with Clustered FL Frameworks \& Heterogeneity Metrics}}
\label{sec:appendix_clustering_comparison}
\textcolor{red}{Client clustering and heterogeneity measurement are foundational techniques in FL/HFL to mitigate non-IID data degradation. However, extending existing frameworks to resource-constrained LLM fine-tuning at the network edge reveals fundamental structural limitations. Below, we systematically review and contrast representative clustering schemes with ELSA across three major paradigms.}

\textcolor{red}{\textit{(i) Model- and Parameter-Space Clustering Paradigms.} Existing works heavily rely on parameter updates, gradient trajectories, or model weight snapshots to cluster clients with similar representations:  Specifically, CFL-GP \cite{CFLGP} executes SVD and $K$-means spectral clustering on concatenated multi-model gradients under a rigid, predefined cluster count $K$; Fed-CM \cite{FedCM} tracks personalized layer gradient trajectories to construct cosine-similarity graphs; and FedClust \cite{FedClust} performs one-shot hierarchical clustering on classification layer weights following initial full-model warm-up iterations. These frameworks fundamentally mandate hosting or fine-tuning the complete model architecture on edge devices merely to extract clustering signals, a requirement that is computationally prohibitive in LLM scenarios. Furthermore, relying on static snapshots or optimization dynamics in parameter space fails to capture intrinsic data-processing behaviors, rendering them vulnerable to gradient noise and temporal distribution drift. \textit{In sharp contrast, ELSA eliminates full-model client dependencies, extracting behavioral clustering signals exclusively via lightweight public probes without requiring clients to maintain or fine-tune full LLM parameters.}}

\textcolor{red}{\textit{(ii) Label Distribution and Divergence-Based Metrics.} Another line of research employs KLD or label distribution statistics to measure data heterogeneity:
HT-FL \cite{HTFL}: Utilizes standard KLD as a statistical divergence metric to quantify categorical label skewness across client partition tiers; 
KL-FedDis \cite{KLFedDis}: Formulates a KLD-driven metric on categorical histogram distributions to guide client partition selection under non-IID settings; 
AdaptiveHFL \cite{AdaptiveUAV}: Adopts ground-truth class occurrence probabilities to estimate statistical data heterogeneity for client association and UAV deployment. While these methods target vision or classification tasks where scalar label histogram differences sufficiently capture domain variance. In complex natural language understanding (NLU) and generative tasks, identical label distributions often mask profound semantic divergence, contextual nuance, and feature-space heterogeneity. Label-centric metrics fail to capture such high-dimensional behavioral differences. Furthermore, existing KLD and distribution metrics operate in isolation, they ignore physical network conditions (e.g., edge-to-client propagation delay and geographic proximity) as well as device trustworthiness (e.g., mitigating data poisoning or noisy updates).}

\textcolor{red}{\noindent\textit{(iii) Inference- and Knowledge-Based Client Organization.}
Beyond parameter- and label-based clustering, FLIS \cite{FLIS} organizes
clients according to inference similarity derived from predictive outputs
or loss patterns, thereby mitigating non-IID data effects in conventional
FL. FedTKD \cite{FedTKD} instead employs adaptive knowledge distillation
and trustworthy evaluation to coordinate heterogeneous local models. These
methods provide useful mechanisms for organizing heterogeneous clients, but
they assume conventional FL settings in which clients execute complete
local models. Consequently, they do not directly address client organization
under client--edge--client split execution or use clustering decisions to
jointly guide model partitioning, client--edge association, and hierarchical
aggregation. In contrast, ELSA constructs behavioral fingerprints from
intermediate representations generated by shared public probes and uses
the resulting clusters throughout its split and hierarchical training
workflow.}

\textcolor{red}{\textit{(iv) Summary of ELSA's Key Differentiators.} Compared with the aforementioned literature, ELSA bridges the gap between learning quality, system efficiency, and environmental robustness through three core designs:}
\begin{itemize}
	\item \textcolor{red}{\textit{Task-Agnostic Behavioral Fingerprinting}: By leveraging intermediate smashed representations generated by a shared public probe, combined with Gaussian moment extraction and Symmetric KLD, ELSA directly captures behavioral and semantic feature distributions without exposing raw private data.}
	
	\item \textcolor{red}{\textit{Dynamic Adaptability}: ELSA supports periodic, low-overhead re-clustering and intra-cluster hierarchical aggregation, continuously adapting to data distribution drift and time-varying wireless topologies.}
	
	\item \textcolor{red}{\textit{Joint Trust and Network Proximity Co-Design}: Beyond statistical distribution measurement, ELSA explicitly integrates device reliability scoring (filtering poisoned/unreliable clients) and physical distance/latency parameters into the cluster formation process, achieving robust and communication-efficient edge learning.}
\end{itemize}

\textcolor{red}{In summary, while traditional CFL frameworks and metrics decouple communication, learning, and topology into isolated sub-problems, ELSA redefines client clustering in edge LLM fine-tuning. By anchoring clustering decisions to lightweight behavioral fingerprints rather than prohibitive full-model gradients or superficial label statistics, ELSA seamlessly unifies semantic heterogeneity awareness, physical network proximity, and client trustworthiness into a holistic, split-learning-compatible paradigm.}

\subsection{\textcolor{red}{Empirical and Theoretical Analysis of Behavioral Probe Sets}}
\label{app:probe_sets}
\textcolor{red}{To validate the reliability, privacy guarantees, and robustness of the public probe set utilized in our behavior-aware clustering mechanism, this section presents detailed theoretical analyses and empirical evaluations addressing domain mismatch, privacy considerations, cross-task comparability, probe parameter sensitivity and alternative similarity metrics.}

\textcolor{red}{\textit{(i) Addressing Probe-Domain Mismatch and Sensitivity to Probe Size and Composition.} To investigate whether client clustering is overly reliant on the choice, domain, or size of the public probe set, we conducted extensive sensitivity experiments. Specifically, we replaced the baseline mixed probes (GLUE/SQuAD) with out-of-domain probes from legal (\textit{CaseHOLD}), medical (\textit{MedQA}), and news (\textit{AG\_News}) domains. Furthermore, we evaluated probe-set size reductions by taking random 50\% and 70\% sub-samples of the baseline set.} 

\textcolor{red}{As summarized in Table~\ref{tab:probe_sensitivity}, the pairwise client-distance rankings under all out-of-domain probe sets maintain a strong Spearman rank correlation ($\rho \ge $0.85) with the baseline ranking. Moreover, reducing the probe size from 100 down to 50 yields minimal impact on both the silhouette score (0.42 vs. 0.41) and downstream test accuracy (82.35\% vs. 82.15\%). These results confirm that our behavior-aware fingerprinting is remarkably robust to probe-domain shifts, sample composition, and probe quantity.}

\begin{table}[h!]
	\centering
	\caption{\textcolor{red}{Sensitivity of behavior-aware fingerprints across different probe domains and probe-set sizes (evaluating client-distance ranking consistency via Spearman $\rho$ relative to the Mixed baseline).}}
	\label{tab:probe_sensitivity}
	\resizebox{0.98\linewidth}{!}{
		\begin{tabular}{lcccc}
			\hline
			\textbf{Probe set} & \textbf{Probe size ($Q$)} & \textbf{Spearman $\rho$} & \textbf{Silhouette score} & \textbf{Test accuracy} \\
			\hline
			Mixed baseline (GLUE/SQuAD) & 100 & 1.00 & 0.42 & 82.35\% \\
			MedQA (medical) & 100 & 0.87 & 0.40 & 81.90\% \\
			CaseHOLD (legal) & 100 & 0.85 & 0.39 & 81.75\% \\
			AG News (news) & 100 & 0.89 & 0.41 & 82.10\% \\
			Random 50\% subset & 50 & 0.92 & 0.41 & 82.15\% \\
			Random 70\% subset & 70 & 0.95 & 0.42 & 82.25\% \\
			\hline
	\end{tabular}}
    \vspace{1mm}
    \\
    \raggedright \tiny{\textdagger~Subsampled from the Mixed baseline set. All Spearman $\rho$ values are calculated relative to the client pairwise distances under the full Mixed baseline set ($Q=$ 100).}
\end{table}

\textcolor{red}{\textit{(ii) Addressing Potential Privacy Leakage of Task-Related Information via Probes.} We clarify that the public probe set consists exclusively of unlabelled, non-sensitive, and publicly accessible data instances.} 
\begin{itemize}
	\item \textcolor{red}{\textit{No Raw Probe Data Transmission:} Clients never transmit raw probe instances nor individual sequence embeddings. Instead, only aggregated moment statistics $\mathcal{R}_n=(\mu_n,\Sigma_n)$, or their sketched variants, over the public probe batch are shared with the aggregator.}
	
	\item \textcolor{red}{\textit{No Direct Transmission of Private Samples:} Because the probe set is independent of the client's local private dataset $\mathcal{D}_n$, evaluating the probe set does not directly transmit any instance-level raw text from $\mathcal{D}_n$. Nevertheless, the uploaded aggregate profiles may reflect model behavior learned from local data; therefore, we do not claim a formal differential privacy or mutual-information bound for these profiles.}
\end{itemize}

\textcolor{red}{\textit{(iii) Addressing Comparability of Probe Representations Across Tasks.} Behavior-aware clustering is performed \textit{within} a specific collaborative training task among participating clients sharing the same global backbone architecture at round $t$.} 
\begin{itemize}
	\item \textcolor{red}{\textit{Intra-Task Homogeneity:} Within a single task, all clients process the exact same public probe set through their local models. Thus, their output activation profiles $\mathcal{R}_n$ share a unified semantic coordinate system, making pairwise Symmetric KLD distances directly comparable and meaningful for grouping clients with similar behaviors.}
	
	\item \textcolor{red}{\textit{Cross-Task Independence:} Clustering is task-specific and does not require absolute representation alignment across distinct downstream tasks. For a new task, the public probe set serves as a common benchmark relative to that task's global initialization, ensuring valid relative distance rankings.}
\end{itemize}

\textcolor{red}{\textit{(iv) Comparison with Alternative Similarity Metrics.} In addition to probe selection, the formulation of the distance metric plays a critical role in accurately capturing client behavioral discrepancies. We compare our default Gaussian Symmetric KLD against two mean-only baselines (Euclidean distance and Cosine similarity) as well as the Gaussian Wasserstein-2 distance. As summarized in Table~\ref{tab:metric_comparison}, mean-only metrics suffer from a noticeable decline in cluster quality and accuracy because they discard second-order covariance information ($\Sigma_n$).}

\begin{table}[t]
	\centering
	\caption{\textcolor{red}{Comparison of alternative clustering metrics.}}
	\label{tab:metric_comparison}
	\resizebox{0.98\linewidth}{!}{
		\begin{tabular}{lcccc}
			\hline
			\textbf{Metric} & \textbf{Profile input} & \textbf{Silhouette score} & \textbf{Test accuracy} & \textbf{Computational property} \\
			\hline
			Euclidean distance & Mean vector & 0.28 & 78.40\% & Closed form \\
			Cosine similarity & Mean vector & 0.31 & 79.15\% & Closed form \\
			Gaussian Wasserstein-2 & $(\mu,\Sigma)$ & 0.41 & 82.10\% & Closed form; matrix square root \\
			\textbf{Symmetric KLD (ours)} & \textbf{$(\mu,\Sigma)$} & \textbf{0.42} & \textbf{82.35\%} & \textbf{Closed form} \\
			\hline
	\end{tabular}}
	\vspace{-3mm}
\end{table}

\textcolor{red}{While the Gaussian Wasserstein-2 metric incorporates covariance statistics and yields competitive results, it introduces non-trivial computational overhead due to the required matrix square-root operations ($\Sigma^{1/2}$). In contrast, Symmetric KLD achieves optimal cluster separation (Silhouette score of 0.42) and downstream test accuracy (82.35\%) while strictly retaining an efficient, closed-form analytical expression. These empirical trade-offs validate our selection of Gaussian moment-based fingerprints combined with Symmetric KLD.}

\textcolor{red}{Consequently, empirical evaluations confirm that ELSA's public probe set is highly resilient to domain shifts, leak-free regarding private client data, and capable of providing robust, reliable client clustering across diverse downstream tasks. Furthermore, pairing moment-based profiles $(\mu, \Sigma)$ with Symmetric KLD yields an optimal trade-off between clustering fidelity and computational efficiency.}

\subsection{\textcolor{red}{Architectural Extension of ELSA to Decoder-Only LLMs}}
\label{app:decoder_extension}
\textcolor{red}{As discussed previously, although the main formulation adopts encoder-only notation for behavioral fingerprinting and model partitioning, the key components of ELSA operate at the Transformer-block level and are therefore directly applicable beyond encoder architectures. Accordingly, we present a principled extension of ELSA from encoder-only models to current decoder-only LLMs (e.g., Qwen and Llama). Specifically, we describe the block-level partitioning of decoder-only LLMs across the client--edge--client pipeline, together with the corresponding causal language-modeling objective and local loss computation. We further explain how behavioral fingerprints can be constructed without \texttt{[CLS]} tokens and discuss KV-cache management during split training and inference.}

\textcolor{red}{For a decoder-only LLM comprising $M$ Transformer blocks, ELSA's three-tier execution pipeline naturally accommodates the causal architecture as follows:}
{\small
	\begin{align*}
		&\textcolor{red}{\text{Client (Part 1):} \quad \text{Input Embedding} + \text{Blocks } 1 : p_n,} \\
		& \textcolor{red}{\rightarrow \text{Edge Server (Part 2):} \quad \text{Blocks } (p_n + 1) : (p_n + q_n),} \\
		& \textcolor{red}{\rightarrow \text{Client (Part 3):} \quad \text{Blocks } (p_n + q_n + 1) : M + \text{LM Head}.}
	\end{align*}
}

\textcolor{red}{Under this partition, the dynamic optimization for $p_n$ and $q_n$ remains directly applicable. The client retains the input embedding, final causal blocks, and the language model (LM) head, while the computationally intensive intermediate blocks are transferred to the ES.}

\begin{itemize}
	\item \textcolor{red}{\textit{Privacy and Loss Computation:} During causal language model (CLM) fine-tuning, the client computes the next-token prediction loss locally: $\mathcal{L}_{\mathrm{CLM}} = -\frac{1}{T-1} \sum_{t=1}^{T-1} \log p_\theta(x_{t+1}\mid x_{\leq t})$. Because raw prompts, target labels, and logits stay entirely within the client boundary, the privacy guarantees of our 3-part split architecture are fully preserved. When weight tying is used between embeddings and the LM head, both modules naturally reside on the client; under PEFT/LoRA settings, shared base weights remain frozen while only adapter parameters are aggregated at the edge/global server.}
	
	\item \textcolor{red}{\textit{Behavioral Fingerprinting without \texttt{[CLS]} Tokens:} Instead of relying on a \texttt{[CLS]} token, a decoder-only model extracts behavioral fingerprints using the hidden state of the EOS token (or the last valid prompt token): $T_n^{(j)} = h_{n, t_{\mathrm{eos}}}^{(j)} \in \mathbb{R}^{D^{\mathsf{hidden}}}$. Since this state causally incorporates the prompt context, it provides a robust sequence-level representation for Gaussian profiling and symmetric-KLD clustering without altering the underlying protocol.}
	
	\item \textcolor{red}{\textit{KV Cache in Training vs. Inference:} \textcolor{red}{For training/fine-tuning under teacher forcing, gradients are backpropagated through the full sequence, making KV caching unnecessary during split backpropagation. For inference extensions, local KV caches are maintained layer-wise at the client (Part 1/3) and ES, using session affinity to ensure generation consistency.}}
\end{itemize}

\textcolor{red}{Overall, ELSA extends naturally to decoder-only LLMs with only minor adaptations to behavioral fingerprinting, loss computation, and KV-cache management.}

\subsection{\textcolor{red}{Cryptographic Setup, Salt Management, and Privacy Boundaries of $\mathcal{Q}_n$}}
\label{app:cryptographic_setup}

\textcolor{red}{This appendix clarifies the cryptographic assumptions, threat model, and privacy boundaries associated with the SS-OP transformation $\mathcal{Q}_n$. In particular, we describe the lifecycle of the master salt $s$, the trust boundaries among clients, edge servers, and the cloud server, and the reconstruction ambiguity introduced by SS-OP and dimensionality-reducing sketching.}

\textcolor{red}{\textit{(i) Salt Management and Key Generation.} The master salt $s$ is generated and administered by the trusted CS. During the initial network handshake or a dynamic parameter update phase, the CS distributes $s$ directly to each participating client $n$ through a secure and authenticated communication channel (e.g., TLS), while the ES are excluded from receiving $s$. Each client independently derives its local pseudo-random number generator seed as $\text{seed}_n=\operatorname{Hash}(s\parallel n)$, as specified in Eq.~(18), and uses it to generate $\mathbf{V}_n$. The protocol may periodically renew $s$, for example using epoch indices or timestamps, to limit prolonged exposure. We acknowledge that physical compromise of a participating client may disclose $s$ and weaken the SS-OP protection. Such endpoint compromise is outside our primary semi-honest ES threat model, and we do not claim resistance to SS-OP inversion after the relevant salt and transformation parameters have been disclosed.}

\textcolor{red}{\textit{(ii) Threat Model and Trust Boundaries.} The CS is considered trusted and is responsible for salt distribution and final model aggregation, while it does not receive intermediate activations or gradients. The ES are modeled as semi-honest entities that follow the prescribed protocol while attempting to infer information from the intermediate representations they observe. Because $s$ and $\text{seed}_n$ are withheld from the ES, an ES cannot directly reproduce $\mathbf{V}_n$ or the complete client-specific transformation $\mathcal{Q}_n$. This assumption defines the protection boundary of SS-OP rather than constituting a formal cryptographic secrecy guarantee.}

\textcolor{red}{\textit{(iii) Protection Conditions and Reconstruction Ambiguity under SS-OP and Sketching.} The protection provided by SS-OP relies on the local retention of the semantic subspace basis $\mathbf{U}_n$, the secrecy of $\mathbf{V}_n$ under the stated semi-honest ES threat model, and the irreversible dimensionality reduction introduced by sketching:}
\begin{itemize}
\item \textcolor{red}{\textit{Local Retention of $\mathbf{J}_n$ and $\mathbf{U}_n$:} The reference hidden-state matrix $\mathbf{J}_n$, from which $\mathbf{U}_n$ is derived via truncated SVD, is computed and retained exclusively on the client device and is never transmitted to the ES. Consequently, the ES does not directly observe either $\mathbf{J}_n$ or the resulting semantic subspace basis $\mathbf{U}_n$.}

\item \textcolor{red}{\textit{Secrecy of the Client-Specific Transformation:} The matrix $\mathbf{V}_n$ is generated locally using $\text{seed}_n=\operatorname{Hash}(s\parallel n)$. Provided that $s$ and $\text{seed}_n$ remain undisclosed under the assumed threat model, the ES cannot directly reproduce $\mathbf{V}_n$ or the complete SS-OP matrix $\mathcal{Q}_n=\mathbf{U}_n\mathbf{V}_n\mathbf{U}_n^\top+\left(\mathbf{I}_d-\mathbf{U}_n\mathbf{U}_n^\top\right)$.}

\item \textcolor{red}{\textit{Decompression and Transformation Boundary:} When the ES performs sketch decompression, it obtains only an approximation of the perturbed activation $\widetilde{\mathbf{H}}_n^{\mathsf{up}}=\mathcal{Q}_n\mathbf{H}_n^{\mathsf{up}}$, rather than the original activation $\mathbf{H}_n^{\mathsf{up}}$. Without the client-specific matrix $\mathcal{Q}_n$, the ES lacks the direct inverse transformation required to map the recovered perturbed representation back to the original activation space.}

\item \textcolor{red}{\textit{Reconstruction Ambiguity from Lossy Sketching:} The dimensionality-reducing sketching operator has a nontrivial null space, so multiple activation vectors may correspond to the same observed sketch. Reconstruction is therefore underdetermined without additional distributional or auxiliary prior information. As formalized in Lemma~\ref{lem:reconstruction_lower_bound} under the stated activation model, sketching produces a nonzero conditional reconstruction-error lower bound, while the undisclosed SS-OP matrix introduces additional reconstruction ambiguity that is not quantified by this bound.}
\end{itemize}

\textcolor{red}{Taken together, SS-OP and sketching provide a conditional reconstruction-error guarantee and empirical resistance against the evaluated attacks under the stated semi-honest ES threat model. This analysis does not constitute an information-theoretic secrecy, formal cryptographic security, or $(\epsilon,\delta)$-differential privacy guarantee.}

\subsection{\textcolor{red}{Comparison with Computational Sketching Frameworks}}
\label{sec:appendix_sketching}
\textcolor{red}{Computational sketching in FL, represented by the FedSKETCH family \cite{fedsketch,Fetchsgd,li2026sketched,shrivastavasketching,chengpersonalized} and data sketching methods \cite{liuone}, primarily applies linear random projections (e.g., Count Sketch, Gaussian/One-Bit projection) for gradient compression or distribution estimation. Although effective in reducing communication costs, these approaches operate under fundamentally different paradigms than ELSA across four main dimensions:}

\textcolor{red}{\textit{(i) Sketching objects:} Existing schemes sketch full-model cumulative gradients, clipped noisy updates, or raw data statistics \cite{liuone}, requiring clients to host full model parameters and execute complete forward-backward passes. In contrast, ELSA's sketching applies exclusively to intermediate activations (\textit{smashed representations}) within the split learning pipeline or edge-side LoRA increments, allowing front-end truncation without hosting the full LLM.}
	
\textcolor{red}{\textit{(ii) Sketching Roles \& Purposes:} Prior works employ sketching as the core primitive either solely for gradient compression or for distribution metric calculation. In ELSA, sketch-based communication compression serves as an auxiliary channel; the core heterogeneity awareness relies on public probes and Gaussian moments, decoupling sketching from clustering decisions.}
	
\textcolor{red}{\textit{(iii) Architectural Premises:} Standard sketching frameworks assume classical FL/PFL settings where the gradient dimension $d$ scales quadratically or linearly with massive LLM parameters. ELSA couples sketching with a three-tier SL+HFL architecture, where the sketching dimension depends strictly on the hidden layer width or adapter rank, remaining invariant to total LLM parameter size.}
	
\textcolor{red}{\textit{(iv) Privacy Semantics:} Traditional sketching privacy revolves around preventing the server from reconstructing full gradients or raw data. ELSA combines activation-level transmission with SS-OP perturbations, guaranteeing that raw private samples never leave the client while relaxing dependencies on gradient-reconstruction-based security assumptions.}

\textcolor{red}{Consequently, ELSA redefines the application of computational sketching in edge LLM training, shifting it from traditional full-model gradient compression to a lightweight, split-architecture-compatible communication primitive.}

\subsection{
\textcolor{red}{Proof of Lemma 1: Reconstruction Error Lower Bound}}

\textcolor{red}{
We first verify that the SS-OP transformation used in Lemma~1 is orthogonal. Let $ \mathbf{P}_n=\mathbf{U}_n\mathbf{U}_n^\top$. Since $\mathbf{P}_n$ is an orthogonal projector,
$\mathbf{P}_n^2=\mathbf{P}_n$, and
$\mathbf{U}_n^\top(\mathbf{I}_d-\mathbf{P}_n)=\mathbf{0}$.
Using $\mathbf{V}_n^\top\mathbf{V}_n=\mathbf{I}_r$, we obtain
\begin{align}
\mathcal{Q}_n^\top\mathcal{Q}_n
&=
\left[
\mathbf{U}_n\mathbf{V}_n^\top\mathbf{U}_n^\top
+
(\mathbf{I}_d-\mathbf{P}_n)
\right]
\left[
\mathbf{U}_n\mathbf{V}_n\mathbf{U}_n^\top
+
(\mathbf{I}_d-\mathbf{P}_n)
\right]
\notag\\
&=
\mathbf{U}_n\mathbf{V}_n^\top
\mathbf{V}_n\mathbf{U}_n^\top
+
(\mathbf{I}_d-\mathbf{P}_n)
\notag\\
&=
\mathbf{P}_n+(\mathbf{I}_d-\mathbf{P}_n)
=
\mathbf{I}_d.
\end{align}
Therefore, $\mathcal{Q}_n$ is orthogonal and invertible.
}

\textcolor{red}{
To derive a conservative reconstruction-error lower bound, we consider
a strengthened adversary that is additionally given the secret matrix
$\mathcal{Q}_n$. Providing this information can only improve the
adversary's reconstruction capability. Hence,
\begin{equation}
\resizebox{0.96\columnwidth}{!}{$
\displaystyle
\inf_{\mathcal{A}}
\mathbb{E}\!\left[
\left\|H_n-\mathcal{A}(\mathbf{S}_n,\mathbf{R})\right\|_F^2
\right]
\geq
\inf_{\mathcal{A}^{+}}
\mathbb{E}\!\left[
\left\|H_n-\mathcal{A}^{+}
(\mathbf{S}_n,\mathbf{R},\mathcal{Q}_n)\right\|_F^2
\right]
$}
\label{eq:stronger_adversary}
\end{equation}
Consequently, it is sufficient to establish a lower bound for the strengthened adversary.
}

\textcolor{red}{
Conditioned on $\mathbf{R}$ and $\mathcal{Q}_n$, define the effective
linear observation matrix as
\[
\mathbf{A}_n=\mathbf{R}\mathcal{Q}_n.
\]
Because $\mathcal{Q}_n$ is invertible,
$
\operatorname{rank}(\mathbf{A}_n)
=
\operatorname{rank}(\mathbf{R})
\leq d'
$.
For each activation vector $h_{n,t}$, the adversary observes
\[
s_{n,t}=\mathbf{A}_nh_{n,t}.
\]
Under the isotropic Gaussian activation model, the minimum mean squared
error estimator is the conditional expectation
\[
\widehat{h}_{n,t}^{\,*}
=
\mathbb{E}
\left[
h_{n,t}\mid
s_{n,t},\mathbf{A}_n
\right].
\]
The corresponding conditional covariance is
\begin{align}
\operatorname{Cov}
\left(
h_{n,t}\mid s_{n,t},\mathbf{A}_n
\right)
=
\sigma_H^2
\left[
\mathbf{I}_d
-
\mathbf{A}_n^\top
(\mathbf{A}_n\mathbf{A}_n^\top)^\dagger
\mathbf{A}_n
\right],
\label{eq:conditional_covariance}
\end{align}
where $(\cdot)^\dagger$ denotes the Moore--Penrose pseudoinverse.
}

\textcolor{red}{
The matrix $
\mathbf{A}_n^\top
(\mathbf{A}_n\mathbf{A}_n^\top)^\dagger
\mathbf{A}_n
$ is the orthogonal projector onto the row space of $\mathbf{A}_n$.
Its trace is therefore equal to
$\operatorname{rank}(\mathbf{A}_n)$. It follows that the minimum
reconstruction error for each activation vector satisfies
\begin{align}
\inf_{\widehat{h}_{n,t}}
\mathbb{E}
\left[
\left\|
h_{n,t}-\widehat{h}_{n,t}
\right\|_2^2
\right]
&=
\operatorname{Tr}
\left[
\operatorname{Cov}
\left(
h_{n,t}\mid s_{n,t},\mathbf{A}_n
\right)
\right]
\notag\\
&=
\sigma_H^2
\left[
d-\operatorname{rank}(\mathbf{A}_n)
\right]
\notag\\
&\geq
\sigma_H^2(d-d').
\label{eq:per_token_mmse}
\end{align}
}

\textcolor{red}{
Summing Eq.~(\ref{eq:per_token_mmse}) over the $T$ independently
generated activation vectors gives
\begin{align}
\mathbb{E}
\left[
\left\|
H_n-\widehat{H}_n
\right\|_F^2
\right]
&\geq
T\sigma_H^2(d-d')
\notag\\
&=
T\sigma_H^2d
\left(
1-\frac{d'}{d}
\right)
\notag\\
&=
T\left(
1-\frac{1}{\rho}
\right)
\operatorname{Tr}(\Sigma_H).
\end{align}
Since $\rho>1$ and
$\operatorname{Tr}(\Sigma_H)>0$, the lower bound is strictly positive.
This completes the proof.
}
\hfill$\blacksquare$

\begin{remark}[
\textcolor{red}{Scope and Privacy Interpretation of Lemma 1}]
\label{remark:lemma_privacy_scope}

\textcolor{red}{
The bound in Lemma~\ref{lem:reconstruction_lower_bound} is conservative
because it is derived for a strengthened adversary that is additionally
given the secret SS-OP matrix $\mathcal{Q}_n$. In the actual ELSA threat
model, the edge server does not possess $\mathcal{Q}_n$. Therefore,
SS-OP introduces additional reconstruction ambiguity beyond the
irreversible information loss caused by sketching.
}

\textcolor{red}{
The independent isotropic Gaussian activation assumption is introduced
to obtain a closed-form Bayesian MMSE lower bound. It should not be
interpreted as claiming that practical Transformer activations follow
an exact isotropic Gaussian distribution. For a general Gaussian model
$h_{n,t}\sim\mathcal{N}(0,\Sigma_H)$, the conditional MMSE covariance
for the strengthened adversary is
}
\begin{equation}
\resizebox{0.96\columnwidth}{!}{$
	\textcolor{red}{
\displaystyle
\Sigma_{\mathrm{post}}
=
\Sigma_H
-
\Sigma_H\mathbf{A}_n^\top
\left(
\mathbf{A}_n\Sigma_H\mathbf{A}_n^\top
\right)^\dagger
\mathbf{A}_n\Sigma_H,}
$}
\end{equation}
\textcolor{red}{where $\mathbf{A}_n=\mathbf{R}\mathcal{Q}_n$. Accordingly,
}
\begin{equation}
\textcolor{red}{
\mathbb{E}
\left[
\left\|H_n-\widehat{H}_n\right\|_F^2
\right]
\geq
T\,\operatorname{Tr}(\Sigma_{\mathrm{post}}).
}
\end{equation}

\textcolor{red}{
When $\Sigma_H=\sigma_H^2\mathbf{I}_d$, this expression reduces to the
closed-form result in Lemma~\ref{lem:reconstruction_lower_bound}.
For non-Gaussian or cross-token-correlated activations, the exact
reconstruction error depends on the underlying distribution and the
adversary's prior knowledge. Therefore, Lemma~\ref{lem:reconstruction_lower_bound}
provides a conditional analytical reconstruction-error guarantee under
explicitly stated assumptions. It does not constitute an
$(\epsilon,\delta)$-differential privacy guarantee and is complemented
by empirical evaluations against activation inversion, generative
embedding inversion, and membership inference attacks.
}
\end{remark}

\subsection{Proof of Theorem 1}
To analyze the convergence of ELSA, we first note that only the adapter parameters $\theta$ are trainable, while the pre-trained LLM backbone remains frozen throughout fine-tuning and is thus omitted from the optimization arguments. We then establish convergence under the following standard assumptions~\cite{DeFedGCN}:

\begin{assumption}[Lipschitz Smoothness of Local Loss Functions]
	There exists $\mathcal{L} > 0$ such that for each client $n$ and any parameter vectors $\theta$, $\theta'$,
	\begin{equation}
		\left\| \nabla F_n(\theta) - \nabla F_n(\theta') \right\| \leq \mathcal{L} \left\| \theta - \theta' \right\|, \quad \forall n, \theta, \theta'.
	\end{equation}
\end{assumption}

\begin{assumption}[Bounded SGD Variance and non-IID Heterogeneity]
	There exist constants ${\sigma_1}^2, {\sigma_2}^2 > 0$ such that
	\begin{align}
		\mathbb{E} \left[ \left\| \nabla F_n(\theta) - \nabla f_n(\theta; \mathbf{b}) \right\|^2 \right] \leq \sigma_1^2, \quad \text{(Local SGD variance)} \\
		\left\| \nabla F_n(\theta) - \nabla \widetilde{F}(\theta) \right\|^2 \leq \sigma_2^2, \quad \text{(non-IID data heterogeneity)}
	\end{align}
	where $f_n(\cdot; \mathbf{b})$ denotes the loss on a mini-batch $\mathbf{b}$ sampled from client $n$'s local data, and $\widetilde{F}(\theta) = \frac{1}{N} \sum_{n=1}^N F_n(\theta)$ is the global objective.
\end{assumption}

\begin{assumption}[Error from Sketching and Median Reconstruction]
	The SS-OP transformation is orthogonal and preserves norms and inner products without bias. However, the subsequent hash-based sketching and median-based reconstruction introduce approximation errors. Specifically, the reconstructed hidden state $\widetilde{\mathbf{H}}^{\mathsf{up}}_n$ satisfies:
	\begin{align}
		\left\| \mathbb{E} [\widetilde{\mathbf{H}}^{\mathsf{up}}_n] - \mathbf{H}^{\mathsf{up}}_n \right\| &\leq \epsilon_{\text{sketch}}, \quad \text{(Bias)} \\
		\mathbb{E} \left[ \left\| \widetilde{\mathbf{H}}^{\mathsf{up}}_n - \mathbb{E}[\widetilde{\mathbf{H}}^{\mathsf{up}}_n] \right\|^2 \right] &\leq \sigma_{\text{sketch}}^2, \quad \text{(Variance)}
	\end{align}
	where $\epsilon_{\text{sketch}}$ arises from the bias of the median estimator, and $\sigma_{\text{sketch}}^2$ accounts for variance due to hashing collisions. The total mean squared error is bounded by $\delta^2 \triangleq \epsilon_{\text{sketch}}^2 + \sigma_{\text{sketch}}^2$.
\end{assumption}

Under Assumption 1, the reconstruction error in hidden states induces a bounded error in the computed gradients. Extending Lemma 1 of~\cite{fedsketch} to the biased setting, the gradient $\mathbf{g}_n$ computed using the reconstructed states satisfies:
\begin{align}
	\mathbb{E}\left[ \left\| \mathbf{g}_n - \nabla F_n(\theta) \right\|^2 \right] 
	&\leq \underbrace{\mathbb{E}\left[ \left\| \nabla f_n(\theta; \mathbf{b}) - \nabla F_n(\theta) \right\|^2 \right]}_{\leq \sigma_1^2} \nonumber
	\\ & + \underbrace{\mathcal{L}^2 \delta^2}_{\text{sketching error}} 
	\triangleq \sigma_{\text{local}}^2.
\end{align}
Here, $\sigma_{\text{local}}^2$ captures the combined effect of stochastic sampling and sketching-induced noise at each client.

Critically, the non-IID data heterogeneity (Assumption 2) is \emph{not} absorbed into $\sigma_{\text{local}}^2$; instead, it manifests as a systematic deviation between local and global gradients. Let $\mathbf{r}_g$ denote the aggregated update direction after client-side computation, compression, and server-side averaging. Then, the total error relative to the true global gradient satisfies:
\begin{align}
	\mathbb{E}\left[ \left\| \mathbf{r}_g - \nabla \widetilde{F}(\theta_g) \right\|^2 \right] 
	\leq \frac{1}{N} \sum_{n=1}^N \mathbb{E}\left[ \left\| \mathbf{g}_n - \nabla F_n(\theta_g) \right\|^2 \right] \nonumber \\
	+ \frac{1}{N} \sum_{n=1}^N \left\| \nabla F_n(\theta_g) - \nabla \widetilde{F}(\theta_g) \right\|^2 
	\leq \sigma_{\text{local}}^2 + \sigma_2^2.
\end{align}
This decomposition explicitly separates the impact of \emph{local randomness} ($\sigma_{\text{local}}^2$) from \emph{data heterogeneity} ($\sigma_2^2$), ensuring a theoretically sound convergence characterization.

Building upon the above error decomposition, we establish the following convergence guarantee for ELSA.

\begin{proof}
	By Assumption 1 (Lipschitz smoothness of $\widetilde{F}(\cdot)$), the following inequality holds for any global iteration $g$:
	\begin{equation}
		\widetilde{F}(\theta_{g+1}) \leq \widetilde{F}(\theta_g) - \eta \langle \nabla \widetilde{F}(\theta_g), \mathbf{r}_g \rangle + \frac{\mathcal{L} \eta^2}{2} \|\mathbf{r}_g\|^2.
	\end{equation}
	Taking expectation over the randomness in SGD and sketching, we decompose the update direction as:
	\begin{equation}
		\textcolor{blue}{\mathbb{E}[\mathbf{r}_g] = \nabla \widetilde{F}(\theta_g) + \left( \mathbb{E}[\mathbf{r}_g] - \nabla \widetilde{F}(\theta_g) \right),}
	\end{equation}
	but more importantly, from the error decomposition in Section~III.C, we have:
	\begin{equation}
		\mathbb{E}\left[ \left\| \mathbf{r}_g - \nabla \widetilde{F}(\theta_g) \right\|^2 \right] \leq \sigma_{\text{local}}^2 + \sigma_2^2.
	\end{equation}
	
	We now bound the two key terms in the smoothness inequality.
	
	\noindent\textbf{\textcolor{blue}{(Step 1) Inner product term.}} \textcolor{blue}{Using the identity $\mathbb{E}[\langle a, X \rangle] = \langle a, \mathbb{E}[X] \rangle$ and our bias definition $\mathbf{e}_g = \mathbb{E}[\mathbf{r}_g] - \nabla \widetilde{F}(\theta_g)$, we obtain:}
	\begin{align}
		\textcolor{blue}{-\mathbb{E}[\langle \nabla \widetilde{F}(\theta_g), \mathbf{r}_g \rangle] }
		&\textcolor{blue}{= -\langle \nabla \widetilde{F}(\theta_g), \mathbb{E}[\mathbf{r}_g] \rangle} \nonumber \\
		&\textcolor{blue}{= -\|\nabla \widetilde{F}(\theta_g)\|^2 - \langle \nabla \widetilde{F}(\theta_g), \mathbf{e}_g \rangle.}
	\end{align}
	\textcolor{blue}{Applying Young's inequality ($-\langle a, b \rangle \leq \frac{1}{2}\|a\|^2 + \frac{1}{2}\|b\|^2$):}
	\begin{equation}
		\textcolor{blue}{-\langle \nabla \widetilde{F}(\theta_g), \mathbf{e}_g \rangle \leq \frac{1}{2} \|\nabla \widetilde{F}(\theta_g)\|^2 + \frac{1}{2} \|\mathbf{e}_g\|^2.}
	\end{equation}
	\textcolor{blue}{By Jensen's inequality and the error bound in Eq.~(4), the squared expected bias satisfies $\|\mathbf{e}_g\|^2 = \|\mathbb{E}[\mathbf{r}_g] - \nabla \widetilde{F}(\theta_g)\|^2 \leq \mathbb{E}[\|\mathbf{r}_g - \nabla \widetilde{F}(\theta_g)\|^2] \leq \sigma_{\text{local}}^2 + \sigma_2^2$. Thus,}
	\begin{equation}
		\textcolor{blue}{-\mathbb{E}[\langle \nabla \widetilde{F}(\theta_g), \mathbf{r}_g \rangle] \leq -\frac{1}{2} \|\nabla \widetilde{F}(\theta_g)\|^2 + \frac{1}{2} (\sigma_{\text{local}}^2 + \sigma_2^2).}
	\end{equation}
	
	\noindent\textbf{(Step 2) Quadratic term.} Using $\|a + b\|^2 \leq 2\|a\|^2 + 2\|b\|^2$,
	\begin{align}
		\mathbb{E}[\|\mathbf{r}_g\|^2] 
		&= \mathbb{E}[\|\mathbf{r}_g - \nabla \widetilde{F}(\theta_g) + \nabla \widetilde{F}(\theta_g)\|^2] \nonumber \\
		&\leq 2\mathbb{E}[\|\mathbf{r}_g - \nabla \widetilde{F}(\theta_g)\|^2] + 2\|\nabla \widetilde{F}(\theta_g)\|^2 \nonumber \\
		&\leq 2(\sigma_{\text{local}}^2 + \sigma_2^2) + 2\|\nabla \widetilde{F}(\theta_g)\|^2.
	\end{align}
	
	Substituting (Step 1) and (Step 2) into the smoothness inequality:
	\begin{align}
		\mathbb{E}[\widetilde{F}(\theta_{g+1})] 
		&\leq \widetilde{F}(\theta_g) - \eta \left( \frac{1}{2} \|\nabla \widetilde{F}(\theta_g)\|^2 - \frac{1}{2} (\sigma_{\text{local}}^2 + \sigma_2^2) \right) \nonumber \\
		&\quad + \frac{\mathcal{L} \eta^2}{2} \left( 2(\sigma_{\text{local}}^2 + \sigma_2^2) + 2\|\nabla \widetilde{F}(\theta_g)\|^2 \right) \nonumber \\
		&= \widetilde{F}(\theta_g) - \left( \frac{\eta}{2} - \mathcal{L} \eta^2 \right) \|\nabla \widetilde{F}(\theta_g)\|^2 \nonumber \\
		&\quad + \frac{\eta}{2} (\sigma_{\text{local}}^2 + \sigma_2^2) + \mathcal{L} \eta^2 (\sigma_{\text{local}}^2 + \sigma_2^2).
	\end{align}
	
	Choose learning rate $\eta = \frac{1}{\mathcal{L} \sqrt{G}}$. For $G \geq 4$, we have $\mathcal{L} \eta \leq \frac{1}{2}$, so $\frac{\eta}{2} - \mathcal{L} \eta^2 \geq \frac{\eta}{4}$. Rearranging:
	\begin{equation}
		\frac{\eta}{4} \|\nabla \widetilde{F}(\theta_g)\|^2 \leq \widetilde{F}(\theta_g) - \mathbb{E}[\widetilde{F}(\theta_{g+1})] + \left( \frac{\eta}{2} + \mathcal{L} \eta^2 \right) (\sigma_{\text{local}}^2 + \sigma_2^2).
	\end{equation}
	
	Summing over $g = 0$ to $G-1$ and taking total expectation:
	\begin{equation}
		\hspace{-4.5mm}
		\resizebox{0.48\textwidth}{!}{$
			\begin{aligned}
				\frac{\eta}{4} \sum_{g=0}^{G-1} \mathbb{E}[\|\nabla \widetilde{F}(\theta_g)\|^2] 
				&\leq \widetilde{F}(\theta_0) - \widetilde{F}(\theta_G) + G \left( \frac{\eta}{2} + \mathcal{L} \eta^2 \right) (\sigma_{\text{local}}^2 + \sigma_2^2) \\
				&\leq \widetilde{F}(\theta_0) - \widetilde{F}^* + G \left( \frac{\eta}{2} + \mathcal{L} \eta^2 \right) (\sigma_{\text{local}}^2 + \sigma_2^2).
			\end{aligned}
			$}\hspace{-3.15mm}
	\end{equation}
	where $\widetilde{F}^* = \min_\theta \widetilde{F}(\theta)$. Then, dividing both sides by $\frac{\eta}{4} G$:
	\begin{align}
		&\frac{1}{G} \sum_{g=0}^{G-1} \mathbb{E}[\|\nabla \widetilde{F}(\theta_g)\|^2] 
		\leq \frac{4(\widetilde{F}(\theta_0) - \widetilde{F}^*)}{\eta G} + \nonumber \\  &~~~~~~~~~~~~~~~~~~~~~~~~~~~~~~4 \left( \frac{1}{2} + \mathcal{L} \eta \right) (\sigma_{\text{local}}^2 + \sigma_2^2) \nonumber \\
		&= \frac{4\mathcal{L}(\widetilde{F}(\theta_0) - \widetilde{F}^*)}{\sqrt{G}} + 4 \left( \frac{1}{2} + \frac{1}{\sqrt{G}} \right) (\sigma_{\text{local}}^2 + \sigma_2^2).
	\end{align}
	
	For sufficiently large $G$ (e.g., $G \geq 4$), $\frac{1}{\sqrt{G}} \leq \frac{1}{2}$, so $\frac{1}{2} + \frac{1}{\sqrt{G}} \leq 1$. Thus,
	\begin{equation}
		\frac{1}{G} \sum_{g=0}^{G-1} \mathbb{E}[\|\nabla \widetilde{F}(\theta_g)\|^2] 
		\leq \frac{4\mathcal{L}(\widetilde{F}(\theta_0) - \widetilde{F}^*)}{\sqrt{G}} + \frac{\sigma_{\text{local}}^2}{\sqrt{G}} + \sigma_2^2,
	\end{equation}
	where we absorb constants into the $\sigma_{\text{local}}^2 / \sqrt{G}$ term for clarity (since $\sigma_{\text{local}}^2$ already includes problem-dependent constants). This matches the bound in Theorem~1, completing the proof.
\end{proof}

\begin{figure}[!t]
	\centering
	
	\subfigure[]{
		\includegraphics[trim=0.5cm 0.7cm 0.6cm 0.5cm, clip, width=0.46\columnwidth]{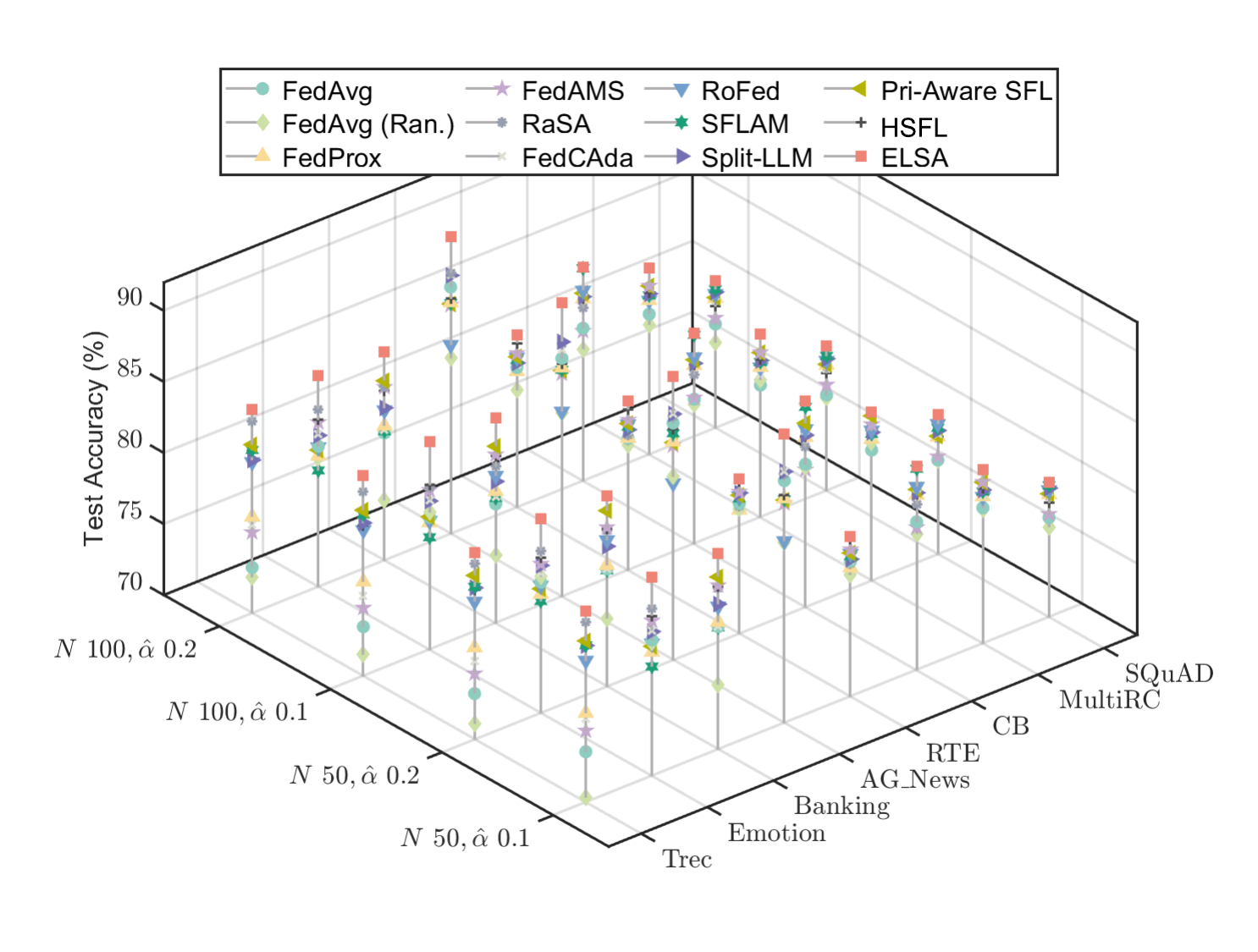}
	}
	\subfigure[]{
		\includegraphics[trim=0.5cm 0.7cm 0.6cm 0.5cm, clip, width=0.46\columnwidth]{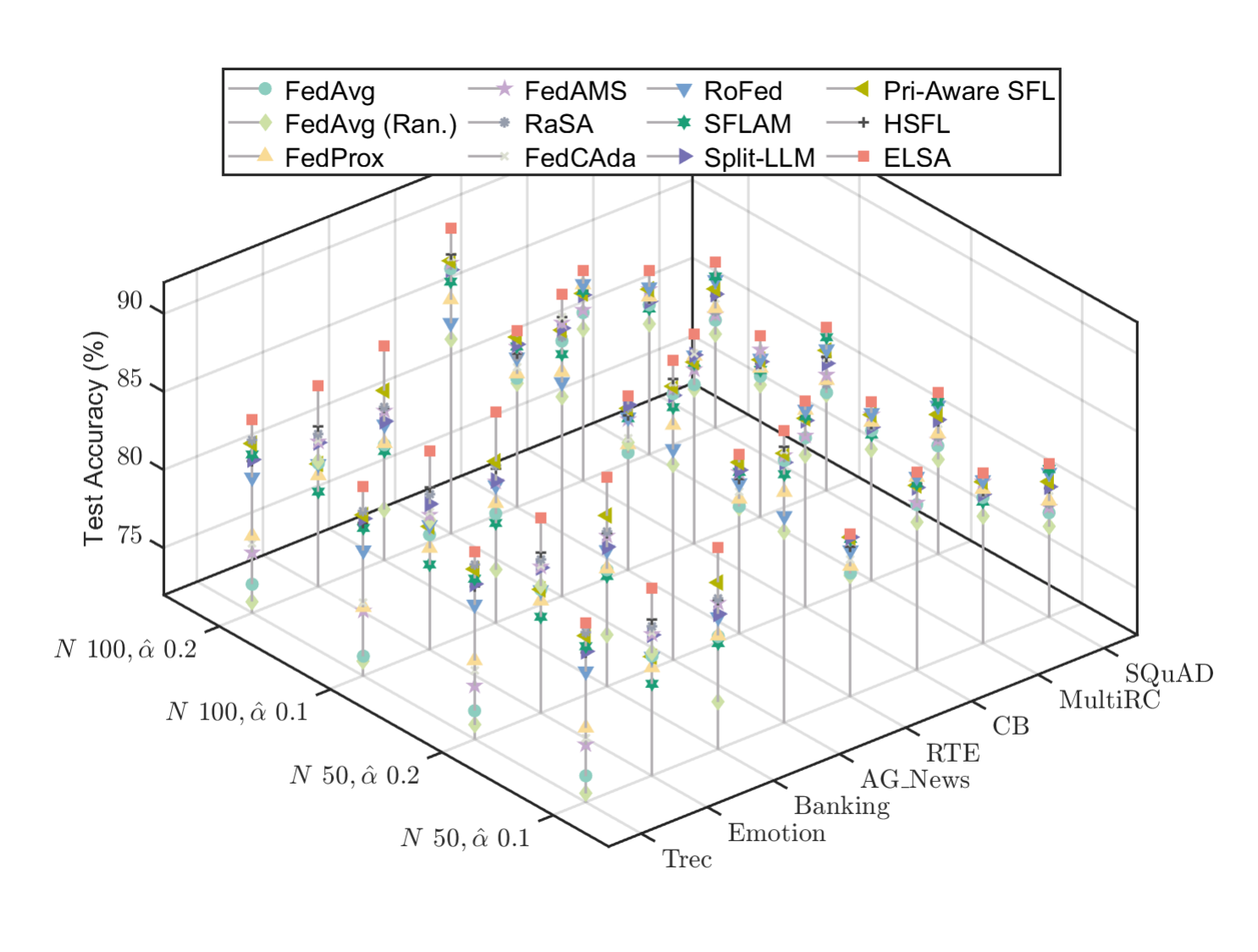}
	}
	
	\vspace{-3mm}
	\caption{\textcolor{red}{Performance of different models under varying client counts and data heterogeneity settings (BERT on the left, Qwen3-4B on the right).}}
	\vspace{-5mm}
	\label{fig:10}
\end{figure} 

\subsection{\textcolor{red}{On-Device Hardware Profiling, Active Memory Breakdown, and Probe Overhead Analysis}}
\label{app:hardware_profiling}

\textcolor{red}{To rigorously evaluate the physical feasibility of ELSA under realistic edge constraints, we implement a dedicated \textit{On-Device Profiling Layer}. Following the empirical profiling methodologies of SplitCom\cite{SplitCom} and FwdLLM \cite{FwdLLM}, we measure peak memory footprints, energy consumption, and per-stage latency on two physical platforms representing distinct edge tiers:}
\begin{itemize}
	\item \textcolor{red}{\textit{NVIDIA RTX 4090 (24\,GB GDDR6X):} Proxy for high-end edge servers, edge nodes, or vehicular domain controllers.}
	\item \textcolor{red}{\textit{NVIDIA Jetson AGX Orin (64\,GB LPDDR5 Unified Memory):} Direct representative of resource-constrained mobile and embedded edge devices.}
\end{itemize}

\textcolor{red}{\textit{(i) Empirical On-Device Resource Benchmarks.}
Table~\ref{tab:appendix_resource_profile} provides a breakdown of resource usage across execution stages, measured using BERT-base ($110$\,M parameters, FP32, LoRA rank $r=8$, local depth $p=3$, batch size $B_n=1$, sequence length $S=128$). Memory allocation is logged via \texttt{torch.cuda.max\_memory\_allocated()} on the RTX 4090 and resident set size (RSS) via \texttt{tegrastats} on the Jetson Orin. Energy is measured via \texttt{pynvml} and onboard INA sensors, respectively.}

\textcolor{red}{We explicitly refrain from linearly scaling measurements from desktop-class GPUs to embedded platforms due to structural gaps in memory bandwidth (1.01\,TB/s vs. $204$\,GB/s) and Thermal Design Power (TDP: 450\,W vs. 50\,W). Under small batch sizes, fine-tuning edge models is strictly memory-bandwidth-bound rather than compute-bound\footnote{\textcolor{red}{The memory footprint scaling across $p \in$ \{1, 3, 9\} reflects non-linear physical allocation due to the PyTorch CUDA memory caching mechanism. Beyond layer weight additions, transitioning from $p=$ 1 to $p=$ 3 incorporates the fixed overhead of output head projection and autograd graph contexts. For $p=$ 9, the prolonged backpropagation path triggers higher-tier CUDA allocator block reservations, yielding a stepped peak memory growth.}}.}

\begin{table*}[!t]
    \centering
    \caption{\textcolor{red}{Per-client resource profile on two representative edge platforms: 
        an RTX-4090 workstation (proxy for high-end edge) 
        and a Jetson AGX Orin 64GB (resource-constrained mobile edge). 
        Model: BERT-base (110M), full-precision FP32, LoRA $r{=}$8, local layers $p{=}$3, 
        batch size 1, sequence length 128. 
        Memory reported as \texttt{torch.cuda.max\_memory\_allocated()} on 4090 
        and resident set size via \texttt{tegrastats} on Orin. 
        Energy measured via \texttt{pynvml} (4090) and onboard INA sensors (Orin). 
        All numbers reported as mean $\pm$ std over 5 independent runs.}}
    \label{tab:appendix_resource_profile}
    \small
    \renewcommand{\arraystretch}{1.15}
    \setlength{\tabcolsep}{3.5pt}
    
    \begin{adjustbox}{max width=\textwidth}
        \begin{tabular}{@{} l *{3}{c} *{3}{c} @{}}
            \toprule
            
            \multicolumn{1}{c}{\multirow{2}{*}{\textbf{Stage}}}
            & \multicolumn{3}{c}{\textbf{RTX-4090 (High-End Edge)}}
            & \multicolumn{3}{c}{\textbf{Jetson AGX Orin (Mobile Edge)}} \\
            \cmidrule(lr){2-4} \cmidrule(lr){5-7}
            
            \multicolumn{1}{c}{}
            & \textbf{\shortstack{Peak Mem\\(GB)}}
            & \textbf{\shortstack{Energy\\(J)}}
            & \textbf{\shortstack{Latency\\(ms)}}
            & \textbf{\shortstack{Peak Mem\\(GB)}}
            & \textbf{\shortstack{Energy\\(J)}}
            & \textbf{\shortstack{Latency\\(ms)}} \\
            \midrule
            
            Probe fingerprint extraction (1 fwd pass)
            & 0.29 $\pm$ 0.01 & 0.04 $\pm$ 0.00 & 26.30 $\pm$ 1.20
            & 0.37 $\pm$ 0.02 & 0.07 $\pm$ 0.01 & 71.50 $\pm$ 3.80 \\
            \addlinespace[0.3em]
            
            Local training step (fwd + bwd + LoRA update)
            & 0.96 $\pm$ 0.03 & 0.39 $\pm$ 0.02 & 218.40 $\pm$ 6.70
            & 1.17 $\pm$ 0.05 & 0.62 $\pm$ 0.03 & 486.20 $\pm$ 14.30 \\
            \addlinespace[0.3em]
            
            Sketch \& SS-OP encoding (CPU-side)
            & 0.09 $\pm$ 0.00 & 0.01 $\pm$ 0.00 & 8.20 $\pm$ 0.50
            & 0.14 $\pm$ 0.01 & 0.01 $\pm$ 0.00 & 19.50 $\pm$ 1.60 \\
            \addlinespace[0.3em]
            
            Communication (14.2\,MB upload)
            & 0.00 $\pm$ 0.00 & 0.00 $\pm$ 0.00 & 36.00 $\pm$ 2.10
            & 0.00 $\pm$ 0.00 & 0.00 $\pm$ 0.00 & 108.00 $\pm$ 5.40 \\
            \addlinespace[0.3em]
            
            \textbf{Total per round}
            & \textbf{0.97 $\pm$ 0.03} & \textbf{0.44 $\pm$ 0.02} & \textbf{288.90 $\pm$ 8.40}
            & \textbf{1.18 $\pm$ 0.06} & \textbf{0.71 $\pm$ 0.04} & \textbf{685.20 $\pm$ 18.70} \\
            \midrule
            
            Static $p{=}$9 (conservative)
            & 1.84 $\pm$ 0.06 & 0.79 $\pm$ 0.03 & 442.50 $\pm$ 12.60
            & 2.63 $\pm$ 0.11 & 1.34 $\pm$ 0.06 & 1048.30 $\pm$ 42.50 \\
            \addlinespace[0.3em]
            
            Static $p{=}$1 (aggressive)
            & 0.42 $\pm$ 0.02 & 0.21 $\pm$ 0.01 & 118.60 $\pm$ 4.30
            & 0.59 $\pm$ 0.03 & 0.36 $\pm$ 0.02 & 289.40 $\pm$ 11.20 \\
            
            \bottomrule
            
            \multicolumn{7}{@{}l}{\scriptsize \textcolor{red}{All measurements at 25$^{\circ}$C ambient; Orin operates in 50\,W MAXN mode.}} \\
            \multicolumn{7}{@{}l}{\scriptsize \textcolor{red}{Probe fingerprint uses 32 representative prompts from the probe set.}} \\
            \multicolumn{7}{@{}l}{\scriptsize \textcolor{red}{Thermal note: on Orin, sustained 50-round training raised SoC temperature from 42$^{\circ}$C to 68$^{\circ}$C,}} \\
            \multicolumn{7}{@{}l}{\scriptsize \textcolor{red}{triggering $\sim$12\% frequency throttling (latency degrades from 486\,ms to 542\,ms). No throttling on 4090 (stable at 62$^{\circ}$C).}} \\
        \end{tabular}
    \end{adjustbox}
    \vspace*{-3mm}
\end{table*}

\textcolor{red}{\textit{(ii) Active Memory Breakdown \& U-Shaped Feasibility}
Under the client-side U-shaped structure ($p=$ 3), the theoretical active memory lower bound consists of:}
\begin{itemize}
	\item \textcolor{red}{\textit{Vocabulary Embedding Matrix (30522$\times$768, FP32):} 30522$\times$768$\times$ 4 Bytes $\approx$ 89.42 MB.}
	
	\item \textcolor{red}{\textit{Output Head (Part 3):} Weight-tied with the embedding layer in standard BERT implementations (0 MB additional static parameters); dynamic loss/logits tensor allocations are negligible ($<$ 0.5 MB).}
	
	\item \textcolor{red}{\textit{Local Transformer Layers ($p=$ 3):} $\approx$ 7.08 M parameters/layer $\times$ 4Bytes = 28.32MB; totaling 84.96 MB for $p=$ 3.}
	
	\item \textcolor{red}{\textit{Cached Activations (with Gradient Checkpointing):} $\approx$ 0.33 MB/layer, totaling 1.0 MB for $p=$3.}
	
	\item \textcolor{red}{\textit{Intermediate Tensors and KV Caches:} 2$\times p \times B_n \times S \times H \times \text{4 Bytes} \approx \text{2.4 MB}$.}
	
	\item \textcolor{red}{\textit{Runtime Framework Context (PyTorch + CUDA/cuDNN):} $\approx$ 350 MB on RTX 4090 and $\approx$ 500 MB on Jetson AGX Orin (due to unified memory architecture overhead).}
\end{itemize}

\textcolor{red}{Combining these components yields a theoretical lower bound of:}
\begin{equation}
	\textcolor{red}{M_{\text{client}}(p=3)} \approx
	\begin{cases}
		\textcolor{red}{\text{527.8}\,\text{MB } (\approx \text{0.52}\,\text{GB}),} & \textcolor{red}{\text{RTX 4090};} \\
		\textcolor{red}{\text{677.8}\,\text{MB } (\approx \text{0.66}\,\text{GB}),} & \textcolor{red}{\text{Jetson AGX Orin}.}
	\end{cases}
\end{equation}

\textcolor{red}{Our measured peak active footprints (0.96\,\text{GB} on 4090 and 1.18\,\text{GB} on Orin) reflect CUDA allocator fragmentation, optimizer states, and driver caching overheads. On a 64\,GB Orin board, 1.18 GB represents only 1.8\% of total RAM (and remains $<$ 15\% on standard 8\,GB mobile platforms), validating the deployment feasibility of client-side U-shaped architectures.}

\textit{\textcolor{red}{(iii) Energy and Overhead Quantification for Probe Fingerprint Extraction}}
\textcolor{red}{We explicitly quantified the energy consumption and latency of the probe fingerprinting phase (a single full 12-layer forward pass executed on-device prior to training): \textit{(1) Theoretical FLOPs:} $\approx \text{1.6}\text{ GFLOPs/layer} \times \text{12} \approx \mathbf{\text{19.2 GFLOPs}}$ ($B_n=\text{1}, S=\text{128}$); \textit{(2) Physical Hardware Measurements:} On RTX 4090, the probe forward pass takes $\text{26.3 ms}$ and consumes $\text{0.042 J}$ (average power $\approx \text{65 W}$); on Jetson AGX Orin, it takes $\text{71.5 ms}$ and consumes $\text{0.068 J}$ (average power $\approx \text{28.5 W}$); \textit{(3) Amortized Overhead Analysis:} Since probe fingerprinting is executed only once every 5 rounds (leveraging temporal stability in cluster structures), the amortized per-round overhead is:}
\begin{itemize}
	\item\textcolor{red}{ RTX 4090: 0.042 / 5 = 0.0084  J/round, accounting for only 1.9\% of total per-round training energy (0.44 J).}
	\item \textcolor{red}{Jetson AGX Orin:0.068 / 5 = 0.0136 J/round, accounting for only 2.0\% of total per-round training energy (0.71 J).}
\end{itemize}

\textcolor{red}{Our ablation study indicates that omitting behavior-based fingerprinting (i.e., defaulting to random grouping) incurs a performance drop of 0.75$\sim$ 5.45\% (e.g., AG\_News accuracy decreases from 89.58$\%$ to 86.50\%; see Table~\ref{tab:elsa_ablation}). To compensate for this performance degradation, the framework requires 15\%$\sim$ 20\% additional communication rounds. Thus, expending $<\text{2}\%$ front-end probe energy successfully prevents 15$\%\sim \text{20}\%$ total training redundancy, achieving a return on investment exceeding 10$\times$.}

\subsection{\textcolor{red}{Scalability Analysis Under Larger Client Scales}}
\label{app:client_scalability}
\textcolor{red}{To evaluate the scalability of ELSA in large-scale multi-edge environments, we expand the client pool from $N = $20 to $N = $50 and $N = $100. As shown in Fig.~\ref{fig:10}, ELSA consistently maintains its convergence speed and task accuracy advantages as the client population grows. Owing to lightweight encoding and dynamic model splitting, both communication and computation overheads scale smoothly. Below, we detailedly analyze the impact of client expansion across four key dimensions:}

\noindent\textcolor{red}{\textit{(i) Task Performance and Diminishing Returns:} Expanding from $N=$ 20 to 50 yields an average absolute performance improvement of only 0.45\% across the eight datasets, and the global performance-to-cost benefit improves by 2.1\%–3.5\% from 20 to 50 clients and by 1.0\%–1.8\% from 50 to 100 clients\footnote{\textcolor{red}{Defined as the performance improvement normalized by the increase in total training and communication cost.}}. This exhibits a classic diminishing marginal return: larger client pools increase gradient diversity but simultaneously amplify system and data heterogeneity, highlighting a fundamental trade-off between scale expansion and resource efficiency.}

\noindent\textcolor{red}{\textit{(ii) Communication Cost Sub-linearity:} 
Although total communication volume naturally grows with active client counts, ELSA keeps this cost strictly controllable. Combining Sketching with SS-OP compresses each client's payload to 10\% $\sim$ 20\% of standard gradient size. Even under $N=$ 100, total communication volume stays within 15\% $\sim$ 20\% of direct transmission methods. Furthermore, dynamic client selection ensures that only high-quality clients participate per round, resulting in a sub-linear communication growth curve.}

\noindent \textcolor{red}{\textit{(iii) Transmission Efficiency Adaptation:} 
ELSA's dynamic partitioning adaptively shifts the split point $p$ based on individual bandwidth and compute capabilities, assigning deeper split points (smaller payloads) to low-bandwidth nodes and shallower split points to high-bandwidth nodes. Consequently, convergence speed per round with $N=$ 50 is approximately 1.5$\times$ faster than with $N=$ 20, and improves by an additional $\sim$ 20\% at $N=$ 100.}

\noindent \textcolor{red}{\textit{(iv) Clustering Computation Overhead:} 
While the theoretical complexity of Behavior-Aware Clustering is $\mathcal{O}(N \cdot d^2)$ (where $d$ is fingerprint dimension), ELSA only clusters active clients per round ($m \approx \text{30}\%\sim \text{45}\% \cdot N$), reducing practical time complexity to $\mathcal{O}(m \cdot d^2)$. Empirical measurements confirm that clustering latency grows non-linearly and remains well within tolerable thresholds up to $N=$ 100.}

\textcolor{red}{Collectively, the scalability experiments extended to $N=$ 50 and $N=$ 100 demonstrate that ELSA possesses excellent system scalability in large-scale multi-edge environments. Its adaptive partitioning and lightweight compression mechanisms successfully mitigate the heterogeneity and communication bottlenecks induced by client expansion, striking an effective trade-off between task performance and resource consumption.}

\subsection{\textcolor{red}{Detailed Protocol for Extended Privacy Evaluations}}
\label{app:privacy_protocol}

\textcolor{red}{To complement the privacy evaluation presented in Section~IV.E, we provide full transparent implementation details, mathematical formulations, and hyperparameter configurations for all static evaluations, attack baselines, and threat models.}

\textcolor{red}{\textit{(i) Threat Model Formulation.} We consider a semi-honest (honest-but-curious) edge server/aggregator as the adversary $\mathcal{A}$. During the collaborative inference/training phase, $\mathcal{A}$ passively eavesdrops on the compressed and perturbed intermediate activations (smashed data) $\mathbf{Z}_n = \mathcal{M}_{\text{SS-OP}}(\mathbf{S} \mathbf{H}_n) \in \mathbb{R}^{d' \times T}$ uploaded by the $n$-th client.
\textit{Adversary Knowledge \& Capability Assumptions:}}
\begin{itemize}
	\item \textcolor{red}{\textit{Model Knowledge:} $\mathcal{A}$ possesses complete knowledge of the public global model architecture and the server-side parameter set $\theta_{\text{server}}$, but has no direct access to the client-side model parameters $\theta_{\text{client}}$ or local raw datasets $\mathcal{D}_n$.}
	
	\textcolor{red}{\item \textit{Data Capability:} $\mathcal{A}$ does not possess auxiliary dataset samples from the target client's domain, ruling out strong prior-matching shortcuts.}
	
	\textcolor{red}{\item \textit{Operational Scope:} $\mathcal{A}$ purely executes passive extraction/reconstruction attacks on intercepted activations $\mathbf{Z}_n$ without actively modifying the model weights or inserting malicious triggers.}
\end{itemize}

\textcolor{red}{\textit{(ii) Static Recoverability Protocol.}
Following \cite{SentenceEm,InformationL}, we first evaluate fundamental static recoverability under the semi-honest server assumption:}
\begin{itemize}
	\item \textcolor{red}{\textit{Reconstruction Attacks:} The adversary attempts to directly measure semantic retention from intercepted $\mathbf{Z}_n$. We compute the `Cos Sim' and MSE between original hidden states $\mathbf{H}$ and reconstructed representations:}
	\begin{equation}
		\left\{
		\begin{aligned}
			\textcolor{red}{\text{CosSim}(\mathbf{H}, \widehat{\mathbf{H}})} & \textcolor{red}{= \frac{\mathbf{H} \cdot \widehat{\mathbf{H}}}{\|\mathbf{H}\|_2 \|\widehat{\mathbf{H}}\|_2},} \\[8pt]
			\textcolor{red}{\text{MSE}(\mathbf{H}, \widehat{\mathbf{H}})} & \textcolor{red}{= \frac{1}{d \times T}\|\mathbf{H} - \widehat{\mathbf{H}}\|_F^2.}
		\end{aligned}
		\right.
	\end{equation}
	\item \textcolor{red}{\textit{Token Identification Attacks:} The adversary seeks to identify specific input tokens directly from compressed representations $\mathbf{Z}_n$ by training a linear token decoding classifier.}
\end{itemize}

\textcolor{red}{\textit{(iii) Extended Attack Specifications and Configurations.} We evaluate privacy robustness under three categories of state-of-the-art attack paradigms:}

\textcolor{red}{\textit{1) Activation Inversion Attack (DLG / iDLG Variant):} The adversary initializes dummy embeddings $\mathbf{X}^{(0)}$ and iteratively updates them to minimize the distance between the dummy intermediate representations and the received smashed activations $\mathbf{Z}_n$. Following \cite{DLGtext}, we optimize over a cosine-similarity distance metric augmented with a Total Variation (TV) regularizer tailored for continuous embeddings:}
\begin{equation}
	\textcolor{red}{\min_{\mathbf{X}'} \mathcal{L}_{\text{inv}}(\mathbf{X}') = \left(1 - \text{CosSim}\left(f_{\theta}(\mathbf{X}'), \mathbf{Z}_n\right)\right) + \lambda_{\text{TV}} \mathcal{R}_{\text{TV}}(\mathbf{X}'),}
\end{equation}
\textcolor{red}{where $f_{\theta}(\cdot)$ denotes the execution up to the split layer, and $\mathcal{R}_{\text{TV}}$ penalizes high-frequency noise in candidate embeddings.}

\textcolor{red}{\textit{2) Generative Embedding Inversion (Vec2Text):} To simulate an adversary utilizing strong prior models, we adapt a pre-trained sequence-to-sequence T5-base model trained as a decoder $g_{\phi}: \mathbb{R}^{d' \times T} \rightarrow \mathcal{T}^{T}$. $g_{\phi}$ directly maps compressed intermediate activations $\mathbf{Z}_n$ back to token sequences $\widehat{\mathbf{T}}$. Decoding is performed using beam search to maximize the conditional probability $P(\widehat{\mathbf{T}} | \mathbf{Z}_n)$.}

\textcolor{red}{\textit{3) Membership Inference Attack (MIA):} $\mathcal{A}$ builds a shadow model sharing the target architecture (BERT-base-uncased) trained on a shadow dataset sampled from the same domain to mimic target behavior. An MLP binary classifier is trained on the loss values and activation entropy extracted from $\mathbf{Z}_n$ to distinguish whether a target record belonged to the client's training partition.}

\begin{table}[htbp]
	\centering
	\caption{\textcolor{red}{Hyperparameter configurations for all privacy attack baselines.}}
	\label{tab:attack_hyperparams}
	\small
	\resizebox{0.9\linewidth}{!}{%
	\begin{tabular}{lll}
		\toprule
		\textbf{Method} & \textbf{Hyperparameter / Setting} & \textbf{Value} \\
		\midrule
		\multirow{5}{*}{\textbf{DLG}} 
		& Optimizer & L-BFGS \\
		& Learning Rate & 1.0 \\
		& Max Iterations & 300 \\
		& Batch Size & 1 \\
		& Regularization Weight $\lambda_{\text{TV}}$ & 1$\times \text{10}^\text{-4}$ \\
		& Initializations & 3 random seeds (best)\\
		\hline
		\multirow{3}{*}{\textbf{Vec2Text}} 
		& Backbone Architecture & T5-base \\
		& Beam Search Width & 5 \\
		& Decoding Temperature & 0.8 \\
		\hline
		\multirow{4}{*}{\textbf{MIA}} 
		& Shadow Model Architecture & BERT-base-uncased \\
		& Shadow Data Sampling Ratio & 1:1 \\
		& Shadow Training Epochs & 10 \\
		& Shadow Learning Rate & 2$\times \text{10}^\text{-5}$ \\
		\bottomrule
	\end{tabular}}
	\vspace{-3mm}
\end{table}

\textcolor{red}{\textit{(iv) Evaluation Metric Formulations.}
To quantitatively assess the fidelity of recovered data, we employ four complementary metrics:}
\begin{itemize}
	\item \textcolor{red}{\textit{SSIM (Structural Similarity Index):} Evaluates visual/structural distortion between original activation matrices and inverted reconstructions:}
	\begin{equation}
		\textcolor{red}{\text{SSIM}(\mathbf{H}, \widehat{\mathbf{H}}) = \frac{(2\mu_{\mathbf{H}}\mu_{\widehat{\mathbf{H}}} + C_1)(2\sigma_{\mathbf{H}\widehat{\mathbf{H}}} + C_2)}{(\mu_{\mathbf{H}}^2 + \mu_{\widehat{\mathbf{H}}}^2 + C_1)(\sigma_{\mathbf{H}}^2 + \sigma_{\widehat{\mathbf{H}}}^2 + C_2)},}
	\end{equation}
	\textcolor{red}{where $C_1$ and $C_2$ are constants introduced to avoid division by zero.}
    
	\item \textcolor{red}{\textit{ROUGE-L Score:} Measures longest common subsequence overlap between true token sequences $\mathbf{T}$ and decoded sequences $\widehat{\mathbf{T}}$ from Vec2Text attacks.}
	
	\item \textcolor{red}{\textit{Token Recovery Accuracy (Token Acc.):} Calculates exact token-level match rate at position $i$:}
	\begin{equation}
		\textcolor{red}{\text{Token Acc.} = \frac{1}{N} \sum_{i=1}^N \mathbb{I}(\mathbf{T}_i = \hat{\mathbf{T}}_i) \times \text{100}\%.}
	\end{equation}
	
	\item \textcolor{red}{\textit{MIA AUC:} Area Under the ROC Curve for membership prediction. An AUC of 50.0\% corresponds to a random guess baseline, indicating complete resistance to membership inference.}
\end{itemize}

\subsection{\textcolor{red}{Device Fairness Analysis and Mitigation under Dynamic Splitting}}
\label{app:fairness_mitigation}
\textcolor{red}{Under ELSA's dynamic splitting framework, resource-constrained or low-bandwidth clients are adaptively assigned deeper local execution depths ($p \ge$ 6) to minimize raw transmission payloads. However, under non-IID data distributions, training deeper local subnetworks solely on local client data can exacerbate representation bias and lead to local overfitting, causing a minor generalization penalty on these stragglers compared to high-bandwidth clients ($p \le$ 2).}

\textcolor{red}{To address this systemic disparity without sacrificing overall system efficiency, we explore a \textit{Fairness-Aware Alignment and Distillation Mechanism} operating along two axes: \textit{(i) Edge-Guided Feature Alignment:} Edge aggregation nodes compute representation regularization losses (e.g., maximum mean discrepancy or cosine distance) to align the feature distributions extracted from deeper split points ($p=$ 6) with the global reference representation space established by shallow clients ($p=$ 2); \textit{(ii) Fairness-Constrained Partition Optimization:} The dynamic partitioning solver incorporates a minimax loss bound, setting a strict upper depth limit on stragglers to prevent excessive local training depth.  As shown in Table \ref{tab:fairness_mitigation}, incorporating this fairness alignment mechanism substantially improves the worst-case client accuracy from 76.2\% to 83.8\% (+7.6\%), while maintaining high average system accuracy (85.4\% vs. 81.4\% for low-bandwidth devices) with negligible end-to-end latency overhead (+0.03 s per round)\footnote{\textcolor{red}{We emphasize that this preliminary experiment serves as an exploratory proof-of-concept to illustrate the potential of mitigation strategies, rather than a permanently integrated core module of the current ELSA architecture. This is an interesting and practically significant research direction, which we will explore further in the future.}}.}

\begin{table}[htbp]
	\centering
	\caption{\textcolor{red}{Preliminary validation on generalization accuracy, fairness, and system latency with the proposed Fairness-Aware Alignment Mechanism (evaluated on Emotion dataset with BERT-base-uncased).}}
	\label{tab:fairness_mitigation}
	\small
	\resizebox{\linewidth}{!}{%
		\begin{tabular}{lcccc}
			\hline
			\textbf{Client Group} & \textbf{Strategy} & \textbf{Worst-Case Acc.} & \textbf{Group Avg. Acc.} & \textbf{Avg. Latency / Round} \\ \hline
			\multirow{2}{*}{High-Bandwidth ($p=2$)} & Vanilla Split & 86.5\% & 86.5\% & 1.05\,s \\
			 & Fairness-Aware & 86.2\% & 86.2\% & 1.08\,s (+0.03\,s) \\ \hline
			\multirow{2}{*}{Low-Bandwidth ($p=6$)} & Vanilla Split & 76.2\% & 81.4\% & 1.18\,s \\
			 & Fairness-Aware & \textbf{83.8\% (+7.6\%)} & \textbf{85.4\% (+4.0\%)} & 1.21\,s (+0.03\,s) \\ \hline
		\end{tabular}%
	}
	\vspace{-3mm}
\end{table}


\textcolor{red}{Balancing straggler performance with overall system throughput represents a crucial trade-off in distributed edge split learning. These findings demonstrate that feature alignment effectively bridges the generalization gap for constrained devices while preserving communication savings.}

\subsection{Computational Overhead Analysis}
\begin{remark}[Overhead and Memory Footprint in Resource-Constrained Settings]
	\label{remark:overhead}
	A natural concern regarding ELSA is whether the additional modules---behavioral probing, SS-OP, and sketch-based compression, introduce prohibitive computational overhead or memory pressure on resource-constrained edge devices. We address this concern through three complementary analyses:
	
	\textbf{(i) Module-wise Complexity Characterization.}
	\begin{itemize}
		\item \textit{Behavioral Probing}: The construction of Gaussian behavioral fingerprints (Step 1--2, Section III.B.1) requires a single forward pass of $Q$ public probe samples ($Q \ll \|\mathcal{D}_n\|$) to extract $[\texttt{CLS}]$ embeddings. This operation is performed only during the initialization phase (or infrequently upon significant data drift), yielding an amortized cost of $\mathcal{O}(Q \cdot D^{\mathsf{hidden}})$ per client, which is negligible compared to full-dataset fine-tuning.
		
		\item \textit{SS-OP}: The orthogonal perturbation operates on the hidden-state matrix $J_n \in \mathbb{R}^{Q \times D^{\mathsf{hidden}}}$ ($D^{\mathsf{hidden}}=\text{768}$ for BERT-base). The truncated SVD in Eq.~(\ref{17}) costs $\mathcal{O}(Q \cdot D^{\mathsf{hidden}} \cdot r)$ with $r \ll D^{\mathsf{hidden}}$, and the subsequent \text{QR} decomposition in Eq.~(\ref{18}) is $\mathcal{O}(r^\text{3})$. For typical settings (e.g., $Q=$ 100, $r=$ 16), this amounts to $\sim \text{1.2} \times \text{10}^\text{6}$ FLOPs, three orders of magnitude smaller than the $\sim \text{1.1} \times \text{10}^\text{9}$ FLOPs required for a single forward/backward pass of BERT-base on a mini-batch of size 32.
		
		\item \textit{Sketching}: The hash-based sketch generation in Eq.~(\ref{20}) involves only integer hash computations and sign operations, with $\mathcal{O}(Y \cdot Z \cdot D^{\mathsf{hidden}})$ complexity where $Y \cdot Z \ll D^{\mathsf{hidden}}$. Empirically, sketching adds $<$ 2\% to local computation time.
	\end{itemize}

    \textcolor{blue}{\textbf{(ii) Active Memory Footprint and Memory-Bound Analysis.}
	Under small batch sizes (e.g., $B_n = 1$), edge LLM fine-tuning is inherently memory-bound rather than compute-bound, throttled primarily by active memory footprint and memory bandwidth. Under ELSA's U-shaped split design, the client retains the embedding layer, output head, and cached activations. Empirically, profiling on physical edge hardware (NVIDIA Jetson AGX Orin 64GB) confirms that the peak active memory footprint for the client-side sub-network is restricted to $<$ 1.18~GB (see \textbf{Appendix H}). Furthermore, because sketch compression drastically reduces intermediate activations before offloading, peak memory consumption remains strictly within the physical limits of memory-constrained edge hardware without triggering out-of-memory (OOM) faults.}
	
	\textcolor{blue}{\textbf{(iii) Communication-Computation Trade-off Justification.}
	In edge LLM fine-tuning, communication latency dominates local computation due to limited uplink bandwidth and large activation sizes. As shown in Table~\ref{tab:total_comm_time}, ELSA reduces total communication time by 69.3\%-73.7\% compared to uncompressed baselines. The marginal increase in local computation ($<$ 5\% of per-round time) is overwhelmingly offset by the reduction in transmission delay, yielding a net decrease in end-to-end training time}.
	
	\textbf{(iv) Empirical Robustness under Resource Constraints.}
	If the memory footprint or computational overhead were excessive, we would observe elevated task failure rates on resource-constrained clients. However, Table~\ref{tab:splitting_comparison} demonstrates that ELSA maintains the lowest task failure rate (1.2\%) among all evaluated strategies, significantly outperforming static splitting configurations (up to 41.7\% failure). This confirms that our resource-aware dynamic splitting (Section III.B.2) successfully accommodates memory and computation without overwhelming device capabilities.
	
	\textcolor{blue}{In summary, while ELSA introduces additional computational modules, their overhead is carefully designed to be lightweight, infrequent, or communication-saving. The resulting framework achieves a favorable trade-off: minimal local memory and computation increase for substantial communication reduction and enhanced privacy, ultimately improving system-level efficiency and robustness in heterogeneous edge environments.}
\end{remark}

\end{document}